\documentclass[10pt,journal,compsoc]{IEEEtran}
\usepackage{booktabs}
\usepackage{url}
\usepackage{epsfig}
\usepackage{subfigure}
\usepackage{graphics}
\usepackage{graphicx}
\usepackage{formular}
\usepackage{algorithm}
\usepackage{algorithmic}
\usepackage{multirow}
\usepackage{multicol}
\usepackage{xcolor}
\usepackage{clrscode}
\usepackage{makecell}
\usepackage{CJK}
\usepackage{amsthm}
\usepackage{balance}
\usepackage{times}
\usepackage{amsmath}
\usepackage{amsfonts}
\usepackage{stfloats}
\usepackage{array}
\usepackage{float}
\usepackage{url}
\usepackage{epstopdf}
\usepackage[centerlast]{caption2}
\usepackage{subfigure}
\usepackage{stfloats}
\usepackage{float}
\usepackage{cite}
\ifCLASSINFOpdf
\else
\fi
\begin{document}
\title{Micro-supervised Disturbance Learning: A Perspective of Representation Probability Distribution}
\author{Jielei Chu, ~\IEEEmembership{member,~IEEE}, Jing Liu, Hongjun Wang, Hua Meng,\\
Zhiguo Gong,~\IEEEmembership{Senior member,~IEEE}, Tianrui Li, ~\IEEEmembership{Senior member,~IEEE}
\thanks{Jielei Chu, Hongjun Wang, Hua Meng, Tianrui Li (the corresponding author) are with the School of Computing and Artificial Intelligence, Southwest Jiaotong University, Chengdu 611756, China. Tianrui Li is also with National Engineering Laboratory of Integrated Transportation Big Data Application Technology, Southwest Jiaotong University, Chengdu 611756, China. e-mails: \{jieleichu, huameng, wanghongjun, trli\}@swjtu.edu.cn.}
\thanks{Jing Liu is with the School of Business, Sichuan University, Sichuan, 610065, Chengdu, China. e-mail: liujing@scu.edu.cn}
\thanks{Zhiguo Gong is with the State Key Laboratory of Internet of Things for Smart City, Department of Computer and Information Science, University of Macau, Macau, China. Email: fstzgg@um.edu.mo}
}
\maketitle
\begin{abstract}
The instability is shown in the existing methods of representation learning based on Euclidean distance under a broad set of conditions. Furthermore, the scarcity and high cost of labels prompt us to explore more expressive representation learning methods which depends on the labels as few as possible. To address these issues, the small-perturbation ideology is firstly introduced on the representation learning model based on the representation probability distribution. The positive small-perturbation information (SPI) which only depend on two labels of each cluster is used to stimulate the representation probability distribution and then two variant models are proposed to fine-tune the expected representation distribution of RBM, namely, Micro-supervised Disturbance GRBM (Micro-DGRBM) and Micro-supervised Disturbance RBM (Micro-DRBM) models. The Kullback-Leibler (KL) divergence of SPI is minimized in the same cluster to promote the representation probability distributions to become more similar in Contrastive Divergence (CD) learning. In contrast, the KL divergence of SPI is maximized in the different clusters to enforce the representation probability distributions to become more dissimilar in CD learning. To explore the representation learning capability under the continuous stimulation of the SPI, we present a deep Micro-supervised Disturbance Learning (Micro-DL) framework based on the Micro-DGRBM and Micro-DRBM models and compare it with a similar deep structure which has not any external stimulation. Experimental results demonstrate that the proposed deep Micro-DL architecture shows better performance in comparison to the baseline method, the most related shallow models and deep frameworks for clustering.
\\
\end{abstract}
\section{Introduction}
Learning expressive representations is a fundamental and challenging problem in machine learning area \cite{6472238pami}. The success of many tasks such as person re-identification \cite{8607050tip}, zero-shot deep domain adaptation \cite{9361131pami2021}, visual recognition \cite{9052469pami2020}, fingerprint matching \cite{8937828pami}, and many others lie on deep representations of data. To this day, how to capture meaningful representation information from various data has always been the unremitting goal of machine learning. In recent years, much of the effort in researching representation learning methods goes into many new perspectives. To compare with high-dimensional representations of real-valued, compact representations aims to extract a discriminative and compressive feature. A compact representations learning (CRL) sub-network \cite{8606226pami} based on a Faster R-CNN \cite{7485869} has been successfully applied to tattoo examination. Since data usually contains complementary information from different views, multi-view representation learning is used to learn comprehensive representations \cite{8471216tkde}. For sharing representations among the data views, a shared multi-view data representation learning (SMDR) \cite{8618422pami2020} model is devised for dealing with cross-modality and cross-domain event detection. Sun et al. \cite{9113457pami} proposed a multi-view representation learning algorithm with deep Gaussian processes (MvDGPs), which integrates the advantages of multi-view representation learning and deep learning. \\
\indent From the perspective of transferable representation and graph structure representation, many methods have also emerged for representational learning. To match different domain distributions and reproduce kernel Hilbert space, Long et at. \cite{8454781pami2019} developed a deep adaptation networks (DAN) for domain adaptation. This transferable representation learning method enhances significantly feature transferability. Different distributions are exhibited between source domains and target in domain adaptation learning. To improve the universality of data representations, Tamaazousti et al. \cite{8703078pami} proposed an MuLDiPNet which has capability to automatically change an initial source problem and extract new features. Graph data representation and learning has been studied recently \cite{VGAE, WOS:000526526700003, 8128909tnnls, 9394783pami}. To address the effect of some graph structure noises, Jiang et al. presented a Graph elastic Convolution Network (GeCN) \cite{9394783pami} based on GCN \cite{VGAE} for semi-supervised classification (10\% $\sim$ 30\% label rate).\\
\indent To reduce the reliance on the labels, there are various semi-supervised learning methods in the research field of weakly supervised learning which are widely used in classification \cite{8528884, 8662706}, clustering \cite{LIU201919, REN2019121}, discrete choice models \cite{8574052}, sentiment analysis \cite{PARK2019139}, and so on. In these applications, semi-supervised feature learning \cite{ZENG2019104787} is a critical phase to enhance the efficiencies and performances of the following learning tasks. Existing shallow semi-supervised feature learning methods \cite{8444719, 8528552} exploit various semi-supervised strategies to improve learning efficiency and performance while minimize the use of labels. Chen et al. \cite{8528552} presented a Sparse Rescaled Linear Square Regression (SRLSR) method which has capability to obtain more sparse regression coefficients for semi-supervised feature learning. Recently, there are some works which explore deep semi-supervised learning method \cite{8839597}, \cite{8998334} to promote the capabilities of deep feature learning. In general, the conventional semi-supervised learning methods perform within a fixed feature space. The DCS \cite{7937922} is an incremental deep semi-supervised learning method, which propagates information from labeled to unlabeled samples in the procedure of deep feature learning. In the DCS framework, the Co-Space stems from two CNN models by extracting features for all unlabeled and labeled samples. Sellami et al. \cite{SELLAMI2019246} constructed a semi-supervised 3-D Convolutional Neural Network (3-D CNN) for the spectro-spatial classification. The approach not only preserves the information of the relevant spectro-spatial but also enhances the classification using few labeled samples. Xue et al. \cite{8421642} proposed an efficient and fast semi-supervised DIOD method based on weakly deep semi-supervised joint sparse learning and Advanced Region Proposal Networks (ARPNs). Meng et al. \cite{8858038} presented a Semi-supervised Graph Regularized Deep NMF (SGDNMF) with bi-orthogonal constraints for data representation. The bi-orthogonal constraints are introduced into the SGDNMF model on two factor matrices for improving the representation performance with a small fraction of labels. The deep semi-supervised learning procedure is used in the decision making module to maximize the diagnostic efficiency and simultaneously minimize the human interaction. Mercado et al. \cite{mercadoNeurIPS2019} proposed a semi-supervised learning method on Multilayer Graphs (semi-MG) using a regularizer of the generalized matrix mean. In each individual graph layer, the labeled and unlabeled samples are fused together with the encoding information. Recently, an unsupervised feature learning method \cite{9165942} was developed based on ensemble learning and a shallow semi-supervised feature learning model, pcGRBM \cite{Chujielei2018pcGRBM}, was proposed. However, these model based on European distance shows somewhat instabilities. Besides instability, the semi-supervised feature learning model pcGRBM\cite{Chujielei2018pcGRBM} still depends on a certain quantity of labels to obtain higher performance.\\
\indent The motivation of this paper derives from the small-perturbation ideology of physical systems \cite{SMALLPERTURBATIONTHEORY}. In simple terms, the original state of the physical system changes slightly under the stimulation of small disturbance. So, there are two interesting problems: 1) Whether the small disturbance can be used to stimulate the representation learning model to fine-tune the expected representation probability distribution? 2) Whether the representation learning capability can significantly improve under the continuous stimulation of small disturbance? To achieve these goals, the positive small-perturbation information (SPI) is used to stimulate the representation learning process from the perspective of representation probability distribution and two variant models are proposed to fine-tune the expected representation distribution of RBM \cite{hinton2006reducing}, namely, Micro-supervised Disturbance GRBM (Micro-DGRBM) and Micro-supervised Disturbance RBM (Micro-DRBM) models. The SPI only depends on two labels of each cluster. Hence, we term this learning pattern as Micro-supervised Disturbance Learning (Micro-DL). The Kullback-Leibler (KL) divergence \cite{SANKARAN201672} of SPI is minimized in the same cluster to promote the representation probability distributions become more similar in Contrastive Divergence (CD) learning \cite{hinton2002training}. In contrast, the KL divergence of SPI is maximized in the different clusters to enforce the representation probability distributions become more dissimilar in CD learning. Then we present a deep Micro-DL framework (see Fig. 1) to explore the effect of continuous stimulation of positive SPI based on Micro-DGRBM and Micro-DRBM models. Finally, we design the similar deep structure without the stimulation of positive SPI to compare the capability of representation learning with Micro-DL framework. Our Micro-DL method improves the stability \cite{HighDistanc} of representation learning from the perspective of representation probability distribution and significantly reduces the reliance on the labels. The contributions are as follows:
\begin{itemize}
  \item The small-perturbation ideology is firstly introduced on the representation learning model which based on the representation probability distribution. Then the positive small-perturbation information (SPI) is used to stimulate the representation learning process from the perspective of representation probability distribution and two variant models are proposed to fine-tune the expected representation distribution of RBM. They are Micro-supervised Disturbance GRBM (Micro-DGRBM) for modeling real-valued data and Micro-supervised Disturbance RBM (Micro-DRBM) for modeling binary data, respectively.
  \item To explore the representation learning capability under the continuous stimulation of small disturbance, we present a deep Micro-supervised Disturbance Learning (Micro-DL) framework based on the Micro-DGRBM and Micro-DRBM models and compare it with a similar deep structure which has not any stimulation of positive SPI.
  \item Experimental results demonstrate that the proposed deep Micro-DL architecture shows better performance in comparison to the baseline method, the most related shallow models and deep frameworks for clustering. Furthermore, our Micro-DL method improves the stability of representation learning from the perspective of representation probability distribution and significantly reduces the reliance on the labels.
\end{itemize}
\indent The rest of the paper is organized as follows. Section 2 introduces the theoretical background. Section 3 presents Micro-DL method, include two novel shallow micro-supervised disturbance representation learning models, learning algorithms and a novel deep Micro-DL architecture. Section 4 shows all experimental results. Finally, Section 5 summarizes our contributions.
\section{Background}
\subsection{Kullback-Leibler Divergence}
The KL divergence \cite{SANKARAN201672}, \cite{PONTI2017470}, \cite{8666051} has been popularly used to measure the difference between two probability distributions. Let $P(x)$ and $Q(x)$ be two probability distributions of a discrete random variable $x$. The KL divergence is given by
\begin{equation}
\begin{aligned}
  \textbf{\texttt{KL}}(P(x)\parallel Q(x))=\sum\limits_{x}P(x)\mathbf{\log}\frac{P(x)}{Q(x)}
 \end{aligned}
\end{equation}
where KL divergence is always non-negative ($\textbf{\texttt{KL}}(P\parallel Q)\ge0$) and $\textbf{\texttt{KL}}(P\parallel Q)=0$ if and only if $P=Q$. In other words, KL divergence is the expectation of the logarithmic difference between $P$ and $Q$. To obtain a distribution $P$ which is the closest to $Q$, we can minimize KL divergence of them. \\
\indent In neural networks, the method of updating parameters often directs towards minimizing the KL divergence\cite{GUREVICH2020103184}. KL divergence is used to measure the similarity between uncertain objects, then it can be merged with learning algorithms to improve the performance \cite{SHARMA2019100}. In computer vision, the KL divergence is used in training rotation-invariant RBM \cite{8870198} which can offer stability and consistency of representation. In variational Bayes Recurrent Neural Networks (BRNNs) \cite{01228}, the KL divergence between the approximate posterior and the prior distributions is a vital component of the models.
\subsection{Contrastive Divergence Learning}
CD learning \cite{hinton2002training}, \cite{carreira2005contrastive} as a fast learning method has been successfully applied to train RBMs. It proximately follows the gradient of two KL divergences and is defined as
\begin{equation}
\begin{aligned}
  \texttt{\textbf{CD}}_{n}=\texttt{\textbf{KL}}(p_{0}||p_{\infty})-\texttt{\textbf{KL}}(p_{n}||p_{\infty}),
\end{aligned}
 \end{equation}
where $p_{0}$ is the data distribution, $p_{\infty}$ is model distribution and $p_{n}$ is the distribution of the data after running the Markov chain for $n$ step.\\
\indent In the encoding process of RBMs model, the conditional probability $p(h_{j}=1|\texttt{\textbf{v}})$ is given by:
 \begin{equation}
p(h_{j}=1|\texttt{\textbf{v}})=\sigma(b_{j}+\sum\limits_{i}v_{i}w_{ij}),
\end{equation}
where $\texttt{\textbf{v}}=(v_{1},v_{2}, \cdots, v_{i},\cdots,v_{n})$ is a vector of visible layer, $\sigma$ is the sigmoid function, $w_{ij}$ is the connection parameter between the hidden and visible layers, $\mathbf{b}=(b_{1},b_{2}, \cdots, b_{j},\cdots,b_{m})$ is the bias parameter of hidden layers. In the reconstructed process of RBMs, the conditional probability $p(v_{i}=1|\texttt{\textbf{h}})$ is given by:
 \begin{equation}
p(v_{i}=1|\texttt{\textbf{h}})=\sigma(c_{i}+\sum\limits_{j}h_{j}w_{ij}),
\end{equation}
where $\texttt{\textbf{h}}=(h_{1},h_{2}, \cdots, h_{j},\cdots,h_{m})$ is a vector of hidden layer, $\mathbf{c}=(c_{1},c_{2}, \cdots, c_{i},\cdots,c_{n})$, $\mathbf{W}_{n\times m}$ is the bias parameter of visible layer. \\
\indent In the GRBM and its variants \cite{Chujielei2018pcGRBM}, \cite{freund1994unsupervised}, \cite{zhang2014supervised}, the hidden layers remain unchanged binary units, but visible layer units are replaced by Gaussian linear units. The conditional probability of reconstructed process takes the form
\begin{equation}
\begin{aligned}
  p(\textbf{\texttt{v}}|\textbf{\texttt{h}})=\mathcal{N}(\sum\textbf{\texttt{h}}\mathbf{W}^T+\mathbf{c},\sigma^{2}),
\end{aligned}
 \end{equation}
 where $\mathbf{c}$ is a connection matrix, $\mathcal{N}(\cdot)$ is a gaussian density. \\

\indent For fast training RBMs, CD learning with one step Gibbs sampleling (CD-1 learning) \cite{carreira2005contrastive} is defined by:
\begin{equation}
\begin{aligned}
  \texttt{\textbf{CD}}_{1}=\texttt{\textbf{KL}}(p_{0}||p_{\infty})-\texttt{\textbf{KL}}(p_{1}||p_{\infty}),
\end{aligned}
 \end{equation}
where $p_{1}$ is the distribution of the reconstructed data. Then the update rules of model parameters of RBMs are given by
\begin{equation}
\begin{aligned}
   &w_{ij}^{(\tau+1)}=w_{ij}^{(\tau)}+\varepsilon(\langle v_{i}h_{j}\rangle_{0}-\langle v_{i}h_{j}\rangle_{1}),
\end{aligned}
\end{equation}
\begin{equation}
\begin{aligned}
   b_{j}^{(\tau+1)}=b_{j}^{(\tau)}+\varepsilon(\langle h_{j}\rangle_{0}-\langle h_{j}\rangle_{1})
\end{aligned}
\end{equation}
and
\begin{equation}
\begin{aligned}
   &c_{i}^{(\tau+1)}=c_{i}^{(\tau)}+\varepsilon(\langle v_{i}\rangle_{0}-\langle v_{i}\rangle_{1}),
\end{aligned}
\end{equation}
where $\langle\cdot\rangle_{0}$ denotes an average concerning the data distribution, $\langle\cdot\rangle_{1}$ denotes an average concerning the reconstructed data distribution and $\varepsilon$ is learning rate.
\section{Micro-supervised Disturbance Learning}
In this section, we first present two novel shallow Micro-DGRBM and Micro-DRBM models, in which the positive SPI is used to stimulate the representation learning process from the perspective of representation probability distribution. To explore the representation learning capability under the continuous stimulation of SPI, we present a deep Micro-DL framework based on the Micro-DGRBM and Micro-DRBM models.
\subsection{The Micro-DGRBM and Micro-DRBM Models}
 Let $\mathcal{V}=\{\mathbf{\textbf{v}}_{1},\mathbf{\textbf{v}}_{2},\cdots,\mathbf{\textbf{v}}_{i},\cdots,\mathbf{\textbf{v}}_{M}\}$ be a visible layer data set of the Micro-DRBM model, where $\mathbf{\textbf{v}}_{i}$ is the $i$th row vector of $\mathcal{V}$ which represents the $i$th instance with $n$ features ($\mathbf{\textbf{v}}_{i}=(v_{i1},v_{i2},\cdots,v_{ij},\cdots,v_{in}$) and $M$ is the number of instance vector. Each visible layer instance corresponds to a feature vector in the hidden layer. The hidden feature vector set of the Micro-DRBM model is denoted as $\mathcal{H}=\{\mathbf{\textbf{h}}_{1},\mathbf{\textbf{h}}_{2},\cdots,\mathbf{\textbf{h}}_{i},\cdots,\mathbf{\textbf{h}}_{M}\}$, where $m$ is the dimensionality  of hidden layer and $\mathbf{\textbf{h}}_{i}$ is the hidden feature of $\mathbf{\textbf{v}}_{i}$. Similarly, the visible layer data of the Micro-DGRBM model and its corresponding hidden feature sets are denoted as $\widetilde{\mathcal{V}}=\{\widetilde{\mathbf{\textbf{v}}_{1}},\widetilde{\mathbf{\textbf{v}}_{2}},\cdots,\widetilde{\mathbf{\textbf{v}}_{i}},\cdots,\widetilde{\mathbf{\textbf{v}}_{M}}\}$ and $\widetilde{\mathcal{H}}=\{\widetilde{\mathbf{\textbf{h}}_{1}},\widetilde{\mathbf{\textbf{h}}_{2}},\cdots,\widetilde{\mathbf{\textbf{h}}_{i}},\cdots,\widetilde{\mathbf{\textbf{h}}_{M}}\}$, respectively.\\
\indent We randomly select only two visible layer vectors $\mathbf{\textbf{v}}_{f}$ and $\mathbf{\textbf{v}}_{g}$ from the visible layer data set of Micro-DRBM in each class. It is desirable that the probability distributions of hidden layer features $\mathbf{\textbf{h}}_{f}$ and $\mathbf{\textbf{h}}_{g}$ are similar. Thus, a similar feature distribution (SFD) set is defined as $\mathcal{SFD}=\{(\mathbf{\textbf{h}}_{f},\mathbf{\textbf{h}}_{g})|P(\mathbf{\textbf{h}}_{f}|\mathbf{\textbf{v}}_{f}) $ and $P(\mathbf{\textbf{h}}_{g}|\mathbf{\textbf{v}}_{g})$ are similar$\}$ for modeling Micro-DRBM. Supposing the visible layer data set $\mathcal{V}$ consists of $K$ clusters, then the $\mathcal{SFD}$ is a small set with $K$ two-tuples. Similarly, we randomly select only two visible layer vectors $\mathbf{\widetilde{\textbf{v}}_{f}}$ and $\mathbf{\widetilde{\textbf{v}}_{g}}$ from the visible layer data set of Micro-DGRBM in each class. It is desirable that the probability distributions of hidden layer features $\mathbf{\widetilde{\textbf{h}}_{f}}$ and $\mathbf{\widetilde{\textbf{h}}_{g}}$ are also similar. Thus, another SFD set is defined as $\widetilde{\mathcal{SFD}}=\{(\mathbf{\widetilde{\textbf{h}}_{f}},\mathbf{\widetilde{\textbf{h}}_{g}})|P(\mathbf{\widetilde{\textbf{h}}_{f}}|\widetilde{\mathbf{\textbf{v}}_{f}}) $ and $P(\mathbf{\widetilde{\textbf{h}}_{g}}|\widetilde{\mathbf{\textbf{v}}_{g}})$ are similar$\}$ for modeling Micro-DGRBM. Supposing the visible layer data set $\mathcal{\widetilde{\mathcal{V}}}$ consists of $\widetilde{K}$ clusters, then the $\widetilde{\mathcal{SFD}}$ is a small set with $\widetilde{K}$ two-tuples.\\
\indent In consideration of the differences of feature distributions in hidden layer between two classes, a dissimilar feature distributions (DFD) set can be defined as $\mathcal{DFD}=\{(\mathbf{\textbf{h}}_{r},\mathbf{\textbf{h}}_{s})|P(\mathbf{\textbf{h}}_{r}|\mathbf{\textbf{v}}_{r}) $ and $P(\mathbf{\textbf{h}}_{s}|\mathbf{\textbf{v}}_{s})$ are dissimilar$\}$ for modeling Micro-DRBM. Similarly, another DFD set can be defined as $\widetilde{\mathcal{DFD}}=\{(\mathbf{\widetilde{\textbf{h}}_{r}},\mathbf{\widetilde{\textbf{h}}_{s}})|P(\mathbf{\widetilde{\textbf{h}}_{r}}|\mathbf{\widetilde{\textbf{v}}_{r}}) $ and $P(\mathbf{\widetilde{\textbf{h}}_{s}}|\mathbf{\widetilde{\textbf{v}}_{s}})$ are dissimilar$\}$ for modeling Micro-DGRBM.\\
\indent In the process of representation learning of the Micro-DRBM model, we expect the probability distribution between the elements of two-tuples in $\mathcal{SFD}$ subset to be as similar as possible. Moreover, we also expect the probability distribution between the elements of two-tuples in $\mathcal{DFD}$ subset to be as dissimilar as possible. To achieve these targets, the KL divergences between the elements of two-tuples in $\mathcal{SFD}$ and $\mathcal{DFD}$ sets are integrated into CD learning in the process of representation learning of the Micro-DRBM model. Similarly, the KL divergences between the elements of two-tuples in $\widetilde{\mathcal{SFD}}$ and $\widetilde{\mathcal{DFD}}$ sets can be fused into CD learning of Micro-DGRBM model.\\
\indent Clearly, the above $\mathcal{SFD}$, $\widetilde{\mathcal{SFD}}$, $\mathcal{DFD}$ and $\widetilde{\mathcal{DFD}}$ are used as SPI to stimulate the representation learning process from the perspective of representation probability distribution and then to fine-tune the expected representation distribution. The Micro-supervised disturbance method is defined as follows:
 \begin{equation}
\begin{aligned}
  \min\limits_{\theta}&\frac{1}{K_{S}}\sum\limits_{\mathcal{SFD}} \textbf{\texttt{KL}}\big(P(\mathbf{\textbf{h}}_{f}|\mathbf{\textbf{v}}_{f})\parallel P(\mathbf{\textbf{h}}_{g}|\mathbf{\textbf{v}}_{g})\big)\\
  &-\frac{1}{K_{D}}\sum\limits_{\mathcal{DFD}}\textbf{\texttt{KL}}\big(P(\mathbf{\textbf{h}}_{r}|\mathbf{\textbf{v}}_{r})\parallel P(\mathbf{\textbf{h}}_{s}|\mathbf{\textbf{v}}_{s})\big),
\end{aligned}
\end{equation}
where $(\mathbf{\textbf{v}}_{f},\mathbf{\textbf{v}}_{g})\in \mathcal{SFD}$ and $(\mathbf{\textbf{v}}_{r},\mathbf{\textbf{v}}_{s})\in \mathcal{DFD}$,  $\theta=\{\mathbf{W},\mathbf{b}, \mathbf{c}\}$ is the parameter of Micro-DRBM, $K_{S}$ and $K_{D}$ are the numbers of two-tuples in $\mathcal{SFD}$ and $\mathcal{DFD}$ sets, respectively. Under the stimulation of these Micro-supervised disturbance, we expect that the representation probability distributions become more similar and dissimilar in the same and different clusters, respectively. Similarly, the Micro-supervised disturbance method for Micro-DGRBM model is given by:
 \begin{equation}
\begin{aligned}
  \min\limits_{\widetilde{\theta}}&\frac{1}{\widetilde{K_{S}}}\sum\limits_{\widetilde{\mathcal{SFD}}} \textbf{\texttt{KL}}\big(P(\mathbf{\widetilde{\textbf{h}}_{f}}|\widetilde{\mathbf{\textbf{v}}_{f}})\parallel P(\widetilde{\mathbf{\textbf{h}}_{g}}|\widetilde{\mathbf{\textbf{v}}_{g}})\big)\\
  &-\frac{1}{\widetilde{K_{D}}}\sum\limits_{\widetilde{\mathcal{DFD}}}\textbf{\texttt{KL}}\big(P(\widetilde{\mathbf{\textbf{h}}_{r}}|\widetilde{\mathbf{\textbf{v}}_{r}})\parallel P(\widetilde{\mathbf{\textbf{h}}_{s}}|\widetilde{\mathbf{\textbf{v}}_{s}})\big),
\end{aligned}
\end{equation}
where $(\widetilde{\mathbf{\textbf{v}}_{f}},\widetilde{\mathbf{\textbf{v}}_{g}})\in \widetilde{\mathcal{SFD}}$ and $(\widetilde{\mathbf{\textbf{v}}_{r}},\widetilde{\mathbf{\textbf{v}}_{s}})\in \widetilde{\mathcal{DFD}}$, $\widetilde{\theta}=\{\mathbf{\widetilde{W}},\mathbf{\widetilde{b}}, \mathbf{\widetilde{c}}\}$ is the parameter of Micro-DGRBM, $\widetilde{K_{S}}$ and  $\widetilde{K_{D}}$ are the numbers of two-tuples in $\widetilde{\mathcal{SFD}}$ and $\widetilde{\mathcal{DFD}}$ sets, respectively.\\
\indent The main goal here is to enhance the representation capabilities of Micro-DRBM and Micro-DGRBM models by fine-tuning the expected representation probability distribution with the stimulation of small disturbance in CD learning process. After running the Markov chain for one step, Eq. (2) can be further transformed to $\texttt{\textbf{CD}}_{1}=\texttt{\textbf{KL}}(p_{0}||p_{\infty})-\texttt{\textbf{KL}}(p_{1}||p_{\infty})$, which has been proven an effective and feasible CD learning method in previous works \cite{Chujielei2018pcGRBM}, \cite{9165942}. Based on the above considerations, the final objective function of the Micro-DRBM model is expressed as
\begin{equation}
\begin{aligned}
     \min\limits_{\theta}&\bigg\{-(1-\alpha)\Big(\textbf{\texttt{KL}}(p_{0}\parallel p_{\infty})-\textbf{\texttt{KL}}(p_{1}\parallel p_{\infty})\Big)\\
   &+\alpha\bigg[\frac{1}{K_{S}}\sum\limits_{\mathcal{SFD}} \textbf{\texttt{KL}}\big(P(\mathbf{\textbf{h}}_{f}|\mathbf{\textbf{v}}_{f})\parallel P(\mathbf{\textbf{h}}_{g}|\mathbf{\textbf{v}}_{g})\big)\\
   &-\frac{1}{K_{D}}\sum\limits_{\mathcal{DFD}}\textbf{\texttt{KL}}\big(P(\mathbf{\textbf{h}}_{r}|\mathbf{\textbf{v}}_{r})\parallel P(\mathbf{\textbf{h}}_{s}|\mathbf{\textbf{v}}_{s})\big)\bigg]\bigg\},
\end{aligned}
\end{equation}
where $\alpha\in(0,1)$ is a scale coefficient. Similarly, the objective function of the Micro-DGRBM model is defined as
\begin{equation}
\begin{aligned}
    \min\limits_{\widetilde{\theta}}&\bigg\{-(1-\alpha)\Big(\textbf{\texttt{KL}}(\widetilde{p_{0}}\parallel \widetilde{p_{\infty}})-\textbf{\texttt{KL}}(\widetilde{p_{1}}\parallel \widetilde{p_{\infty}})\Big)\\
   &+\alpha\bigg[\frac{1}{\widetilde{K_{S}}}\sum\limits_{\widetilde{\mathcal{SFD}}} \textbf{\texttt{KL}}\big(P(\widetilde{\mathbf{\textbf{h}}_{f}}|\widetilde{\mathbf{\textbf{v}}_{f}})\parallel P(\widetilde{\mathbf{\textbf{h}}_{g}}|\widetilde{\mathbf{\textbf{v}}_{g}})\big)\\
   &-\frac{1}{\widetilde{K_{D}}}\sum\limits_{\widetilde{\mathcal{DFD}}}\textbf{\texttt{KL}}\big(P(\widetilde{\mathbf{\textbf{h}}_{r}}|\widetilde{\mathbf{\textbf{v}}_{r}})\parallel P(\widetilde{\mathbf{\textbf{h}}_{s}}|\widetilde{\mathbf{\textbf{v}}_{s}})\big)\bigg]\bigg\}.
\end{aligned}
\end{equation}
\indent Next, we discuss how to solve above two multiobjective optimization problems and obtain the update rules of parameters $\theta$ and $\widetilde{\theta}$. In Eq. (12), the approximate gradient of $\textbf{\texttt{KL}}(p_{0}\parallel p_{\infty})-\textbf{\texttt{KL}}(p_{1}\parallel p_{\infty})$ with respect to parameters $\mathbf{W}$, $\mathbf{b}$ and $\mathbf{c}$ can be represented in the following form:
\begin{equation}
\begin{aligned}
-\frac{\partial}{\partial w_{ij}}\big(\textbf{\texttt{KL}}(p_{0}\parallel p_{\infty})-\textbf{\texttt{KL}}(p_{1}\parallel p_{\infty})\big)\approx \langle v_{i}h_{j}\rangle_{0}-\langle v_{i}h_{j}\rangle_{1},
\end{aligned}
\end{equation}
\begin{equation}
\begin{aligned}
-\frac{\partial}{\partial b_{j}}\big(\textbf{\texttt{KL}}(p_{0}\parallel p_{\infty})-\textbf{\texttt{KL}}(p_{1}\parallel p_{\infty})\big)\approx \langle h_{j}\rangle_{0}-\langle h_{j}\rangle_{1}
\end{aligned}
\end{equation}
and
\begin{equation}
\begin{aligned}
-\frac{\partial}{\partial c_{i}}\big(\textbf{\texttt{KL}}(p_{0}\parallel p_{\infty})-\textbf{\texttt{KL}}(p_{1}\parallel p_{\infty})\big)\approx \langle v_{i}\rangle_{0}-\langle v_{i}\rangle_{1}.
\end{aligned}
\end{equation}
Similarly, the approximate gradient of $\textbf{\texttt{KL}}(\widetilde{p_{0}}\parallel \widetilde{p_{\infty}})-\textbf{\texttt{KL}}(\widetilde{p_{1}}\parallel \widetilde{p_{\infty}})$ in Eq. (13) with respect to parameters $\mathbf{\widetilde{W}}$, $\mathbf{\widetilde{b}}$ and $\mathbf{\widetilde{c}}$ can be represented in the following form:
\begin{equation}
\begin{aligned}
-\frac{\partial}{\partial \widetilde{w_{ij}}}\big(\textbf{\texttt{KL}}(\widetilde{p_{0}}\parallel \widetilde{p_{\infty}})-\textbf{\texttt{KL}}(\widetilde{p_{1}}\parallel \widetilde{p_{\infty}})\big)\approx \langle \widetilde{v_{i}}\widetilde{h_{j}}\rangle_{0}-\langle \widetilde{v_{i}}\widetilde{h_{j}}\rangle_{1},
\end{aligned}
\end{equation}
\begin{equation}
\begin{aligned}
-\frac{\partial}{\partial \widetilde{b_{j}}}\big(\textbf{\texttt{KL}}(\widetilde{p_{0}}\parallel \widetilde{p_{\infty}})-\textbf{\texttt{KL}}(\widetilde{p_{1}}\parallel \widetilde{p_{\infty}})\big)\approx \langle \widetilde{h_{j}}\rangle_{0}-\langle \widetilde{h_{j}}\rangle_{1}
\end{aligned}
\end{equation}
and
\begin{equation}
\begin{aligned}
-\frac{\partial}{\partial \widetilde{c_{i}}}\big(\textbf{\texttt{KL}}(\widetilde{p_{0}}\parallel \widetilde{p_{\infty}})-\textbf{\texttt{KL}}(\widetilde{p_{1}}\parallel \widetilde{p_{\infty}})\big)\approx \langle \widetilde{v_{i}}\rangle_{0}-\langle \widetilde{v_{i}}\rangle_{1}.
\end{aligned}
\end{equation}
Then, one of critical tasks is how to obtain the partial derivatives of $\frac{1}{K_{S}}\sum\limits_{\mathcal{SFD}} \textbf{\texttt{KL}}\big(P(\mathbf{\textbf{h}}_{f}|\mathbf{\textbf{v}}_{f})\parallel P(\mathbf{\textbf{h}}_{g}|\mathbf{\textbf{v}}_{g})\big)-\frac{1}{K_{D}}\sum\limits_{\mathcal{DFD}}\textbf{\texttt{KL}}\big(P(\mathbf{\textbf{h}}_{r}|\mathbf{\textbf{v}}_{r})\parallel P(\mathbf{\textbf{h}}_{s}|\mathbf{\textbf{v}}_{s})\big)$ with respect to $\mathbf{W}$, $\mathbf{b}$ and $\mathbf{c}$. Another is how to obtain the partial derivatives of $\frac{1}{\widetilde{K_{S}}}\sum\limits_{\widetilde{\mathcal{SFD}}} \textbf{\texttt{KL}}\big(P(\widetilde{\mathbf{\textbf{h}}_{f}}|\widetilde{\mathbf{\textbf{v}}_{f}})\parallel P(\widetilde{\mathbf{\textbf{h}}_{g}}|\widetilde{\mathbf{\textbf{v}}_{g}})\big)-\frac{1}{\widetilde{K_{D}}}\sum\limits_{\widetilde{\mathcal{DFD}}}\textbf{\texttt{KL}}\big(P(\widetilde{\mathbf{\textbf{h}}_{r}}|\widetilde{\mathbf{\textbf{v}}_{r}})\parallel P(\widetilde{\mathbf{\textbf{h}}_{s}}|\widetilde{\mathbf{\textbf{v}}_{s}})\big)$ with respect to $\mathbf{\widetilde{W}}$, $\mathbf{\widetilde{b}}$ and $\mathbf{\widetilde{c}}$.\\
\indent Firstly, we rewrite $\textbf{\texttt{KL}}\big(P(\mathbf{\textbf{h}}_{f}|\mathbf{\textbf{v}}_{f})\parallel P(\mathbf{\textbf{h}}_{g}|\mathbf{\textbf{v}}_{g})\big)$ to the following equivalent form:
\begin{equation}
\begin{aligned}
 &\textbf{\texttt{KL}}\big(P(\mathbf{\textbf{h}}_{f}|\mathbf{\textbf{v}}_{f})\parallel P(\mathbf{\textbf{h}}_{g}|\mathbf{\textbf{v}}_{g})\big)\\
 &=\sum\limits_{x}p(h_{fx}=1|\mathbf{\textbf{v}}_{f})\mathbf{\log}\frac{p(h_{fx}=1|\mathbf{\textbf{v}}_{f})}{p(h_{gx}=1|\mathbf{\textbf{v}}_{g})}\\
 &=\sum\limits_{x}\Big[p(h_{fx}=1|\mathbf{\textbf{v}}_{f})\mathbf{\log}p(h_{fx}=1|\mathbf{\textbf{v}}_{f})\\
 &-p(h_{fx}=1|\mathbf{\textbf{v}}_{f})\mathbf{\log}p(h_{gx}=1|\mathbf{\textbf{v}}_{g})\Big].\\
\end{aligned}
\end{equation}
Thus, the gradient of Eq. (20) with respect to weight parameter $\mathbf{W}$ is given by:
\begin{equation}
\begin{aligned}
 &\frac {\partial}{\partial w_{ij}} \textbf{\texttt{KL}}\big(P(\mathbf{\textbf{h}}_{f}|\mathbf{\textbf{v}}_{f})\parallel P(\mathbf{\textbf{h}}_{g}|\mathbf{\textbf{v}}_{g})\big)=\\
&\sum\limits_{x}\bigg[\frac {\partial}{\partial w_{ij}}p(h_{fx}=1|\mathbf{\textbf{v}}_{f}) \mathbf{\log}p(h_{fx}=1|\mathbf{\textbf{v}}_{f})+\\
&\frac {\partial}{\partial w_{ij}}p(h_{fx}=1|\mathbf{\textbf{v}}_{f})-\frac {\partial}{\partial w_{ij}} p(h_{fx}=1|\mathbf{\textbf{v}}_{f})\mathbf{\log}p(h_{gx}=1|\mathbf{\textbf{v}}_{g})\\
  &-\frac{p(h_{fx}=1|\mathbf{\textbf{v}}_{f})\frac{\partial}{\partial w_{ij}} p(h_{gx}=1|\mathbf{\textbf{v}}_{g})}{p(h_{gx}=1|\mathbf{\textbf{v}}_{g})}\bigg]\\
&=\sum\limits_{x}\Bigg\{\frac {\partial}{\partial w_{ij}}p(h_{fx}=1|\mathbf{\textbf{v}}_{f})\Big[\mathbf{\log}p(h_{fx}=1|\mathbf{\textbf{v}}_{f})\\
&+\mathbf{\log}p(h_{gx}=1|\mathbf{\textbf{v}}_{g})+1\Big]-\frac{p(h_{fx}=1|\mathbf{\textbf{v}}_{f})\frac{\partial }{\partial w_{ij}}p(h_{gx}=1|\mathbf{\textbf{v}}_{g})}{p(h_{gx}=1|\mathbf{\textbf{v}}_{g})}\Bigg \}.
\end{aligned}
\end{equation}
\indent In encoding procedure of Micro-DRBM model, the transformation from the visible layer to the hidden layer is the sigmoid transform. So, the gradient $\frac {\partial}{\partial w_{ij}}p(h_{fx}=1|\mathbf{\textbf{v}}_{f})$ takes the form:
\begin{equation}
\begin{aligned}
 \frac {\partial}{\partial w_{ij}}p(h_{fx}=1|\mathbf{\textbf{v}}_{f})=\frac {\partial}{\partial w_{ij}}\sigma(b_{x}+\sum\limits_{i}^{n}v_{fi}w_{ix}).
\end{aligned}
\end{equation}
\indent When $x=j$, it is easy to obtain the gradient $\frac {\partial }{\partial w_{ij}}p(h_{fx}=1|\mathbf{\textbf{v}}_{f})$ as follows.
\begin{equation}
\begin{aligned}
&\frac {\partial }{\partial w_{ij}}p(h_{fx}=1|\mathbf{\textbf{v}}_{f})=\frac{v_{fi}e^{-(b_{j}+\sum\limits_{i}^{n}v_{fi}w_{ij})}}{\Big(1+e^{-(b_{j}+\sum\limits_{i}^{n}v_{fi}w_{ij})}\Big)^2}\\
 &=h_{fj}(1-h_{fj})v_{fi}.
\end{aligned}
\end{equation}
\indent In other cases, $\frac {\partial}{\partial w_{ij}}p(h_{fx}=1|\mathbf{\textbf{v}}_{f})=0$.
Similarly, the gradient $\frac {\partial}{\partial w_{ij}}p(h_{gx}=1|\mathbf{\textbf{v}}_{g})$ takes the form:
\begin{equation}
\begin{aligned}
 \frac {\partial}{\partial w_{ij}}p(h_{gx}=1|\mathbf{\textbf{v}}_{g})
 &=h_{gj}(1-h_{gj})v_{gi}.
\end{aligned}
\end{equation}
\indent Thus, Eq. (21) can be simplified as the following equivalent form:
\begin{equation}
\begin{aligned}
&\frac {\partial}{\partial w_{ij}}\textbf{\texttt{KL}}\big(P(\mathbf{\textbf{h}}_{f}|\mathbf{\textbf{v}}_{f})\parallel P(\mathbf{\textbf{h}}_{g}|\mathbf{\textbf{v}}_{g})\big)\\
&=v_{fi}h_{fj}(1-h_{fj})(\mathbf{\log}h_{fj}+\mathbf{\log}h_{gj}+1)-v_{gi}h_{gj}(1-h_{gj}).
\end{aligned}
\end{equation}
Similarly, the gradient of $\textbf{\texttt{KL}}\big(P(\mathbf{\textbf{h}}_{r}|\mathbf{\textbf{v}}_{r})\parallel P(\mathbf{\textbf{h}}_{s}|\mathbf{\textbf{v}}_{s})\big)$ with respect to weight parameter $\mathbf{W}$ is given by:
\begin{equation}
\begin{aligned}
&\frac {\partial}{\partial w_{ij}}\textbf{\texttt{KL}}\big(P(\mathbf{\textbf{h}}_{r}|\mathbf{\textbf{v}}_{r})\parallel P(\mathbf{\textbf{h}}_{s}|\mathbf{\textbf{v}}_{s})\big)\\
&=v_{ri}h_{rj}(1-h_{rj})(\mathbf{\log}h_{rj}+\mathbf{\log}h_{sj}+1)-v_{si}h_{sj}(1-h_{sj}).
\end{aligned}
\end{equation}
\indent From Eqs. (14), (25) and (26), the update rules of $\mathbf{W}$ can be represented in the following form:
\begin{equation}
\begin{aligned}
   &w_{ij}^{(\tau+1)}=w_{ij}^{(\tau)}+(1-\alpha)\varepsilon(\langle v_{i}h_{j}\rangle_{0}-\langle v_{i}h_{j}\rangle_{1})+\frac{\alpha}{K_{S}} \sum\limits_{\mathcal{SFD}}\\
   &\bigg[v_{fi}h_{fj}(1-h_{fj})(\mathbf{\log}h_{fj}+\mathbf{\log}h_{gj}+1)-v_{gi}h_{gj}(1-h_{gj})\bigg]\\
   &-\frac{\alpha}{K_{D}}\sum\limits_{\mathcal{DFD}}\bigg[v_{ri}h_{rj}(1-h_{rj})(\mathbf{\log}h_{rj}+\mathbf{\log}h_{sj}+1)\\
   &-v_{si}h_{sj}(1-h_{sj})\bigg],
\end{aligned}
\end{equation}
where $\varepsilon$ is a learning rate. \\
\indent When $x=j$, it is easy to obtain the gradients of $\textbf{\texttt{KL}}\big(P(\mathbf{\textbf{h}}_{f}|\mathbf{\textbf{v}}_{f})\parallel P(\mathbf{\textbf{h}}_{g}|\mathbf{\textbf{v}}_{g})\big)$ and $\textbf{\texttt{KL}}\big(P(\mathbf{\textbf{h}}_{r}|\mathbf{\textbf{v}}_{r})\parallel P(\mathbf{\textbf{h}}_{s}|\mathbf{\textbf{v}}_{s})\big)$ with respect to parameter $\mathbf{b}$ as follows.
\begin{equation}
\begin{aligned}
&\frac {\partial}{\partial b_{j}} \textbf{\texttt{KL}}\big(P(\mathbf{\textbf{h}}_{f}|\mathbf{\textbf{v}}_{f})\parallel P(\mathbf{\textbf{h}}_{g}|\mathbf{\textbf{v}}_{g})\big)\\
&=h_{fj}(1-h_{fj})(\mathbf{\log}h_{fj}+\mathbf{\log}h_{gj}+1)-h_{gj}(1-h_{gj})
\end{aligned}
\end{equation}
and
\begin{equation}
\begin{aligned}
&\frac {\partial}{\partial b_{j}} \textbf{\texttt{KL}}\big(P(\mathbf{\textbf{h}}_{r}|\mathbf{\textbf{v}}_{r})\parallel P(\mathbf{\textbf{h}}_{s}|\mathbf{\textbf{v}}_{s})\big)\\
&=h_{rj}(1-h_{rj})(\mathbf{\log}h_{rj}+\mathbf{\log}h_{sj}+1)-h_{sj}(1-h_{sj}).
\end{aligned}
\end{equation}
\indent From Eqs. (12), (15), (28) and (29), the update rules of $\mathbf{W}$ can be represented in the following form:
\begin{equation}
\begin{aligned}
   &b_{j}^{(\tau+1)}=b_{j}^{(\tau)}+(1-\alpha)\varepsilon(\langle h_{j}\rangle_{0}-\langle h_{j}\rangle_{1})+\\
   &\frac{\alpha}{K_{S}} \sum\limits_{\mathcal{SFD}}\big[h_{fj}(1-h_{fj})(\mathbf{\log}h_{fj}+\mathbf{\log}h_{gj}+1)-\\
   &h_{gj}(1-h_{gj})\big]-\frac{\alpha}{K_{D}}\sum\limits_{\mathcal{DFD}}\big[h_{rj}(1-h_{rj})(\mathbf{\log}h_{rj}+\mathbf{\log}h_{sj}\\
   &+1)-h_{sj}(1-h_{sj})\big]
\end{aligned}
\end{equation}
\indent As for model parameter $\mathbf{c}$, it's obvious that
\begin{equation}
\begin{aligned}
\frac {\partial}{\partial c_{i}} \textbf{\texttt{KL}}\big(P(\mathbf{\textbf{h}}_{f}|\mathbf{\textbf{v}}_{f})\parallel P(\mathbf{\textbf{h}}_{g}|\mathbf{\textbf{v}}_{g})\big)=0
\end{aligned}
\end{equation}
and
\begin{equation}
\begin{aligned}
\frac {\partial}{\partial c_{i}} \textbf{\texttt{KL}}\big(P(\mathbf{\textbf{h}}_{r}|\mathbf{\textbf{v}}_{r})\parallel P(\mathbf{\textbf{h}}_{s}|\mathbf{\textbf{v}}_{s})\big)=0.
\end{aligned}
\end{equation}
\indent From Eqs. (16), (31) and (32), the update rule of $\mathbf{c}$ takes the form:
\begin{equation}
\begin{aligned}
   &c_{i}^{(\tau+1)}=c_{i}^{(\tau)}+(1-\alpha)\varepsilon(\langle v_{i}\rangle_{0}-\langle v_{i}\rangle_{1}).
\end{aligned}
\end{equation}
Finally, the update rules of $\mathbf{W}$, $\mathbf{b}$ and $\mathbf{c}$ of the Micro-DRBM model are Eqs. (27), (30) and (33), respectively.\\
\indent In the Micro-DGRBM model, the hidden units remain binary, but the visible units are the linear units with Gaussian noise. Then, in the encoding process of Micro-DGRBM model, the conditional probability $p(\widetilde{h_{j}}=1|\widetilde{\texttt{\textbf{v}}})$ is given by:
 \begin{equation}
p(\widetilde{h_{j}}=1|\widetilde{\texttt{\textbf{v}}})=\sigma(\widetilde{b_{j}}+\sum\limits_{i}\widetilde{v_{i}}\widetilde{w_{ij}})
\end{equation}
and its conditional probability of reconstructed process takes the form:
\begin{equation}
\begin{aligned}
  p(\widetilde{\textbf{\texttt{v}}}|\widetilde{\textbf{\texttt{h}}})=\mathcal{N}(\sum\widetilde{\textbf{\texttt{h}}}\widetilde{\mathbf{W}}^T+\widetilde{\mathbf{c}},\sigma^{2}).
\end{aligned}
 \end{equation}
 The update rules of the parameters $\widetilde{\mathbf{W}}$, $\widetilde{\mathbf{b}}$ and $\widetilde{\mathbf{c}}$ of the Micro-DGRBM model are similar to the Micro-DRBM model, which takes the form:
\begin{equation}
\begin{aligned}
   &\widetilde{w_{ij}}^{(\tau+1)}=\widetilde{w_{ij}}^{(\tau)}+(1-\alpha)\varepsilon(\langle \widetilde{v_{i}}\widetilde{h_{j}}\rangle_{0}-\langle \widetilde{v_{i}}\widetilde{h_{j}}\rangle_{1})+\frac{\alpha}{\widetilde{K_{S}}} \sum\limits_{\widetilde{\mathcal{SFD}}}\\
   &\bigg[\widetilde{v_{fi}}\widetilde{h_{fj}}(1-\widetilde{h_{fj}})(\mathbf{\log}\widetilde{h_{fj}}+\mathbf{\log}\widetilde{h_{gj}}+1)-\widetilde{v_{gi}}\widetilde{h_{gj}}(1-\widetilde{h_{gj}})\bigg]\\
   &-\frac{\alpha}{\widetilde{K_{D}}}\sum\limits_{\widetilde{\mathcal{DFD}}}\bigg[\widetilde{v_{ri}}\widetilde{h_{rj}}(1-\widetilde{h_{rj}})(\mathbf{\log}\widetilde{h_{rj}}+\mathbf{\log}\widetilde{h_{sj}}+1)\\
   &-\widetilde{v_{si}}\widetilde{h_{sj}}(1-\widetilde{h_{sj}})\bigg],
\end{aligned}
\end{equation}

\begin{equation}
\begin{aligned}
   &\widetilde{b_{j}}^{(\tau+1)}=\widetilde{b_{j}}^{(\tau)}+(1-\alpha)\varepsilon(\langle \widetilde{h_{j}}\rangle_{0}-\langle \widetilde{h_{j}}\rangle_{1})+\\
   &\frac{\alpha}{\widetilde{K_{S}}} \sum\limits_{\widetilde{\mathcal{SFD}}}\big[\widetilde{h_{fj}}(1-\widetilde{h_{fj}})(\mathbf{\log}\widetilde{h_{fj}}+\mathbf{\log}\widetilde{h_{gj}}+1)-\\
   &\widetilde{h_{gj}}(1-\widetilde{h_{gj}})\big]-\frac{\alpha}{\widetilde{K_{D}}}\sum\limits_{\widetilde{\mathcal{DFD}}}\big[\widetilde{h_{rj}}(1-\widetilde{h_{rj}})(\mathbf{\log}\widetilde{h_{rj}}+\mathbf{\log}\widetilde{h_{sj}}\\
   &+1)-\widetilde{h_{sj}}(1-\widetilde{h_{sj}})\big]
\end{aligned}
\end{equation}
and
 \begin{equation}
\begin{aligned}
   &\widetilde{c_{i}}^{(\tau+1)}=\widetilde{c_{i}}^{(\tau)}+(1-\alpha)\varepsilon(\langle \widetilde{v_{i}}\rangle_{0}-\langle \widetilde{v_{i}}\rangle_{1}).
\end{aligned}
\end{equation}
\subsubsection{Learning Algorithms}
In this subsection, we show the learning algorithm of the proposed shallow Micro-DRBM and Micro-DGRBM models according to the update rules of their parameters.\\
\textbf{Algorithm 1: Micro-DRBM learning}\\
\noindent\line(1,0){250}\\
\textbf{Input}:\\
\indent \indent  $\mathcal{V}=\{\mathbf{\textbf{v}}_{1},\mathbf{\textbf{v}}_{2},\cdots,\mathbf{\textbf{v}}_{i},\cdots,\mathbf{\textbf{v}}_{N}\}$: visible layer data;\\
\indent \indent $\mathcal{SFD}$, $\mathcal{DFD}$: Micro-supervised disturbance sets.\\
\textbf{Output}:\\
\indent \indent   $\mathbf{W}$, $\mathbf{b}$ and $\mathbf{c}$: the parameters of Micro-DRBM.\\
\noindent\line(1,0){250}\\
 Step 1: Randomly initialize $\mathbf{W}$, $\mathbf{b}$ and $\mathbf{c}$.\\
 Step 2: Sample the states of the hidden layer units by $p(h_{j}=1|\texttt{\textbf{v}})=\sigma(b_{j}+\sum\limits_{i}v_{i}w_{ij})$.\\
 Step 3: Sample the states of the reconstructed visible layer units \\
  \indent \indent by $p(v_{i}=1|\texttt{\textbf{h}})=\sigma(c_{i}+\sum\limits_{j}h_{j}w_{ij})$.\\
Step 4: Update $\mathbf{W}$ parameter by Eq. (27).\\
Step 5: Update $\mathbf{b}$ parameter by Eq. (30).\\
Step 6: Update $\mathbf{c}$ parameter by Eq. (33).\\
Step 7: \textbf{While} iteration less than maximum \textbf{go to} Step 2.\\
step 8: \textbf{return} $\mathbf{W}$, $\mathbf{b}$ and $\mathbf{c}$.\\
\noindent\line(1,0){250}\\

\textbf{Algorithm 2: Micro-DGRBM learning}\\
\noindent\line(1,0){250}\\
\textbf{Input}:\\
\indent \indent  $\widetilde{\mathcal{V}}=\{\widetilde{\mathbf{\textbf{v}}_{1}},\widetilde{\mathbf{\textbf{v}}_{2}},\cdots,\widetilde{\mathbf{\textbf{v}}_{i}},\cdots,\widetilde{\mathbf{\textbf{v}}_{N}}\}$: visible layer data;\\
\indent \indent $\widetilde{\mathcal{SFD}}$, $\widetilde{\mathcal{DFD}}$: Micro-supervised disturbances sets.\\
\textbf{Output}:\\
\indent \indent   $\widetilde{\mathbf{W}}$, $\widetilde{\mathbf{b}}$ and $\widetilde{\mathbf{c}}$: the parameters of Micro-DGRBM.\\
\noindent\line(1,0){250}\\
 Step 1: Randomly initialize $\widetilde{\mathbf{W}}$, $\widetilde{\mathbf{b}}$ and $\widetilde{\mathbf{c}}$.\\
 Step 2: Sample the states of the hidden layer units by $p(\widetilde{h_{j}}=1|\widetilde{\texttt{\textbf{v}}})=\sigma(\widetilde{b_{j}}+\sum\limits_{i}\widetilde{v_{i}}\widetilde{w_{ij}})$.\\
 Step 3: Sample the states of the reconstructed visible layer units \\
  \indent \indent by $ p(\widetilde{\textbf{\texttt{v}}}|\widetilde{\textbf{\texttt{h}}})=\mathcal{N}(\sum\widetilde{\textbf{\texttt{h}}}\widetilde{\mathbf{W}}^T+\widetilde{\mathbf{c}},\sigma^{2})$.\\
 Step 4: Update $\widetilde{\mathbf{W}}$ parameter by Eq. (36).\\
 Step 5: Update $\widetilde{\mathbf{b}}$ parameter by Eq. (37).\\
 Step 6: Update $\widetilde{\mathbf{c}}$ parameter by Eq. (38).\\
 Step 7: \textbf{While} iteration less than maximum \textbf{go to} Step 2.\\
 step 8: \textbf{return} $\widetilde{\mathbf{W}}$, $\widetilde{\mathbf{b}}$ and $\widetilde{\mathbf{c}}$.\\
\noindent\line(1,0){250}
\subsection{Micro-DL Architecture}
In the proposed shallow Micro-DGRBM and Micro-DRBM models, the small-perturbation ideology is firstly introduced on the representation learning model based on the representation probability distribution. The positive SPI is used to stimulate the representation learning process from the perspective of representation probability distribution. To explore the representation learning capability under the continuous stimulation of small disturbance, we present a deep Micro-DL framework based on the Micro-DGRBM and Micro-DRBM models in this section. The deep Micro-DL framework consists of a stack of one Micro-DGRBM and $N$ Micro-DRBMs to explore superior performances for modeling continuous data. In the proposed deep architecture, the visible layer consists of Gaussian linear units and all hidden layers consist of binary units. In the process of CD learning of shallow Micro-DGRBM model, the SPI is used as small positive disturbance in the encoding and reconstruction procedures which adopt the binary and linear transformation by 1 step Gibbs sampling, respectively. However, the SPI is used as small positive disturbance in the encoding and reconstruction procedures which all adopt the binary transformation by 1 step Gibbs sampling in the process of CD learning of shallow Micro-DRBM model.\\
\indent The architecture of the proposed Micro-DL is shown in Fig. 1. The hidden feature vector of Micro-DGRBM model, $\mathbf{\textbf{h}}_{(1)}$, is the input of the next Micro-DRBM model. And the hidden feature vector of this Micro-DRBM model, $\mathbf{\textbf{h}}_{(2)}$, is the input of its next Micro-DRBM model. The final hidden feature vector of our Micro-DL architecture, $\mathbf{\textbf{h}}_{(N)}$, is the input of the following clustering task. In the stack of Micro-DRBMs, $\mathbf{\textbf{h}}_{(i)}^{'}$ is the reconstructed vector.\\
\indent In order to prove the effectiveness of the continuous stimulation of small disturbance, we design a most related deep representation learning framework, None Micro-supervised Disturbance Learning (NMicro-DL), which consists of one traditional shallow GRBM and $N$ RBM models. The only difference between Micro-DL and NMicro-DL architectures is that the continuous stimulation of small disturbance is used to fine-tune the expected representation distribution in the CD learning process of the former and the latter is an unsupervised deep representation learning framework.
\begin{figure*}[!htbp]
\vspace{0.5mm} \centering
   \includegraphics[scale=0.45025]{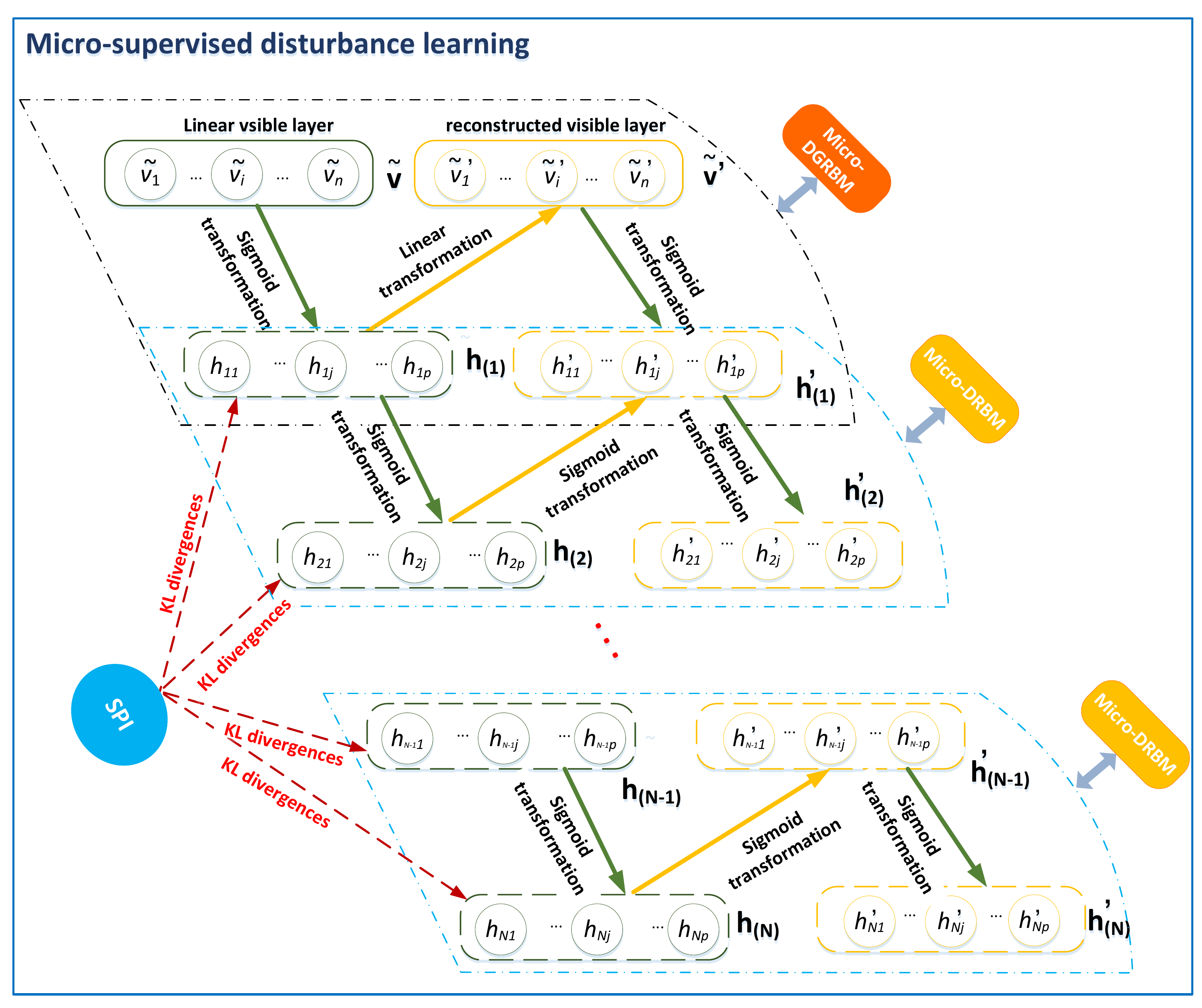}
 \caption{The deep Micro-DL architecture consists of a stack of one shallow Micro-DGRBM and $N-1$ shallow Micro-DRBMs. It has one Gaussian linear visible layer and $N$ binary hidden layers. The SPI is used to explore the representation learning capability of Micro-DL  under the continuous stimulation of small disturbance.}\label{fig:1}
\end{figure*}
\section{Experiments}
In this section, we compare the performance of the proposed Micro-DL with deep unsupervised learning models, shallow and deep semi-supervised learning models. All experiments have been conducted clustering tasks on twelve datasets. The details of experimental datasets are as follows:
\begin{itemize}
  \item Microsoft Research Asia Multimedia image sub-dataset (MSRA-MM)\cite{li2009msra}: Aquarium (922 images, 892 features and 3 classes), bathroom (924 images, 892 features and 3 classes), blog (943 images, 892 features and 3 classes), blood (866 images, 892 features and 3 classes), bouquet (880 images, 892 features and 3 classes), bugat (882 images, 892 features and 3 classes), cactus (919 images, 892 features and 3 classes), voituretuning (799 images, 899 features and 3 classes).
  \item car \footnote[1]{https://archive.ics.uci.edu/ml/datasets/car+evaluation}: This UCI dataset is derived from simple hierarchical decision mode, which contains 1728 instances, 6 features and 4 classes.
  \item KDD99 \footnote[2]{https://archive.ics.uci.edu/ml/datasets/kdd+cup+1999+data}: We choose 5904 instances from a large-scale dataset (KDD Cup 1999, 4000000 instances) randomly. It comprises 41 features and 3 classes.
  \item segmentation \footnote[3]{http://archive.ics.uci.edu/ml/support/image+segmentation}: This image segmentation dataset contains 2100 instances, 19 features and 7 classes.
  \item vowe \footnote[4]{https://sci2s.ugr.es/keel/dataset.php?cod=113}: This dataset contains information about speaker independent recognition which is a merge of the two original databases preset at the UCI repository \footnote[5]{http://archive.ics.uci.edu/ml/datasets.php}. It has 990 instances, 13 features and 11 classes.
\end{itemize}
In order to evaluate the representation capability of the proposed Micro-DL architecture, we compare the clustering performance of our Micro-DL against Semi-SP \cite{rangapuram2015constrained}, VGAE \cite{VGAE}, None Micro-DL (NMicro-DL), pcGRBM \cite{Chujielei2018pcGRBM}, Semi-EAGR \cite{7420705} and Semi-MG \cite{mercadoNeurIPS2019}. For all contrastive shallow models (pcGRBM and Semi-EAGR) and deep frameworks (VGAE, NMicro-DL and Semi-MG), the performance evaluations have two stages: one is feature learning and the other is clustering analysis. The output features of them are the input of Spectral Clustering \cite{ng2002spectral} algorithm. \\
\indent In the representation learning process of our Micro-DL and NMicro-DL architectures, the dimension of the hidden layer is the same as the visible layer. For the first group datasets (MSRA-MM), the Micro-DL architecture consists of one Micro-DGRBM and sixteen Micro-DRBMs. To compare fairness, the learning rates of our Micro-DL architecture are the same as NMicro-DL model for training MSRA-MM datasets. Both of them are $10^{-4}$. For the second group datasets (UCI), the Micro-DL architecture consists of six hidden layers. The learning rates of the weights, biases of visible and hidden layers of Micro-DGRBM and Micro-DRBM models are set to $10^{-8}$. In our Micro-DL architecture, the scale coefficient of Micro-DGRBM and Micro-DRBM are set to 0.3 for all twelve datasets. The number of hidden layer units of Micro-DL is 892 on the aquarium, bathroom, blog, blood, bouquet, bugat and cactus datasets. They are 899, 6, 41, 19 and 13 on the voituretuning, car, KDD99, segmentation and vowe datasets, respectively.
\subsection{Evaluation Metrics}
Four popular external evaluation metrics that are used in the paper to assess the experimental results are the clustering accuracy \cite{6420834}, Jaccard index (Jac index) \cite{DEFRANCA201624}, Fowlkes and Mallows index (FM index) \cite{LIU2018200} and Rand index \cite{Rand}. Furthermore, the Friedman aligned ranks test \cite{garcia2010advanced}) is used to provide fair comparisons among different methods. The calculations of four external metrics are provided as follows:
\begin{enumerate}
\item The clustering accuracy metric is used to calculate the ratio of
the instance assigned to the correct clusters. It is defined to be
\begin{equation}
\begin{aligned}
    accuracy=\frac{\sum\limits_{i=1}^{N}F(l_{i},l_{i}^{'})}{N},
\end{aligned}
\end{equation}
where $N$ is the number of instances, $l_{i}$ and $l_{i}^{'}$ are the target and predicted label of the $i$th instance, respectively. If $l_{i}=l_{i}^{'}$, then $F(l_{i},l_{i}^{'})=1$. Otherwise, $F(l_{i},l_{i}^{'})=0$.
  \item The Jaccard Index measures similarity between sample sets can be written as
  \begin{equation}
\begin{aligned}
    Jac=\frac{|A\cap B|}{|A\cup B|},
\end{aligned}
\end{equation}
where $A$ and $B$ are finite sample sets.
  \item The Fowlkes and Mallows index calculates the similarity between the benchmark classifications and the clusters returned by the clustering algorithm. It is defined as
  \begin{equation}
\begin{aligned}
  FMI=\sqrt{\frac{TP}{TP+FP}\times\frac{TP}{TP+FN}},
   \end{aligned}
\end{equation}
where $TP$, $FP$ and $FN$ are the numbers of true positives, false positives and false negatives, respectively.
 \item The Rand index is a measure of the percentage of correct decisions made by the algorithm. It is defined to be
 \begin{equation}
\begin{aligned}
  Rand=\frac{TP+TN}{TP+FP+FN+TN},
   \end{aligned}
\end{equation}
where $TN$ is the number of true negatives.
\end{enumerate}
\indent The Friedman aligned ranks test is an advanced and popular nonparametric test method which can be used to analyze the performance of algorithms. It can be written as
\begin{equation}
\begin{aligned}
  T=\frac{(n-1)(\sum\limits_{j=1}^{n}\widehat{r}_{.j}^2-nm^2(nm+1)^2/4)}{nm(nm+1)(2nm+1)/6-\sum\limits_{i=1}^{m}\widehat{r}_{i.}^2/n},
\end{aligned}
\end{equation}
 where $\widehat{r}_{.j}$ is the total ranks of the $i$th data set, $\widehat{r}_{i.}$ is the total ranks of the  $j$th algorithm, $n$ is the number of algorithm and $m$ is the number of data set. The test statistic $T$ is compared for significance with a chi-square distribution for $n-1$ degrees of freedom.
\subsection{Clustering Performance}
We first compare the proposed Micro-DL architecture with benchmarking algorithm (Semi-SP). To prove deep representation learning capability of our Micro-DL, we compare it with the most related shallow semi-supervised feature learning models (pcGRBM\cite{Chujielei2018pcGRBM} and Semi-EAGR\cite{7420705}). Then, two unsupervised deep representation models (VGAE and NMicro-DL) are compared with our Micro-DL model to show the effectiveness of the proposed Micro-supervised strategy. Significantly, the NMicro-DL is the most related deep learning model with our Micro-DL architecture. Hence, we compare the performance of them to evaluate that the distributions of deep hidden features of our Micro-DL are whether or not more reasonable than the NMicro-DL model. Finally, we compare further the proposed Micro-DL architecture with deep semi-supervised model (Semi-MG\cite{mercadoNeurIPS2019}) to evaluate impartially the capability of deep representation learning.\\
 \indent All results of clustering accuracy are listed in Table 1. On the whole, the average clustering accuracies of Semi-SP, pcGRBM, Semi-EAGR, Semi-MG, VGAE and NMicro-DL are 0.3724, 0.4335, 0.4656, 0.6692, 0.5789, 0.5619, respectively. However, the clustering accuracy of our Micro-DL architecture increases to 0.7343. It's obvious that the performance of the Micro-DL is significantly increased by 0.3619 when it is compared with the Semi-SP algorithm. These results demonstrates that its output features have reasonable distributions than visible layer data for clustering. In contrast to the pcGRBM and Semi-EAGR models, the performances of the proposed Micro-DL framework are increased by 0.3018 and 0.2687, respectively. Thus, the results indicate that the capability of deep representation learning of our Micro-DL is more powerful than shallow representation learning of the most related contrastive models. When the proposed Micro-DL architecture is compared with the Semi-MG, VGAE and NMicro-DL, the performances are increased by 0.0651, 0.1554 and 0.1724, respectively. Hence, we can conclude that the distributions of hidden layers of our Micro-DL are more reasonable under the stimulation of the SPI than all contrast deep models.\\
 \indent As shown in Table 2, the average Jac of Semi-SP, pcGRBM, Semi-EAGR, Semi-MG, VGAE and NMicro-DL are 0.2521, 0.3123, 0.3224, 0.5073, 0.4148 and 0.4015, respectively. However, our Micro-DL architecture raises the Jac metric to 0.5836 significantly. In contrast to the Semi-SP algorithm, it improves the performance by 0.3315. These results demonstrates again that the hidden features of the Micro-DL architecture have reasonable distributions than visible layer data for clustering. The Jac indices of Micro-DL are increased by 0.2713 and 0.2612 when it is compared with the pcGRBM and Semi-EAGR models, respectively. Furthermore, in contrast to Semi-MG, VGAE and NMicro-DL frameworks, our Micro-DL improves the Jac by 0.0763, 0.1688 and 0.1821, respectively. Hence, these results illustrate that the proposed Micro-DL has more outstanding capability of deep representation learning under the stimulation of the SPI than contrast shallow models and deep frameworks.\\
 \indent In Table 3, the average FMs of Semi-SP, pcGRBM, Semi-EAGR, Semi-MG, VGAE and NMicro-DL  are 0.4077, 0.4710, 0.4935, 0.6807, 0.5913 and 0.5665, respectively. However, the average FM index of our Micro-DL architecture is raised to 0.7564 significantly. In contrast to the Semi-SP algorithm, it improves the performance by 0.3487. When the proposed Micro-DL is compared with the pcGRBM and Semi-EAGR models, the FM indices are increased by 0.2854 and 0.2629, respectively. Furthermore, in contrast to Semi-MG, VGAE and NMicro-DL frameworks, our Micro-DL architecture improves the FM by 0.0757, 0.1651 and 0.1899, respectively. Hence, these comparisons show that our Micro-DL has exciting capability of deep representation learning for clustering.\\
\indent The results of Rand index are presented in Table 4. The average Rand of Semi-SP, pcGRBM, Semi-EAGR, Semi-MG, VGAE and NMicro-DL are 0.5085, 0.5508, 0.5285, 0.5639, 0.5264 and 0.5408, respectively. However, the proposed Micro-DL raises the average Rand to 0.5988. In contrast to the Semi-SP algorithms, it improves the metric of average Rand by 0.0903. The average Rand indices of our Micro-DL are increased by 0.0480 and 0.0703 when it is compared with the pcGRBM and Semi-EAGR models, respectively. Furthermore, in contrast to Semi-MG, VGAE and NMicro-DL frameworks, our Micro-DL improves the metric of average Rand by 0.0349, 0.0724 and 0.0580, respectively.\\
\indent Fig. 2 presents visual contrast of the clustering performances (accuracy, Jac, FM and Rand) among the benchmarking algorithms, shallow models and deep frameworks on twelve datasets. On the whole, we can see that our Micro-DL architecture shows fairly competitive performances in all evaluation metrics.
\begin{figure*}[!htbp]
\vspace{0.5mm} \centering
   \includegraphics[scale=0.4025]{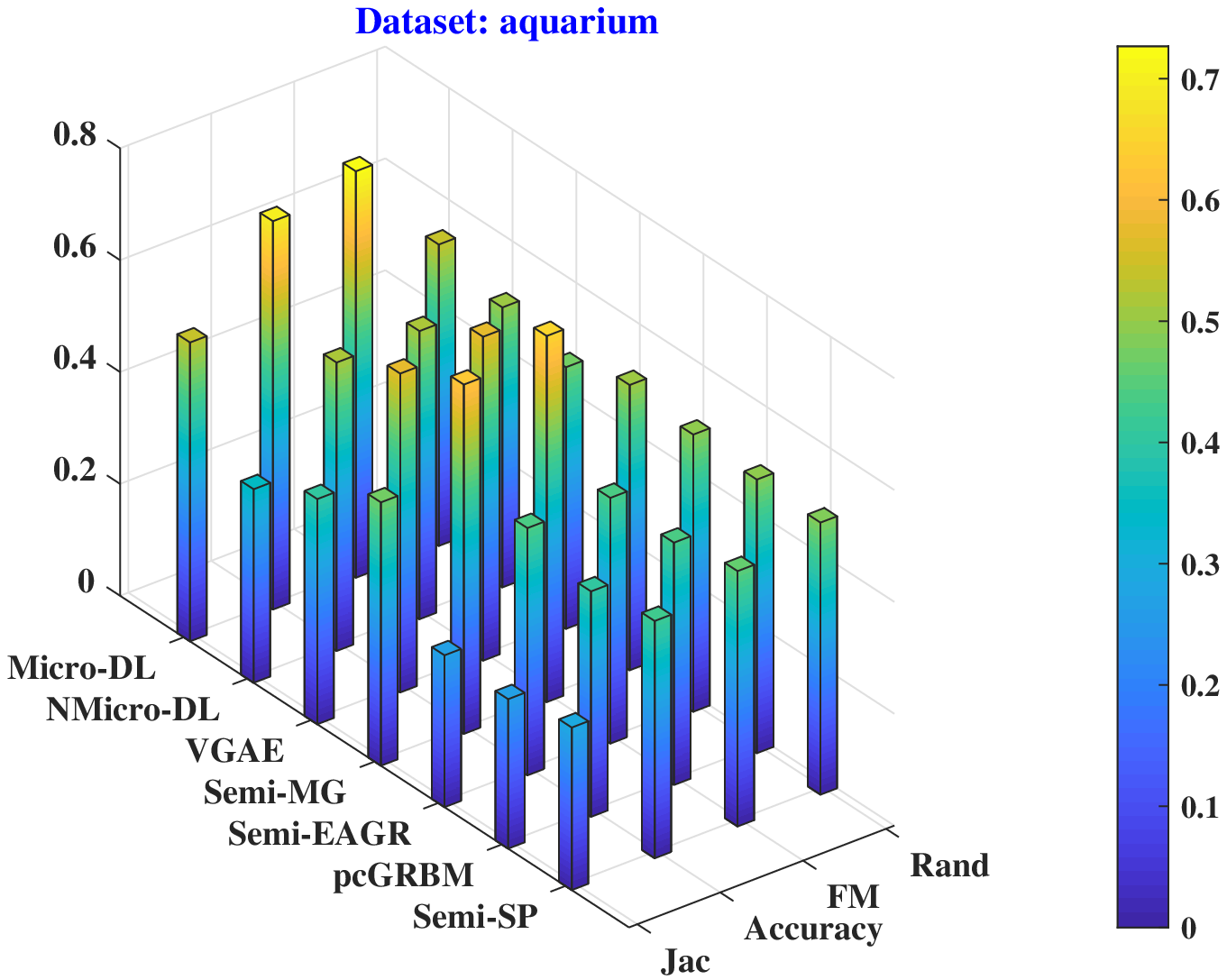}
   \includegraphics[scale=0.4025]{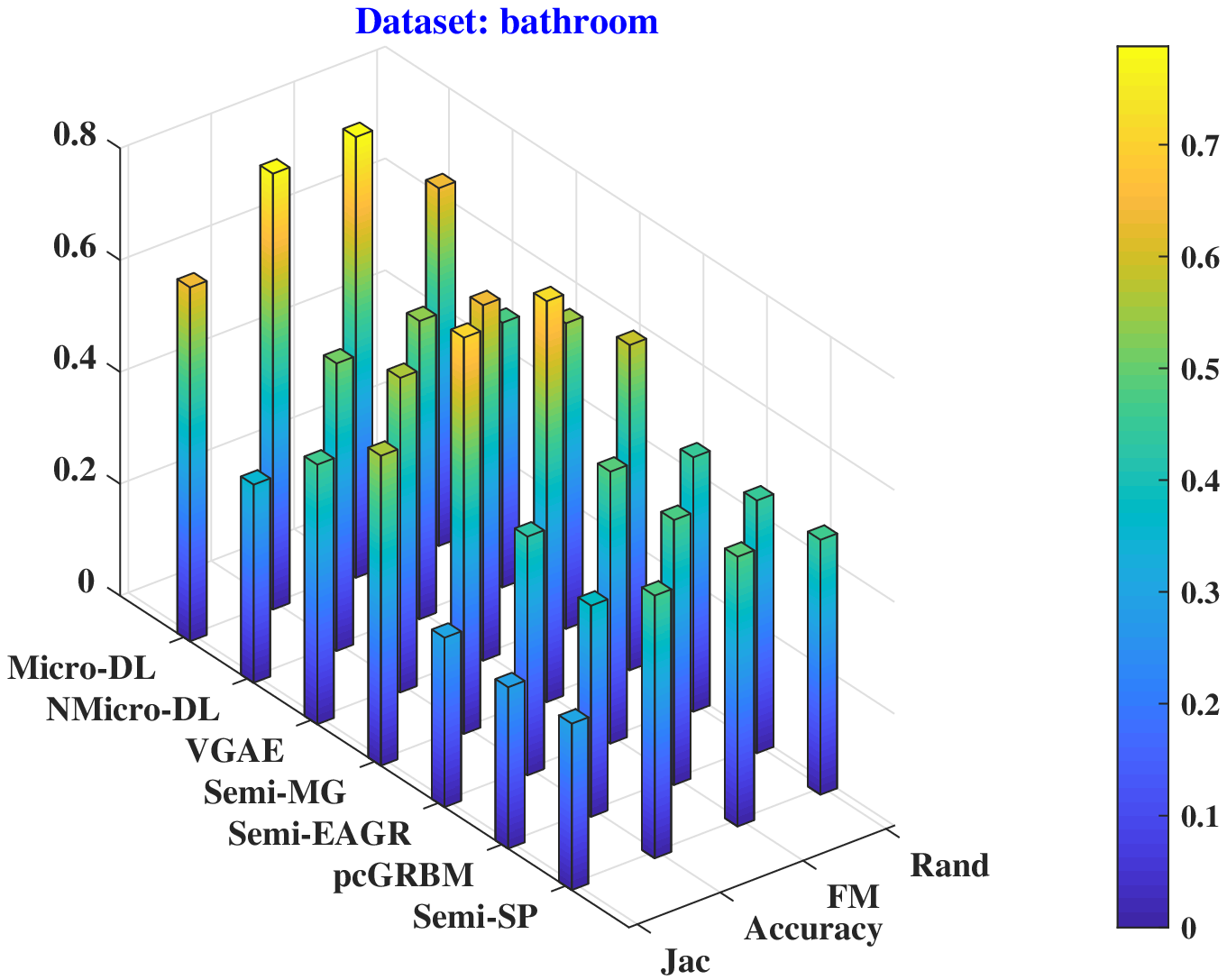}
   \includegraphics[scale=0.4025]{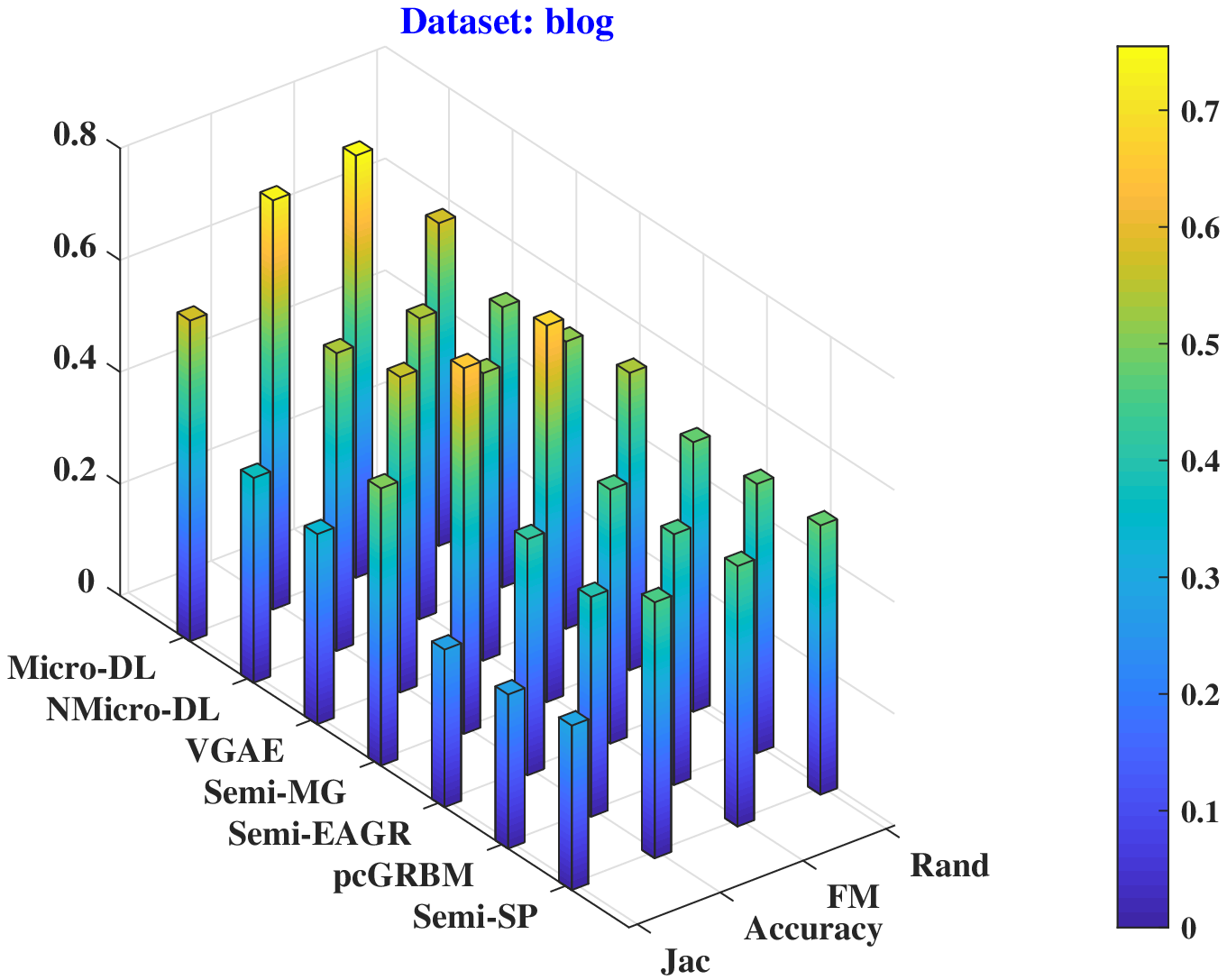}
    \includegraphics[scale=0.4025]{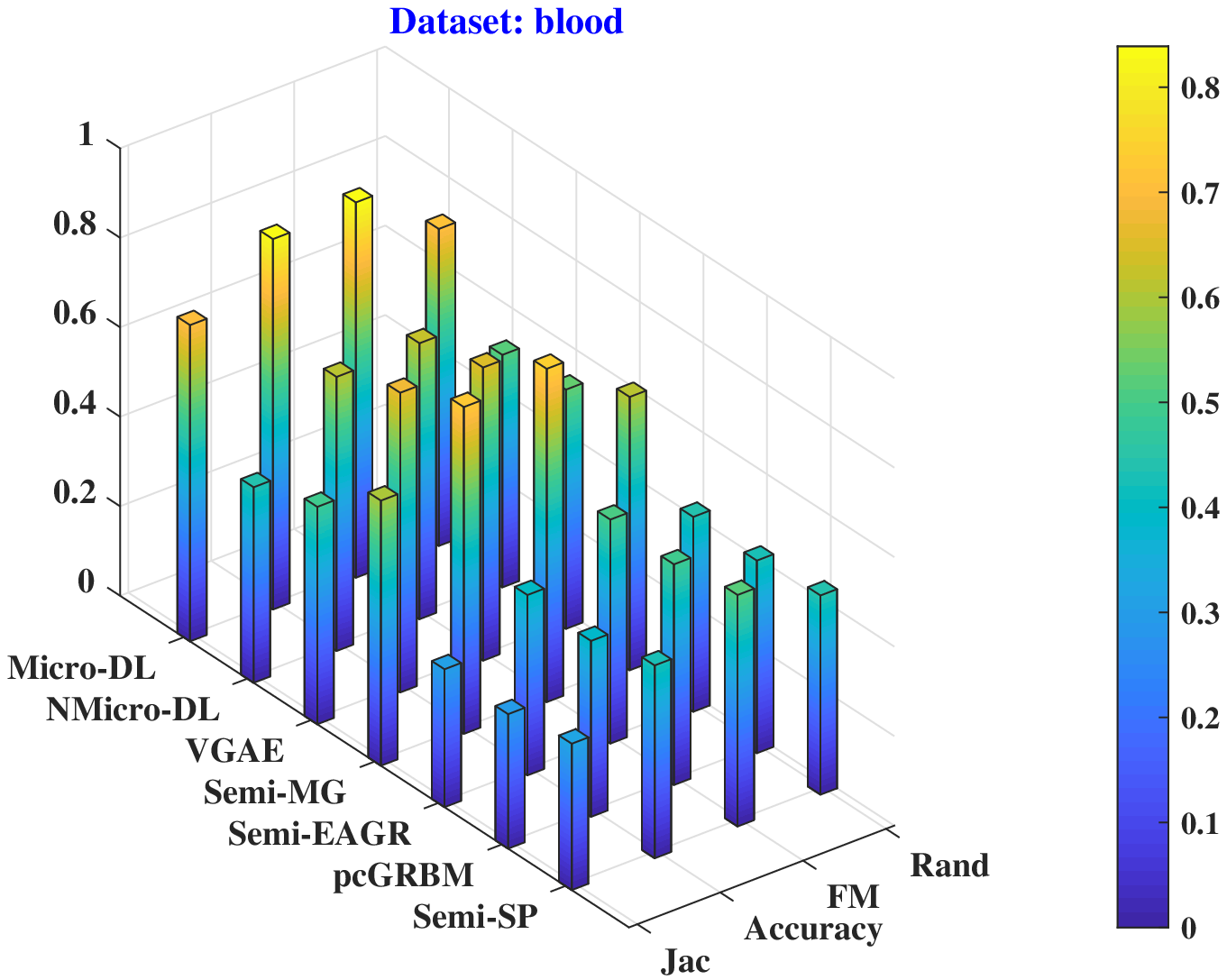}
   \includegraphics[scale=0.4025]{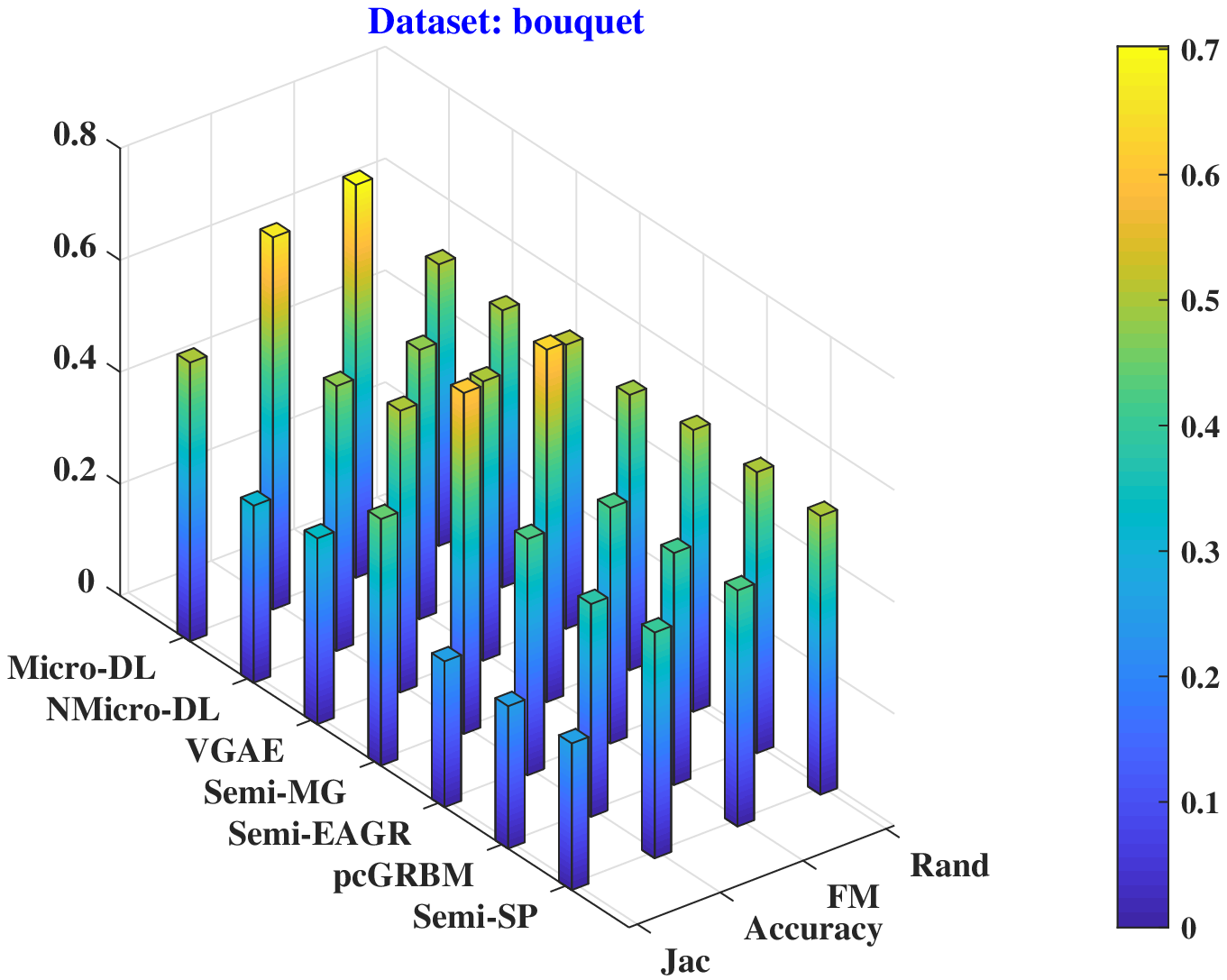}
    \includegraphics[scale=0.4025]{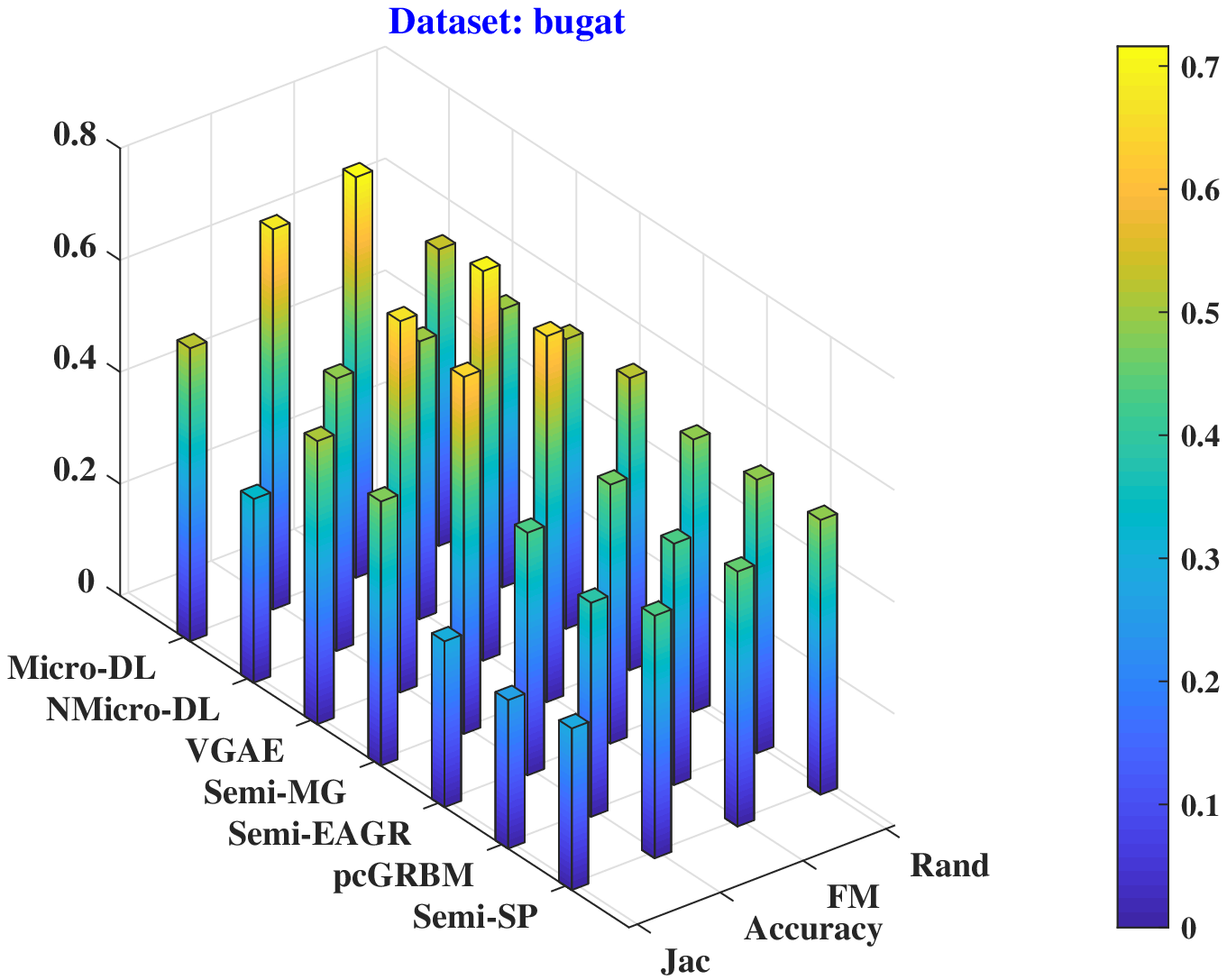}
   \includegraphics[scale=0.4025]{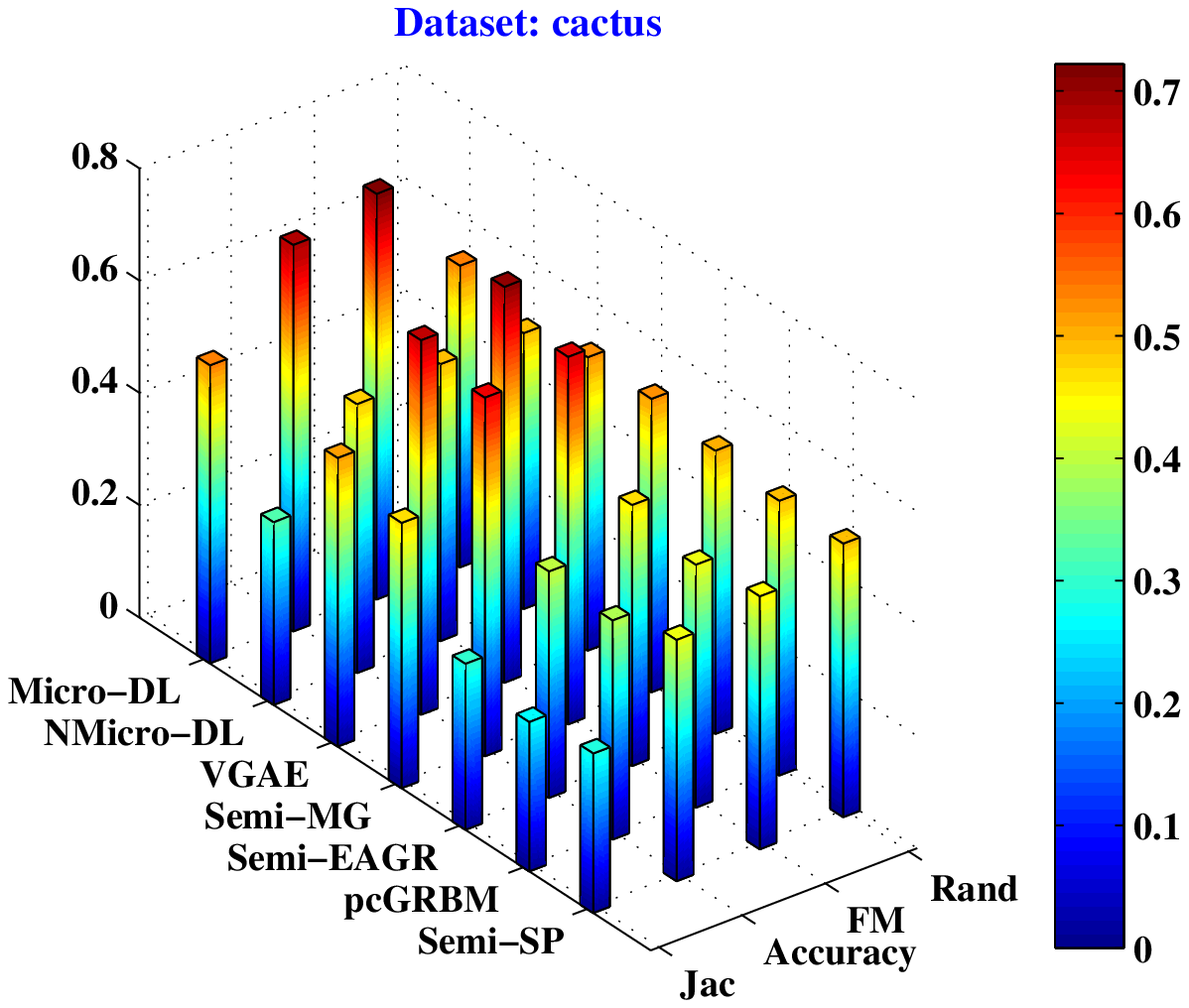}
    \includegraphics[scale=0.4025]{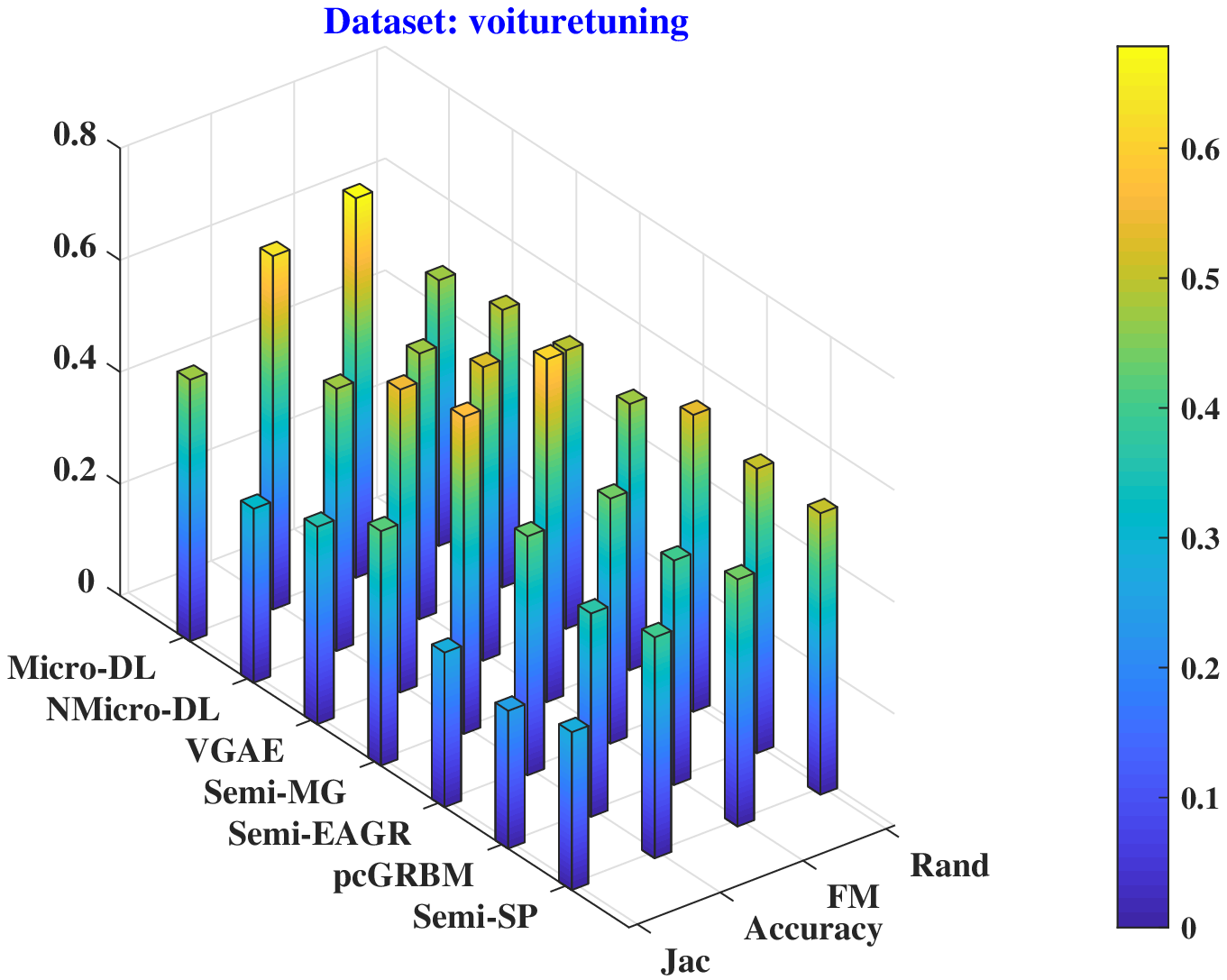}
  \includegraphics[scale=0.4025]{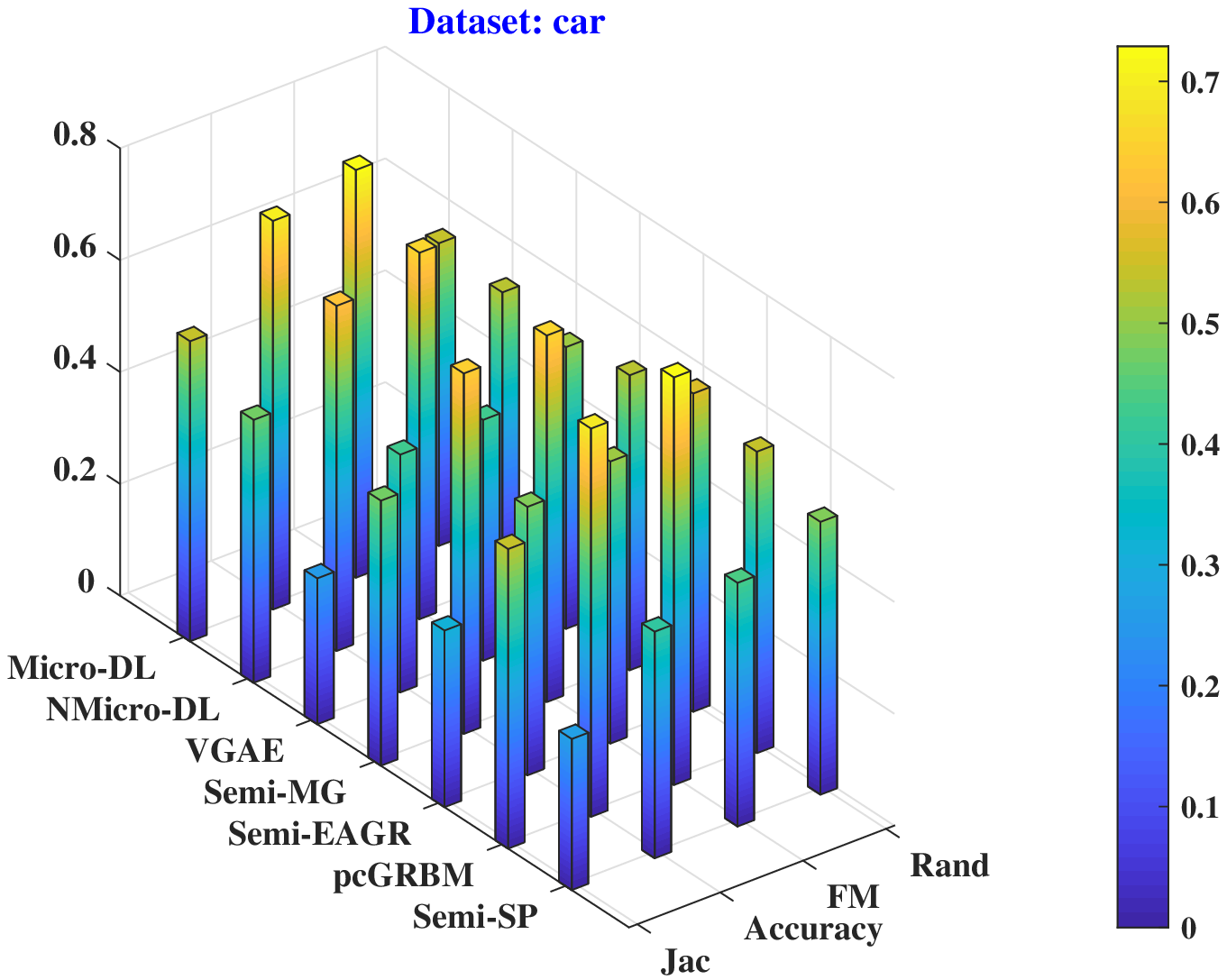}
    \includegraphics[scale=0.4025]{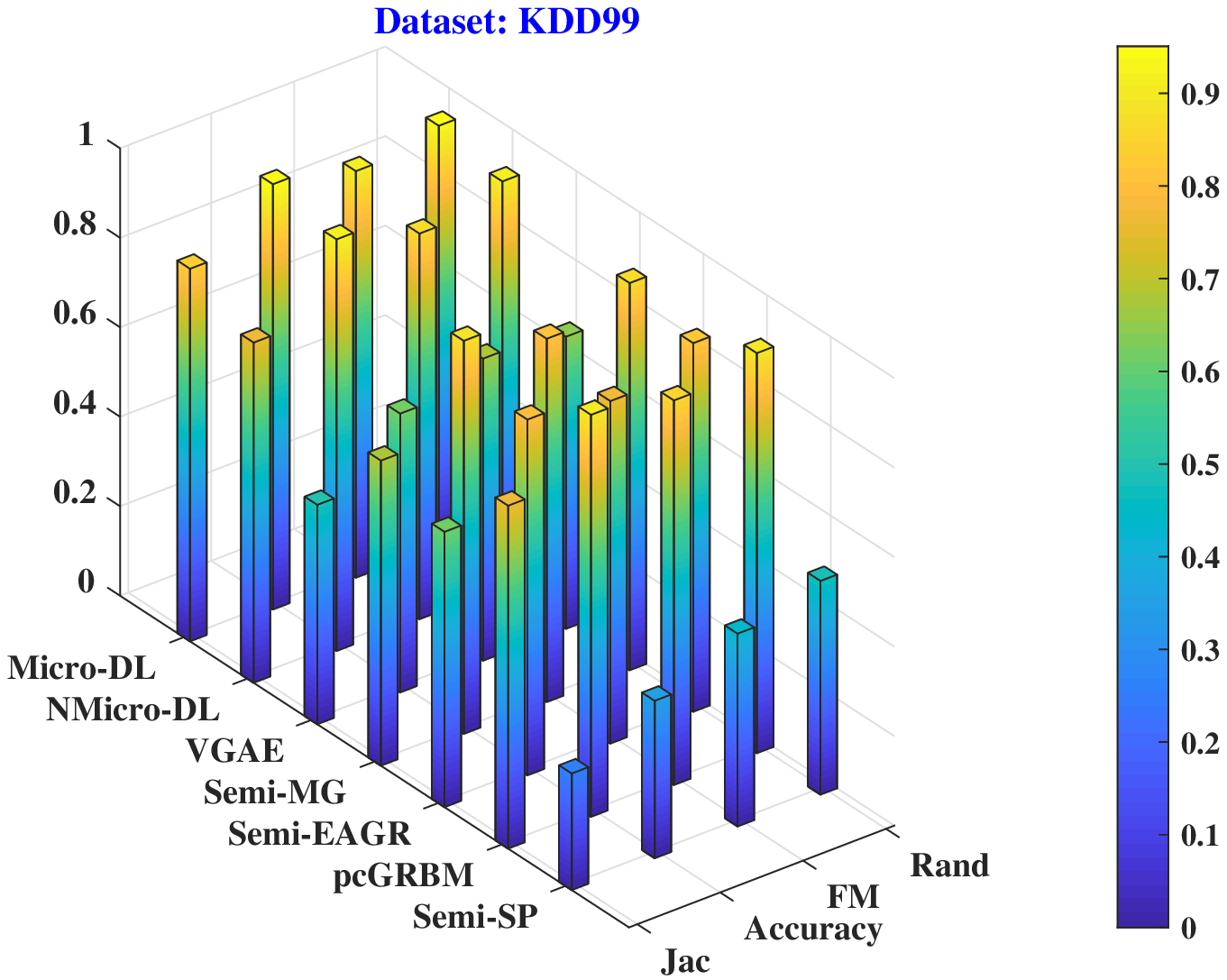}
    \includegraphics[scale=0.4025]{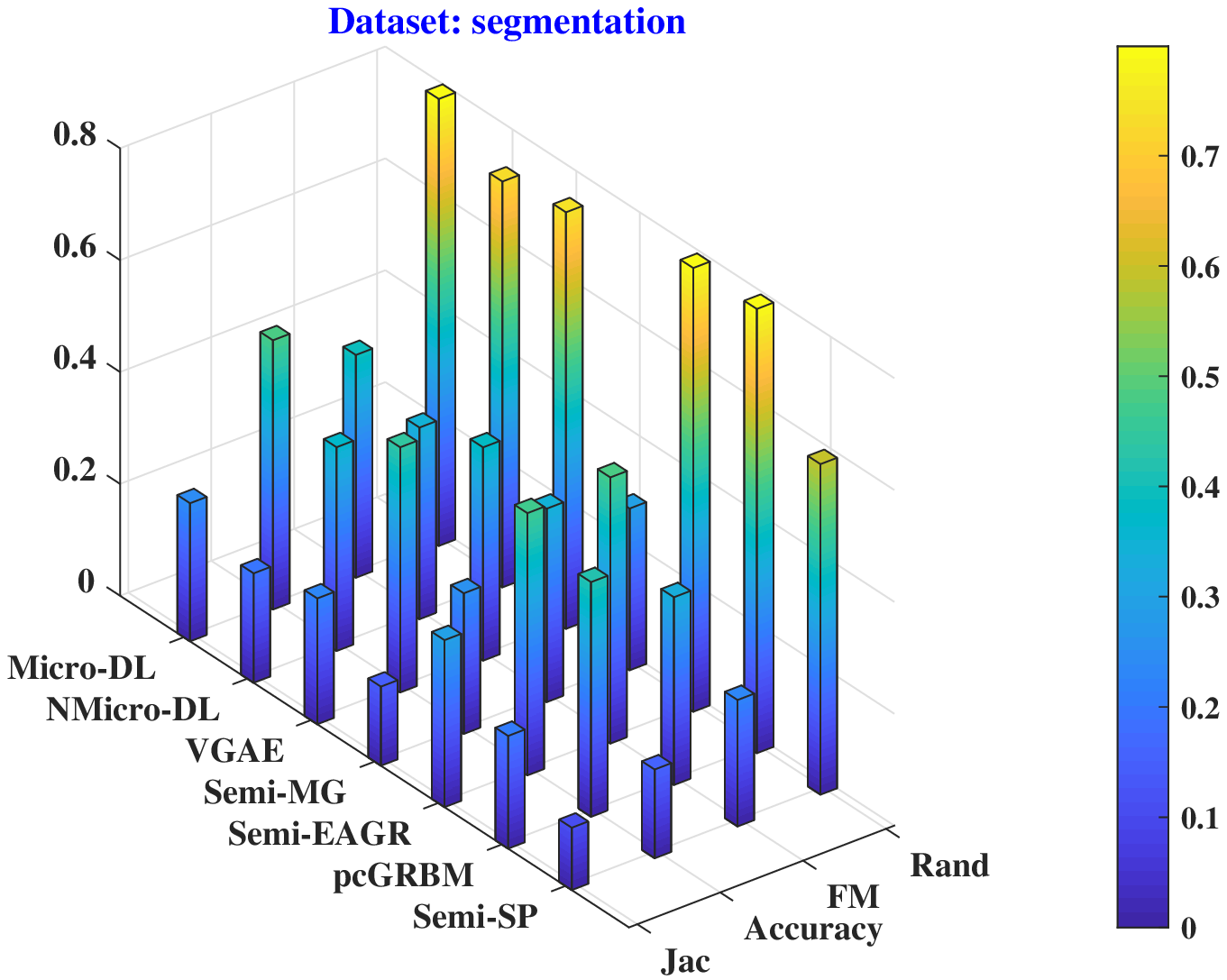}
       \includegraphics[scale=0.4025]{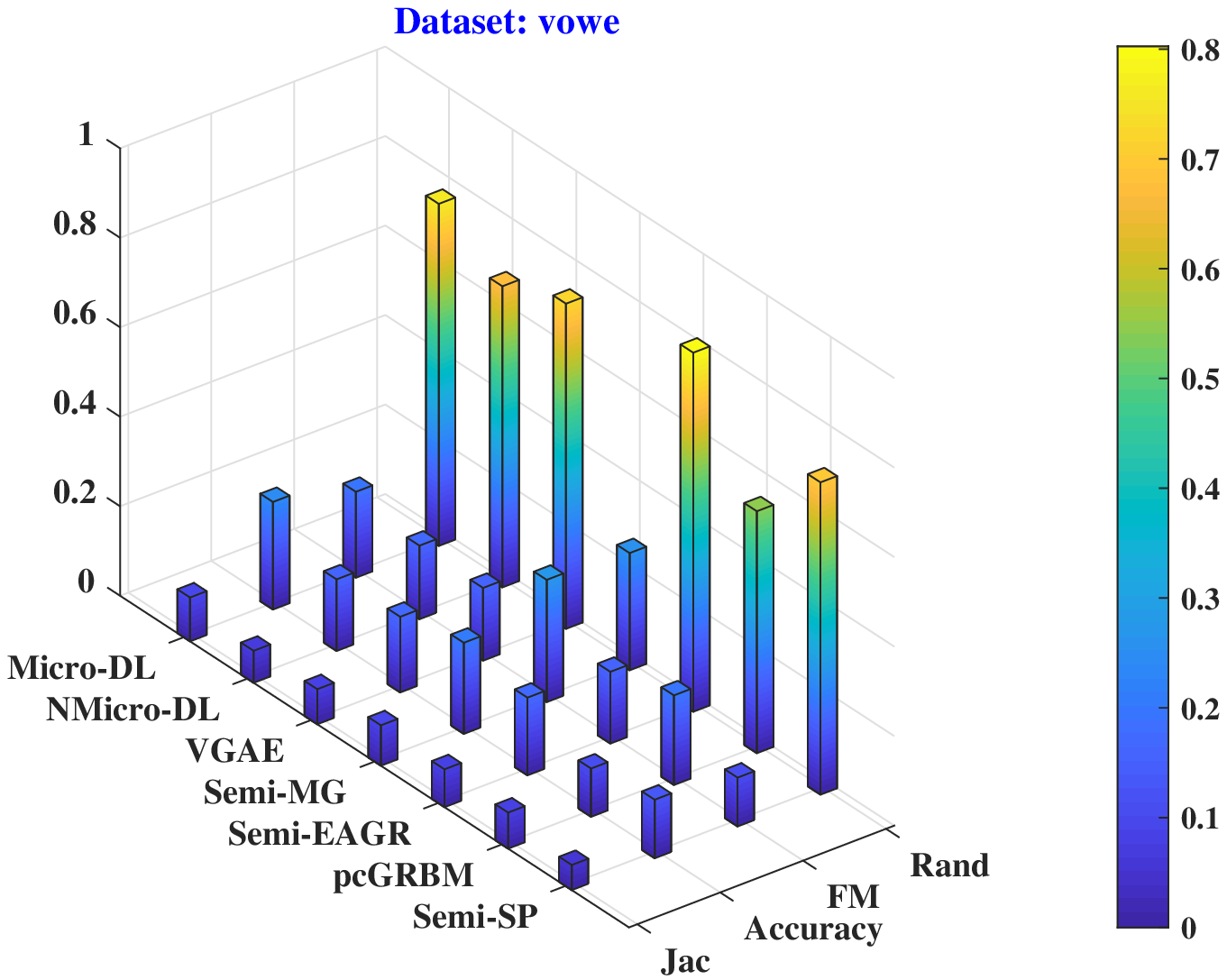}
 \caption{Performance comparisions (Accuracy, Jac index, FM index and Rand index) of benchmarking algorithms (Semi-SP), shallow models (pcGRBM and Semi-EAGR), and deep models (Semi-MG, VGAE, NMicro-DL and Micro-DL). }\label{fig:1}
\end{figure*}
\begin{table*}
\begin{center}
\caption{Performance comparisions (Accuracy) of benchmarking algorithms (Semi-SP), shallow models (pcGRBM and Semi-EAGR), and deep models (Semi-MG, VGAE, NMicro-DL and Micro-DL). The larger clustering accuracy, the better performance. The best performance on each data set is bolded. }
\label{tab:results1} \scalebox{0.9}{
\begin{tabular}{|l|c|c|c|c|c|c|c|}
\hline
\textsf{ \bf{	Dataset	}} & {	Semi-SP			} & {	pcGRBM			} & {	Semi-EAGR			} & {	Semi-MG			} & {	VGAE				} & {	NMicro-DL				} & {	\textbf{Micro-DL}			} \\
\hline \textsf{	aquarium	} & {	0.4242 	$\pm$	0.0489 	} & {	0.4030 	$\pm$	0.0104 	} & {	0.4421 	$\pm$	0.0178 	} & {	0.6245	$\pm$	0.0556 	} & {	0.5701 	$\pm$	0.0455 		} & {	0.5157 	$\pm$	0.0328 		} & {	\textbf{0.6943 	$\pm$	0.0047 }	} \\
\hline \textsf{	bathroom	} & {	0.4695 	$\pm$	0.0610 	} & {	0.3774 	$\pm$	0.0106 	} & {	0.4268 	$\pm$	0.0291 	} & {	0.7083	$\pm$	0.0651 	} & {	0.5624 	$\pm$	0.0579 		} & {	0.5141 	$\pm$	0.0293 		} & {	\textbf{0.7792 	$\pm$	0.0053 }	} \\
\hline \textsf{	blog	} & {	0.4575 	$\pm$	0.0319 	} & {	0.3930 	$\pm$	0.0092 	} & {	0.4221 	$\pm$	0.0197 	} & {	0.6538	$\pm$	0.0588 	} & {	0.5637 	$\pm$	0.0739 		} & {	0.5323 	$\pm$	0.0491 		} & {	\textbf{0.7314 	$\pm$	0.0048 }	} \\
\hline \textsf{	blood	} & {	0.4309 	$\pm$	0.0580 	} & {	0.3931 	$\pm$	0.0111 	} & {	0.4030 	$\pm$	0.0157 	} & {	0.7304	$\pm$	0.0738 	} & {	0.6697 	$\pm$	0.0358 		} & {	0.6125 	$\pm$	0.0158 		} & {	\textbf{0.8279 	$\pm$	0.0057} 	} \\
\hline \textsf{	bouquet	} & {	0.4035 	$\pm$	0.0439 	} & {	0.3803 	$\pm$	0.0133 	} & {	0.4224 	$\pm$	0.0231 	} & {	0.6093	$\pm$	0.0439 	} & {	0.5034 	$\pm$	0.0658 		} & {	0.4735 	$\pm$	0.0328 		} & {	\textbf{0.6653 	$\pm$	0.0045 }	} \\
\hline \textsf{	bugat	} & {	0.4334 	$\pm$	0.0378 	} & {	0.3825 	$\pm$	0.0138 	} & {	0.4336 	$\pm$	0.0201 	} & {	0.6381	$\pm$	0.0428 	} & {	0.6636 	$\pm$	0.0055 		} & {	0.4870 	$\pm$	0.0457 		} & {\textbf{	0.6795 	$\pm$	0.0045 }	} \\
\hline \textsf{	cactus	} & {	0.4294 	$\pm$	0.0440 	} & {	0.3898 	$\pm$	0.0127 	} & {	0.4033 	$\pm$	0.0181 	} & {	0.6381	$\pm$	0.0428 	} & {	0.6664 	$\pm$	0.0047 		} & {	0.4785 	$\pm$	0.0353 		} & {	\textbf{0.6878 	$\pm$	0.0046 }	} \\
\hline \textsf{	voituretuning	} & {	0.3949 	$\pm$	0.0390 	} & {	0.3634 	$\pm$	0.0154 	} & {	0.4272 	$\pm$	0.0312 	} & {	0.5668	$\pm$	0.0383 	} & {	0.5414 	$\pm$	0.0314 		} & {	0.4686 	$\pm$	0.0310 		} & {	\textbf{0.6318 	$\pm$	0.0042 }	} \\
\hline \textsf{	car	} & {	0.4051 	$\pm$	0.0641 	} & {	0.6941 	$\pm$	0.0143 	} & {	0.4800 	$\pm$	0.0331 	} & {	0.6444	$\pm$	0.0398 	} & {	0.4251 	$\pm$	0.0551 		} & {	0.6170 	$\pm$	0.0145 		} & {	\textbf{0.6948 	$\pm$	0.0024 }	} \\
\hline \textsf{	KDD99	} & {	0.3519 	$\pm$	0.0077 	} & {	0.8983 	$\pm$	0.1049 	} & {	0.7950 	$\pm$	0.0941 	} & {	0.8784	$\pm$	0.0190 	} & {	0.6231 	$\pm$	0.0222 		} & {	0.9199 	$\pm$	0.0422 		} & {	\textbf{0.9505 	$\pm$	0.0001 }	} \\
\hline \textsf{	segmentation	} & {	0.1587 	$\pm$	0.0041 	} & {	0.4196 	$\pm$	0.0520 	} & {	0.4688 	$\pm$	0.0579 	} & {	0.2514	$\pm$	0.0779 	} & {	0.4383 	$\pm$	0.0756 		} & {	0.3643 	$\pm$	0.0111 		} & {	\textbf{0.4814 	$\pm$	0.0101} 	} \\
\hline \textsf{	vowe	} & {	0.1309 	$\pm$	0.0042 	} & {	0.1079 	$\pm$	0.0073 	} & {	0.1737 	$\pm$	0.0110 	} & {	0.2037	$\pm$	0.0817 	} & {	0.1691 	$\pm$	0.0147 		} & {	0.1601 	$\pm$	0.0179 		} & {	\textbf{0.2404 	$\pm$	0.0286 }	} \\																															
\hline {\textcolor[rgb]{0.00,0.25,0.50}{\textbf{Average}}}		 & {\textcolor[rgb]{0.00,0.25,0.50}{	0.3742} 			} & {	\textcolor[rgb]{0.00,0.25,0.50}{0.4335 }			} & {	\textcolor[rgb]{0.00,0.25,0.50}{0.4656} 			} & {\textcolor[rgb]{0.00,0.25,0.50}{	0.6692} 			} & {	\textcolor[rgb]{0.00,0.25,0.50}{0.5789 }				} & {	\textcolor[rgb]{0.00,0.25,0.50}{0.5619 	}			} & {	\textbf{\textcolor[rgb]{0.00,0.25,0.50}{0.7343} 	}		} \\

\hline	
\end{tabular}}
\end{center}
\end{table*}
\begin{table*}
\begin{center}
\caption{Performance comparisions (Jac index) of benchmarking algorithms (Semi-SP), shallow models (pcGRBM and Semi-EAGR), and deep models (Semi-MG, VGAE, NMicro-DL and Micro-DL). The larger Jac, the better performance. The best performance on each data set is bolded.}
\label{tab:results1} \scalebox{0.9}{
\begin{tabular}{|l|c|c|c|c|c|c|c|c|}
\hline
\textsf{ \bf{	Dataset	}} & {	Semi-SP			} & {	pcGRBM			} & {	Semi-EAGR			} & {	Semi-MG			} & {	VGAE				} & {	NMicro-DL				} & {	\textbf{Micro-DL}			} \\
\hline \textsf{	aquarium	} & {	0.2906 	$\pm$	0.0047 	} & {	0.2672 	$\pm$	0.0015 	} & {	0.2715 	$\pm$	0.0041 	} & {	0.4708 	$\pm$	0.0477 	} & {	0.4026 	$\pm$	0.0032 		} & {	0.3461 	$\pm$	0.0216 		} & {	\textbf{0.5343 	$\pm$	0.0024 }	} \\
\hline \textsf{	bathroom	} & {	0.2975 	$\pm$	0.0035 	} & {	0.2885 	$\pm$	0.0016 	} & {	0.3029 	$\pm$	0.0216 	} & {	0.5540 	$\pm$	0.0692 	} & {	0.4641 	$\pm$	0.0520 		} & {	0.3542 	$\pm$	0.0215 		} & {	\textbf{0.6327 	$\pm$	0.0190} 	} \\
\hline \textsf{	blog	} & {	0.2940 	$\pm$	0.0055 	} & {	0.2755 	$\pm$	0.0016 	} & {	0.2816 	$\pm$	0.0112 	} & {	0.4957 	$\pm$	0.0547 	} & {	0.3398 	$\pm$	0.0686 		} & {	0.3662 	$\pm$	0.0342 		} & {	\textbf{0.5733 	$\pm$	0.0028 }	} \\
\hline \textsf{	blood	} & {	0.3271 	$\pm$	0.0314 	} & {	0.3005 	$\pm$	0.0019 	} & {	0.3082 	$\pm$	0.0161 	} & {	0.5924 	$\pm$	0.0805 	} & {	0.4858 	$\pm$	0.0709 		} & {	0.4370 	$\pm$	0.0140 		} & {\textbf{	0.7067 	$\pm$	0.0051} 	} \\
\hline \textsf{	bouquet	} & {	0.2622 	$\pm$	0.0100 	} & {	0.2546 	$\pm$	0.0022 	} & {	0.2606 	$\pm$	0.0065 	} & {	0.4408 	$\pm$	0.0421 	} & {	0.3322 	$\pm$	0.0757 		} & {	0.3166 	$\pm$	0.0173 		} & {	\textbf{0.4983 	$\pm$	0.0017 }	} \\
\hline \textsf{	bugat	} & {	0.2889 	$\pm$	0.0145 	} & {	0.2654 	$\pm$	0.0022 	} & {	0.2964 	$\pm$	0.0240 	} & {	0.4722 	$\pm$	0.0474 	} & {	0.5060 	$\pm$	0.0789 		} & {	0.3285 	$\pm$	0.0286 		} & {	\textbf{0.5239 	$\pm$	0.0032 }	} \\
\hline \textsf{	cactus	} & {	0.2842 	$\pm$	0.0162 	} & {	0.2660 	$\pm$	0.0026 	} & {	0.2954 	$\pm$	0.0309 	} & {	0.4722 	$\pm$	0.0474 	} & {	0.5138 	$\pm$	0.0799 		} & {	0.3253 	$\pm$	0.0209 		} & {	\textbf{0.5306 	$\pm$	0.0056} 	} \\
\hline \textsf{	voituretuning	} & {	0.2822 	$\pm$	0.0136 	} & {	0.2463 	$\pm$	0.0026 	} & {	0.2763 	$\pm$	0.0225 	} & {	0.4192 	$\pm$	0.0374 	} & {	0.3529 	$\pm$	0.0838 		} & {	0.3111 	$\pm$	0.0186 		} & {	\textbf{0.4676 	$\pm$	0.0032 }	} \\
\hline \textsf{	car	} & {	0.2701 	$\pm$	0.0428 	} & {	0.5356 	$\pm$	0.0160 	} & {	0.3158 	$\pm$	0.0227 	} & {	0.4738 	$\pm$	0.0455 	} & {	0.2608 	$\pm$	0.0923 		} & {	0.4701 	$\pm$	0.1197 		} & {	\textbf{0.5365 	$\pm$	0.0026} 	} \\
\hline \textsf{	KDD99	} & {	0.2605 	$\pm$	0.0382 	} & {	0.7663 	$\pm$	0.1737 	} & {	0.6153 	$\pm$	0.0007 	} & {	0.6818 	$\pm$	0.0288 	} & {	0.4903 	$\pm$	0.0991 		} & {	0.7603 	$\pm$	0.1076 		} & {	\textbf{0.8326 	$\pm$	0.0008 }	} \\
\hline \textsf{	segmentation	} & {	0.1115 	$\pm$	0.0123 	} & {	0.2015 	$\pm$	0.0308 	} & {	\textbf{0.2986 	$\pm$	0.0023 }	} & {	0.1414 	$\pm$	0.0014 	} & {	0.2250 	$\pm$	0.0462 		} & {	0.1957 	$\pm$	0.0231 		} & {	0.2469 	$\pm$	0.0569 	} \\
\hline \textsf{	vowe	} & {	0.0561 	$\pm$	0.0051 	} & {	0.0806 	$\pm$	0.0023 	} & {	0.0846 	$\pm$	0.0087 	} & {	0.0903 	$\pm$	0.0014 	} & {	0.0784 	$\pm$	0.0076 		} & {	0.0715 	$\pm$	0.0248 		} & {	\textbf{0.0980 	$\pm$	0.0093} 	} \\																															
\hline {\textbf{\textcolor[rgb]{0.00,0.25,0.50}{Average}}}		 & {	\textcolor[rgb]{0.00,0.25,0.50}{0.2521 }			} & {	\textcolor[rgb]{0.00,0.25,0.50}{0.3123} 			} & {\textcolor[rgb]{0.00,0.25,0.50}{	\textcolor[rgb]{0.00,0.25,0.50}{0.3224} 	}		} & {	\textcolor[rgb]{0.00,0.25,0.50}{0.5073} 			} & {	\textcolor[rgb]{0.00,0.25,0.50}{0.4148 }				} & {	\textcolor[rgb]{0.00,0.25,0.50}{0.4015 }				} & {	\textbf{\textcolor[rgb]{0.00,0.25,0.50}{0.5836} 	}		} \\

\hline
\end{tabular}}
\end{center}
\end{table*}
\begin{table*}
\begin{center}
\caption{Performance comparisions (FM index) of benchmarking algorithms (Semi-SP), shallow models (pcGRBM and Semi-EAGR), and deep models (Semi-MG, VGAE, NMicro-DL and Micro-DL). The larger FM, the better performance. The best performance on each data set is bolded.}
\label{tab:results1} \scalebox{0.9}{
\begin{tabular}{|l|c|c|c|c|c|c|c|c|}
\hline
\textsf{ \bf{	Dataset	}} & {	Semi-SP			} & {	pcGRBM			} & {	Semi-EAGR			} & {	Semi-MG			} & {	VGAE				} & {	NMicro-DL				} & {	\textbf{Micro-DL}			} \\
\hline \textsf{	aquarium	} & {	0.4565 	$\pm$	0.0047 	} & {	0.4333 	$\pm$	0.0019 	} & {	0.4392 	$\pm$	0.0052 	} & {	0.6545 	$\pm$	0.0545 	} & {	0.5790 	$\pm$	0.0035 		} & {	0.5151 	$\pm$	0.0227 		} & {	\textbf{0.7266 	$\pm$	0.0026 	}} \\
\hline \textsf{	bathroom	} & {	0.4822 	$\pm$	0.0028 	} & {	0.4737 	$\pm$	0.0020 	} & {	0.4860 	$\pm$	0.0204 	} & {	0.7166 	$\pm$	0.0632 	} & {	0.6353 	$\pm$	0.0466 		} & {	0.5329 	$\pm$	0.0200 		} & {	\textbf{0.7878 	$\pm$	0.0174 }	} \\
\hline \textsf{	blog	} & {	0.4654 	$\pm$	0.0063 	} & {	0.4481 	$\pm$	0.0019 	} & {	0.4538 	$\pm$	0.0127 	} & {	0.6731 	$\pm$	0.0581 	} & {	0.5137 	$\pm$	0.0663 		} & {	0.5376 	$\pm$	0.0344 		} & {	\textbf{0.7545 	$\pm$	0.0027 }	} \\
\hline \textsf{	blood	} & {	0.5173 	$\pm$	0.0282 	} & {	0.4931 	$\pm$	0.0017 	} & {	0.5005 	$\pm$	0.0146 	} & {	0.7448 	$\pm$	0.0672 	} & {	0.6551 	$\pm$	0.0679 		} & {	0.6166 	$\pm$	0.0117 		} & {	\textbf{0.8387 	$\pm$	0.0039} 	} \\
\hline \textsf{	bouquet	} & {	0.4219 	$\pm$	0.0110 	} & {	0.4146 	$\pm$	0.0029 	} & {	0.4213 	$\pm$	0.0080 	} & {	0.6299 	$\pm$	0.0532 	} & {	0.4990 	$\pm$	0.0753 		} & {	0.4813 	$\pm$	0.0194 		} & {	\textbf{0.7021 	$\pm$	0.0015} 	} \\
\hline \textsf{	bugat	} & {	0.4555 	$\pm$	0.0149 	} & {	0.4313 	$\pm$	0.0027 	} & {	0.4627 	$\pm$	0.0245 	} & {	0.6540 	$\pm$	0.0549 	} & {	0.6961 	$\pm$	0.0823 		} & {	0.4964 	$\pm$	0.0303 		} & {	\textbf{0.7158 	$\pm$	0.0036 }	} \\
\hline \textsf{	cactus	} & {	0.4501 	$\pm$	0.0164 	} & {	0.4321 	$\pm$	0.0028 	} & {	0.4643 	$\pm$	0.0330 	} & {	0.6540 	$\pm$	0.0549 	} & {	0.7049 	$\pm$	0.0853 		} & {	0.4929 	$\pm$	0.0218 		} & {	\textbf{0.7222 	$\pm$	0.0052 }	} \\
\hline \textsf{	voituretuning	} & {	0.4415 	$\pm$	0.0152 	} & {	0.4012 	$\pm$	0.0032 	} & {	0.4378 	$\pm$	0.0249 	} & {	0.6125 	$\pm$	0.0509 	} & {	0.5246 	$\pm$	0.0903 		} & {	0.4748 	$\pm$	0.0219 		} & {	\textbf{0.6783 	$\pm$	0.0032 }	} \\
\hline \textsf{	car	} & {	0.4355 	$\pm$	0.0450 	} & {	0.7287 	$\pm$	0.0186 	} & {	0.5043 	$\pm$	0.0247 	} & {	0.6553 	$\pm$	0.0533 	} & {	0.4304 	$\pm$	0.0987 		} & {	0.6551 	$\pm$	0.1312 		} & {	\textbf{0.7289 	$\pm$	0.0029} 	} \\
\hline \textsf{	KDD99	} & {	0.4308 	$\pm$	0.0669 	} & {	0.8594 	$\pm$	0.1151 	} & {	0.7653 	$\pm$	0.0004 	} & {	0.8119 	$\pm$	0.0198 	} & {	0.6750 	$\pm$	0.1028 		} & {	0.8617 	$\pm$	0.0696 		} & {	\textbf{0.9086 	$\pm$	0.0005 }	} \\
\hline \textsf{	segmentation	} & {	0.2261 	$\pm$	0.0430 	} & {	0.3357 	$\pm$	0.0425 	} & {	\textbf{0.4758 	$\pm$	0.0019} 	} & {	0.3458 	$\pm$	0.0196 	} & {	0.3813 	$\pm$	0.0619 		} & {	0.3426 	$\pm$	0.0198 		} & {	0.3982 	$\pm$	0.0692 	} \\
\hline \textsf{	vowe	} & {	0.1091 	$\pm$	0.0134 	} & {	0.2003 	$\pm$	0.0161 	} & {	0.1603 	$\pm$	0.0163 	} & {	\textbf{0.2732 	$\pm$	0.0170} 	} & {	0.1625 	$\pm$	0.0238 		} & {	0.1644 	$\pm$	0.0877 		} & {	0.1920 	$\pm$	0.0098	} \\																																
\hline {\textcolor[rgb]{0.00,0.25,0.50}{\textbf{Average}}}		 & {\textcolor[rgb]{0.00,0.25,0.50}{	0.4077 }			} & {\textcolor[rgb]{0.00,0.25,0.50}{	0.4710 }			} & {	\textcolor[rgb]{0.00,0.25,0.50}{0.4935 	}		} & {\textcolor[rgb]{0.00,0.25,0.50}{	0.6807 	}		} & {	\textcolor[rgb]{0.00,0.25,0.50}{0.5913} 				} & {	\textcolor[rgb]{0.00,0.25,0.50}{0.5665} 				} & {	\textbf{\textcolor[rgb]{0.00,0.25,0.50}{0.7564} 	}		} \\
\hline
\end{tabular}}
\end{center}
\end{table*}
\begin{table*}
\begin{center}
\caption{Performance comparisions (Rand index) of benchmarking algorithms (Semi-SP), shallow models (pcGRBM and Semi-EAGR), and deep models (Semi-MG, VGAE, NMicro-DL and Micro-DL). The larger recall, the better performance. The best performance on each data set is bolded.}
\label{tab:results1} \scalebox{0.9}{
\begin{tabular}{|l|c|c|c|c|c|c|c|c|}
\hline
\textsf{ \bf{	Dataset	}} & {	Semi-SP			} & {	pcGRBM			} & {	Semi-EAGR			} & {	Semi-MG			} & {	VGAE				} & {	NMicro-DL				} & {	\textbf{Micro-DL}			} \\
\hline \textsf{	aquarium	} & {	0.4864 	$\pm$	0.0007 	} & {	0.4892 	$\pm$	0.0018 	} & {	0.4955 	$\pm$	0.0047 	} & {	0.5100 	$\pm$	0.0187 	} & {	0.4677 	$\pm$	0.0028 		} & {	0.5005 	$\pm$	0.0077 		} & {	\textbf{0.5388 	$\pm$	0.0033 }	} \\
\hline \textsf{	bathroom	} & {	0.4555 	$\pm$	0.0001 	} & {	0.4517 	$\pm$	0.0017 	} & {	0.4548 	$\pm$	0.0099 	} & {	0.5813 	$\pm$	0.0478 	} & {	0.5462 	$\pm$	0.0480 		} & {	0.4731 	$\pm$	0.0099 		} & {\textbf{	0.6395 	$\pm$	0.0141 }	} \\
\hline \textsf{	blog	} & {	0.4812 	$\pm$	0.0043 	} & {	0.4812 	$\pm$	0.0013 	} & {	0.4817 	$\pm$	0.0088 	} & {	0.5315 	$\pm$	0.0277 	} & {	0.5134 	$\pm$	0.0408 		} & {	0.5008 	$\pm$	0.0103 		} & {	\textbf{0.5768 	$\pm$	0.0029 }	} \\
\hline \textsf{	blood	} & {	0.4451 	$\pm$	0.0163 	} & {	0.4304 	$\pm$	0.0011 	} & {	0.4354 	$\pm$	0.0084 	} & {	0.6108 	$\pm$	0.0661 	} & {	0.5346 	$\pm$	0.0376 		} & {	0.5196 	$\pm$	0.0088 		} & {	\textbf{0.7084 	$\pm$	0.0051 }	} \\
\hline \textsf{	bouquet	} & {	0.4980 	$\pm$	0.0005 	} & {	0.5022 	$\pm$	0.0026 	} & {	0.5035 	$\pm$	0.0057 	} & {	0.4922 	$\pm$	0.0058 	} & {	0.5076 	$\pm$	0.0338 		} & {	0.4951 	$\pm$	0.0062 		} & {	\textbf{0.5032 	$\pm$	0.0033 }	} \\
\hline \textsf{	bugat	} & {	0.4909 	$\pm$	0.0014 	} & {	0.4887 	$\pm$	0.0025 	} & {	0.4865 	$\pm$	0.0073 	} & {	0.5224 	$\pm$	0.0109 	} & {	0.5177 	$\pm$	0.0307 		} & {	0.4968 	$\pm$	0.0104 		} & {	\textbf{0.5300 	$\pm$	0.0025 }	} \\
\hline \textsf{	cactus	} & {	0.4873 	$\pm$	0.0002 	} & {	0.4894 	$\pm$	0.0018 	} & {	0.5043 	$\pm$	0.0138 	} & {	0.5224 	$\pm$	0.0109 	} & {	0.5224 	$\pm$	0.0285 		} & {	0.4915 	$\pm$	0.0051 		} & {	\textbf{0.5372 	$\pm$	0.0065 	}} \\
\hline \textsf{	voituretuning	} & {	0.5030 	$\pm$	0.0060 	} & {	0.5084 	$\pm$	0.0014 	} & {	0.5305 	$\pm$	0.0064 	} & {	0.4759 	$\pm$	0.0062 	} & {	0.4976 	$\pm$	0.0274 		} & {	0.4958 	$\pm$	0.0076 		} & {	\textbf{0.4747 	$\pm$	0.0041} 	} \\
\hline \textsf{	car	} & {	0.4876 	$\pm$	0.0075 	} & {	0.5390 	$\pm$	0.0074 	} & {	0.5689 	$\pm$	0.0151 	} & {	0.5274 	$\pm$	0.0073 	} & {	0.5036 	$\pm$	0.0273 		} & {	0.5274 	$\pm$	0.0246 		} & {	\textbf{0.5410 	$\pm$	0.0025} 	} \\
\hline \textsf{	KDD99	} & {	0.4776 	$\pm$	0.0584 	} & {	0.8940 	$\pm$	0.0956 	} & {	0.8237 	$\pm$	0.0006 	} & {	0.8653 	$\pm$	0.0164 	} & {	0.6528 	$\pm$	0.0774 		} & {	0.9072 	$\pm$	0.0466 		} & {	\textbf{0.9389 	$\pm$	0.0003} 	} \\
\hline \textsf{	segmentation	} & {	0.5909 	$\pm$	0.1082 	} & {	0.7942 	$\pm$	0.0166 	} & {	0.7933 	$\pm$	0.0094 	} & {	0.2892 	$\pm$	0.1028 	} & {	0.7447 	$\pm$	0.0456 		} & {	0.7256 	$\pm$	0.0634 		} & {	\textbf{0.7991 	$\pm$	0.0455} 	} \\
\hline \textsf{	vowe	} & {	0.6979 	$\pm$	0.0301 	} & {	0.5407 	$\pm$	0.0612 	} & {	\textbf{0.8024 	$\pm$	0.0091} 	} & {	0.2619 	$\pm$	0.1206 	} & {	0.7266 	$\pm$	0.0514 		} & {	0.6729 	$\pm$	0.2284 		} & {	0.7644 	$\pm$	0.0352 	} \\																															
\hline {\textcolor[rgb]{0.00,0.25,0.50}{\textbf{Average}}}		 & {	\textcolor[rgb]{0.00,0.25,0.50}{0.5085 }			} & {	\textcolor[rgb]{0.00,0.25,0.50}{0.5508 }			} & {	\textcolor[rgb]{0.00,0.25,0.50}{0.5285 }			} & {	\textcolor[rgb]{0.00,0.25,0.50}{0.5639} 			} & {	\textcolor[rgb]{0.00,0.25,0.50}{0.5264 }				} & {\textcolor[rgb]{0.00,0.25,0.50}{	0.5408 	}			} & {	\textbf{\textcolor[rgb]{0.00,0.25,0.50}{0.5988} }			} \\
\hline
\end{tabular}}
\end{center}
\end{table*}
\begin{table*}
\begin{center}
\caption{Performance comparisions (Rank) of benchmarking algorithms (Semi-SP), shallow models (pcGRBM and Semi-EAGR), and deep models (Semi-MG, VGAE, NMicro-DL and Micro-DL). The smaller rank, the better performance.}
\label{tab:results2} \scalebox{1}{
\begin{tabular}{|l|c|c|c|c|c|c|c|c|}
\hline
\textsf{ \bf{	Dataset	}} & {	Semi-SP				} & {	pcGRBM				} & {	Semi-EAGR				} & {	Semi-MG				} & {	VGAE				} & {	NMicro-DL				} & {	\textbf{Micro-DL}				} & {	\textcolor[rgb]{0.00,0.25,0.50}{\textbf{Total}}	} \\
\hline \textsf{	aquarium	} & {	-0.1006 	(	65 	)	} & {	-0.1218 	(	71 	)	} & {	-0.0827 	(	58 	)	} & {	0.0997 	(	25 	)	} & {	0.0453 	(	34 	)	} & {	-0.0091 	(	45 	)	} & {	0.1695 	(	6 	)	} & {	\textcolor[rgb]{0.00,0.25,0.50}{304}	} \\
\hline \textsf{	bathroom	} & {	-0.0787 	(	56 	)	} & {	-0.1708 	(	80 	)	} & {	-0.1214 	(	69 	)	} & {	0.1601 	(	8 	)	} & {	0.0142 	(	39 	)	} & {	-0.0341 	(	49 	)	} & {	0.2310 	(	2 	)	} & {	\textcolor[rgb]{0.00,0.25,0.50}{303}	} \\
\hline \textsf{	blog	} & {	-0.0788 	(	57 	)	} & {	-0.1433 	(	75 	)	} & {	-0.1142 	(	67 	)	} & {	0.1175 	(	18 	)	} & {	0.0274 	(	37 	)	} & {	-0.0040 	(	43 	)	} & {	0.1951 	(	3 	)	} & {	\textcolor[rgb]{0.00,0.25,0.50}{300}	} \\
\hline \textsf{	blood	} & {	-0.1502 	(	77 	)	} & {	-0.1880 	(	82 	)	} & {	-0.1781 	(	81 	)	} & {	0.1493 	(	9 	)	} & {	0.0886 	(	26 	)	} & {	0.0314 	(	36 	)	} & {	0.2468 	(	1 	)	} & {	\textcolor[rgb]{0.00,0.25,0.50}{312}	} \\
\hline \textsf{	bouquet	} & {	-0.0905 	(	61 	)	} & {	-0.1137 	(	66 	)	} & {	-0.0716 	(	55 	)	} & {	0.1153 	(	19 	)	} & {	0.0094 	(	40 	)	} & {	-0.0205 	(	48 	)	} & {	0.1713 	(	5 	)	} & {	\textcolor[rgb]{0.00,0.25,0.50}{294}	} \\
\hline \textsf{	bugat	} & {	-0.0977 	(	63 	)	} & {	-0.1486 	(	76 	)	} & {	-0.0975 	(	62 	)	} & {	0.1070 	(	22 	)	} & {	0.1325 	(	14 	)	} & {	-0.0441 	(	51 	)	} & {	0.1484 	(	10 	)	} & {	\textcolor[rgb]{0.00,0.25,0.50}{298}	} \\
\hline \textsf{	cactus	} & {	-0.0982 	(	64 	)	} & {	-0.1378 	(	73 	)	} & {	-0.1243 	(	72 	)	} & {	0.1105 	(	21 	)	} & {	0.1388 	(	13 	)	} & {	-0.0491 	(	52 	)	} & {	0.1602 	(	7 	)	} & {	\textcolor[rgb]{0.00,0.25,0.50}{302}	} \\
\hline \textsf{	voituretuning	} & {	-0.0900 	(	60 	)	} & {	-0.1215 	(	70 	)	} & {	-0.0577 	(	53 	)	} & {	0.0819 	(	27 	)	} & {	0.0565 	(	31 	)	} & {	-0.0163 	(	47 	)	} & {	0.1469 	(	11 	)	} & {	\textcolor[rgb]{0.00,0.25,0.50}{299}	} \\
\hline \textsf{	car	} & {	-0.1607 	(	79 	)	} & {	0.1283 	(	16 	)	} & {	-0.0858 	(	59 	)	} & {	0.0786 	(	28 	)	} & {	-0.1407 	(	74 	)	} & {	0.0512 	(	32 	)	} & {	0.1290 	(	15 	)	} & {	\textcolor[rgb]{0.00,0.25,0.50}{303}	} \\
\hline \textsf{	KDD99	} & {	-0.4220 	(	84 	)	} & {	0.1244 	(	17 	)	} & {	0.0211 	(	38 	)	} & {	0.1045 	(	23 	)	} & {	-0.1508 	(	78 	)	} & {	0.1460 	(	12 	)	} & {	0.1766 	(	4 	)	} & {	\textcolor[rgb]{0.00,0.25,0.50}{256}	} \\
\hline \textsf{	segmentation	} & {	-0.2102 	(	83 	)	} & {	0.0507 	(	33 	)	} & {	0.0999 	(	24 	)	} & {	-0.1175 	(	68 	)	} & {	0.0694 	(	30 	)	} & {	-0.0046 	(	44 	)	} & {	0.1125 	(	20 	)	} & {	\textcolor[rgb]{0.00,0.25,0.50}{302}	} \\
\hline \textsf{	vowe	} & {	-0.0385 	(	50 	)	} & {	-0.0615 	(	54 	)	} & {	0.0043 	(	41 	)	} & {	0.0343 	(	35 	)	} & {	-0.0003 	(	42 	)	} & {	-0.0093 	(	46 	)	} & {	0.0710 	(	29 	)	} & {	\textcolor[rgb]{0.00,0.25,0.50}{297}	} \\
																																					
\hline {\textcolor[rgb]{0.00,0.25,0.50}{\textbf{Total}}}		 & {			\textcolor[rgb]{0.00,0.25,0.50}{799} 		} & {			\textcolor[rgb]{0.00,0.25,0.50}{713} 		} & {			\textcolor[rgb]{0.00,0.25,0.50}{679} 		} & {			\textcolor[rgb]{0.00,0.25,0.50}{303} 		} & {			\textcolor[rgb]{0.00,0.25,0.50}{458} 		} & {			\textcolor[rgb]{0.00,0.25,0.50}{505} 		} & {			\textbf{\textcolor[rgb]{0.00,0.25,0.50}{113} }		} & {	\textcolor[rgb]{0.00,0.25,0.50}{3570} 	} \\
\hline {\textcolor[rgb]{0.00,0.25,0.50}{\textbf{Average Rank}}}		 & {		\textcolor[rgb]{0.00,0.25,0.50}{	66.5833 }		} & {			\textcolor[rgb]{0.00,0.25,0.50}{59.4167 }		} & {			\textcolor[rgb]{0.00,0.25,0.50}{56.5833 	}	} & {			\textcolor[rgb]{0.00,0.25,0.50}{25.2500 }		} & {		\textcolor[rgb]{0.00,0.25,0.50}{	38.1667 }		} & {			\textcolor[rgb]{0.00,0.25,0.50}{42.0833 }		} & {			\textbf{\textcolor[rgb]{0.00,0.25,0.50}{9.4167} }		} & {		} \\
\hline
\end{tabular}}
\end{center}
\end{table*}
\subsection{Friedman Aligned Ranks Test}
In the experiments, the Friedman Aligned Ranks test is based on 12 datasets and 7 contrast algorithms of ranks. The average ranks provide a fair comparison of these algorithms. On average, the proposed Micro-DL ranks the first with the value of 9.4167; the Semi-MG, VGAE and NMicro-DL ranks the second, third and fourth, with the values of 25.2500, 38.1667 and 42.0833, respectively; and the fifth, sixth and the last are the Semi-EAGR, pcGRBM and Semi-SP with ranks 56.5833, 59.4167 and 66.5833, respectively. Under the null hypothesis, the Friedman Aligned Ranks test is used to check whether the metrical sum of aligned ranks are different from the average of total aligned rank $\widehat{R}_{j}=510$:
\begin{equation}
\begin{aligned}
  &\sum_{j=1}^k\widehat{R}^2_{.,j}\\
  &=799^{2}+713^{2}+679^{2}+303^{2}+458^{2}+505^{2}+113^{2}\\
  &=2177178,
 \end{aligned}
\end{equation}
\begin{equation}
\begin{aligned}
  \sum_{j=1}^k\widehat{R}^2_{i,.}&=304^{2}+303^{2}+300^{2}+312^{2}+294^{2}+298^{2}+302^{2}\\
  &+299^{2}+303^{2}+256^{2}+302^{2}+297^{2}=1064172,
 \end{aligned}
\end{equation}
\begin{equation}
\begin{aligned}
 T&=\frac{(7-1)(2177178-7 \cdot 12^2(7\cdot 12+1)^2/4)}{7 \cdot 12(7\cdot12+1)(2\cdot7\cdot12+1)/6-1064172/7}\\
 &=43.5744,
\end{aligned}
\end{equation}
With seven algorithms and 12 data sets, $T$ is a chi-square distrbution with six degree-of-freedom. The null hypothesis is rejected because the $p$-value of $T$ is $1\times 10^{-8}$ which is far less than 0.05. Therefore, we can conclude that these algorithms are significantly different.
\subsection{Friedman + Post-hoc Nemenyi Tests}
The results of Friedman + pos-hoc Nemenyi tests \cite{2021A} among all contrastive methods are shown in Fig. 3. Besides the results of Micro-DL versus Semi-MG, most of the test results between Micro-DL framework and other contrastive methods are less than 0.05. Hence, Friedman + pos-hos Nemenyi tests confirm the striking differences between Micro-DL framework and most other contrastive methods at the significance level (5\%). Furthermore, the result of Friedman + pos-hoc Nemenyi tests between Micro-DL and NMicro-DL is 0.0023 which is far below 0.05. So, there is striking difference between the Micro-DL and NMicro-DL frameworks under the stimulation of SPI in the former. In fact, the results of Micro-DL versus Semi-SP and Micro-DL versus pcGRBM are approximately $3\times10^{-7}$, which are far less than 0.05. It's obvious that there are significant difference between Micro-DL versus Semi-SP and pcGRBM.
\begin{figure}[h]
 \centering
    \includegraphics[scale=0.4025]{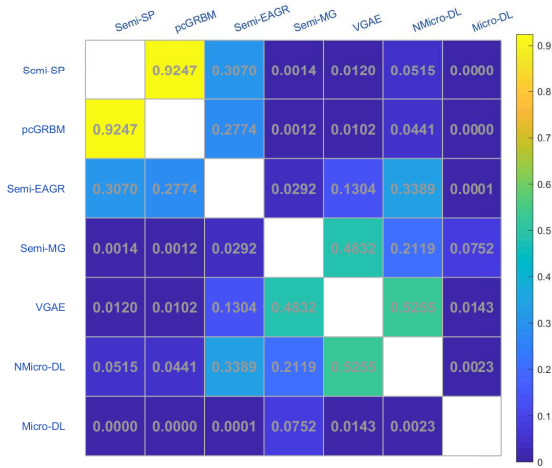}
\\
\caption{The results of Friedman + post-hoc Nemenyi tests among all contrastive algorithms.
} \label{fig:1}
\end{figure}
\subsection{Effectiveness of the SPI}
\begin{figure*}[!htbp]
\vspace{0.5mm} \centering
 \includegraphics[scale=0.4015]{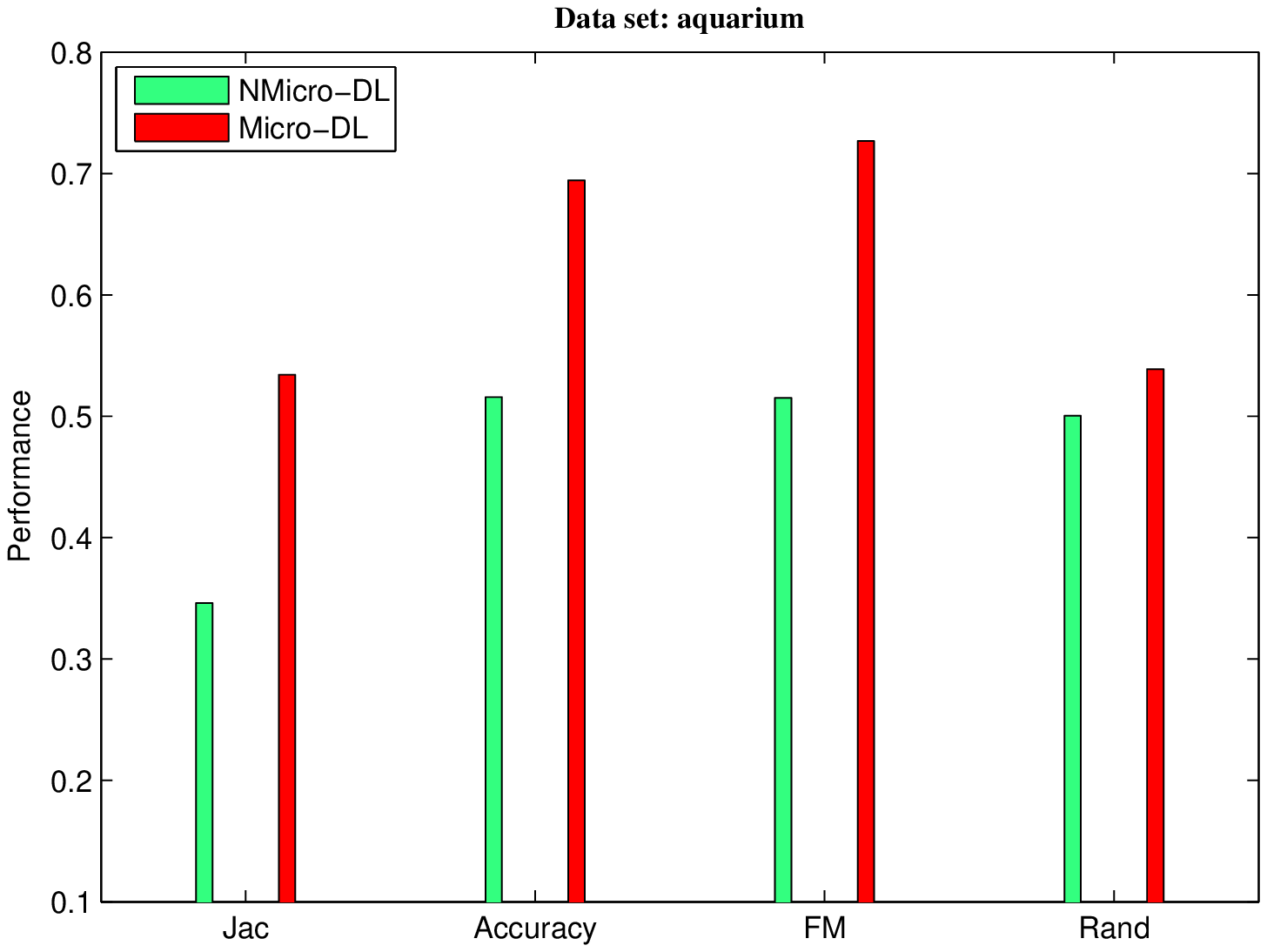}
   \includegraphics[scale=0.4015]{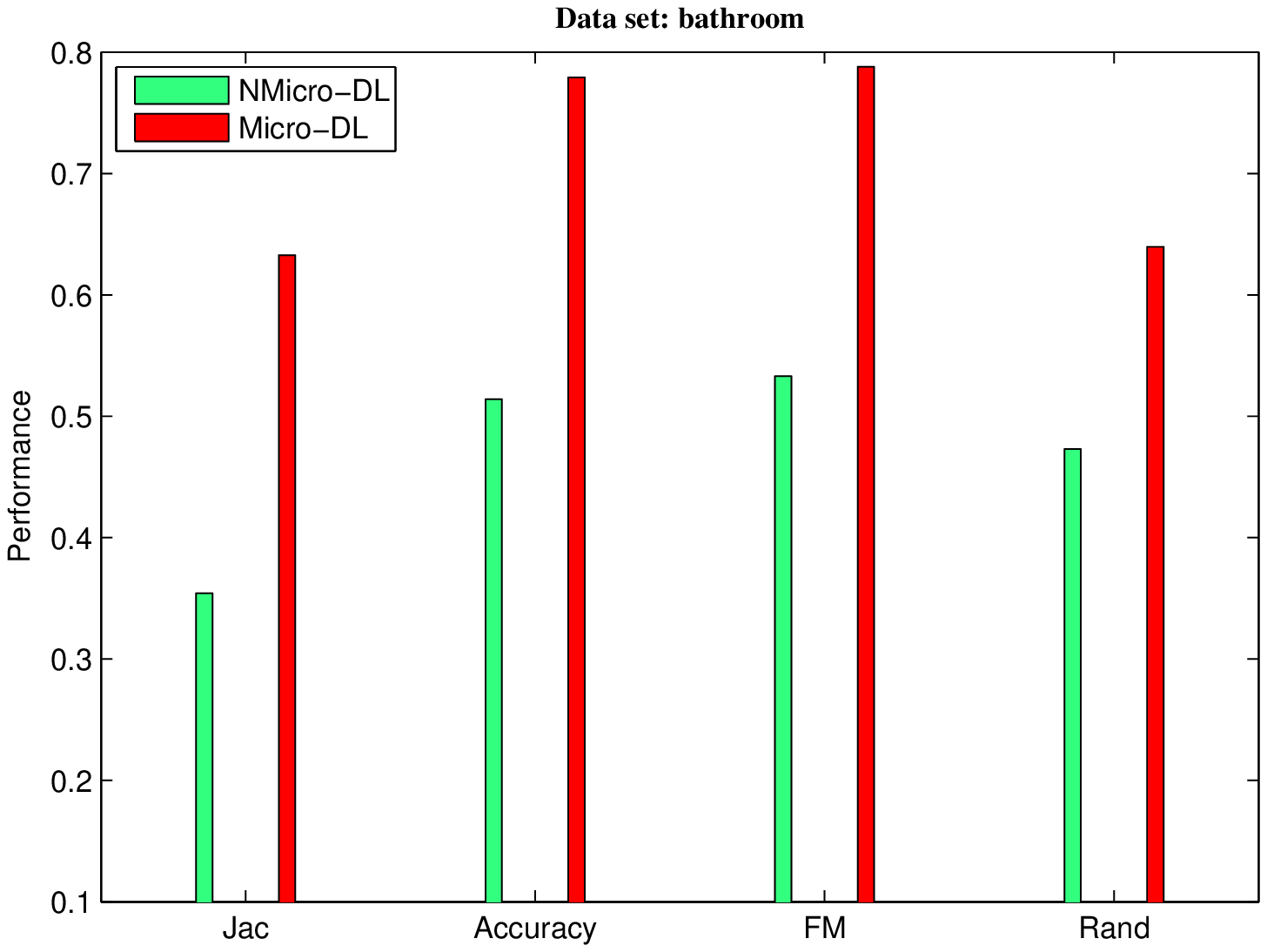}
   \includegraphics[scale=0.4015]{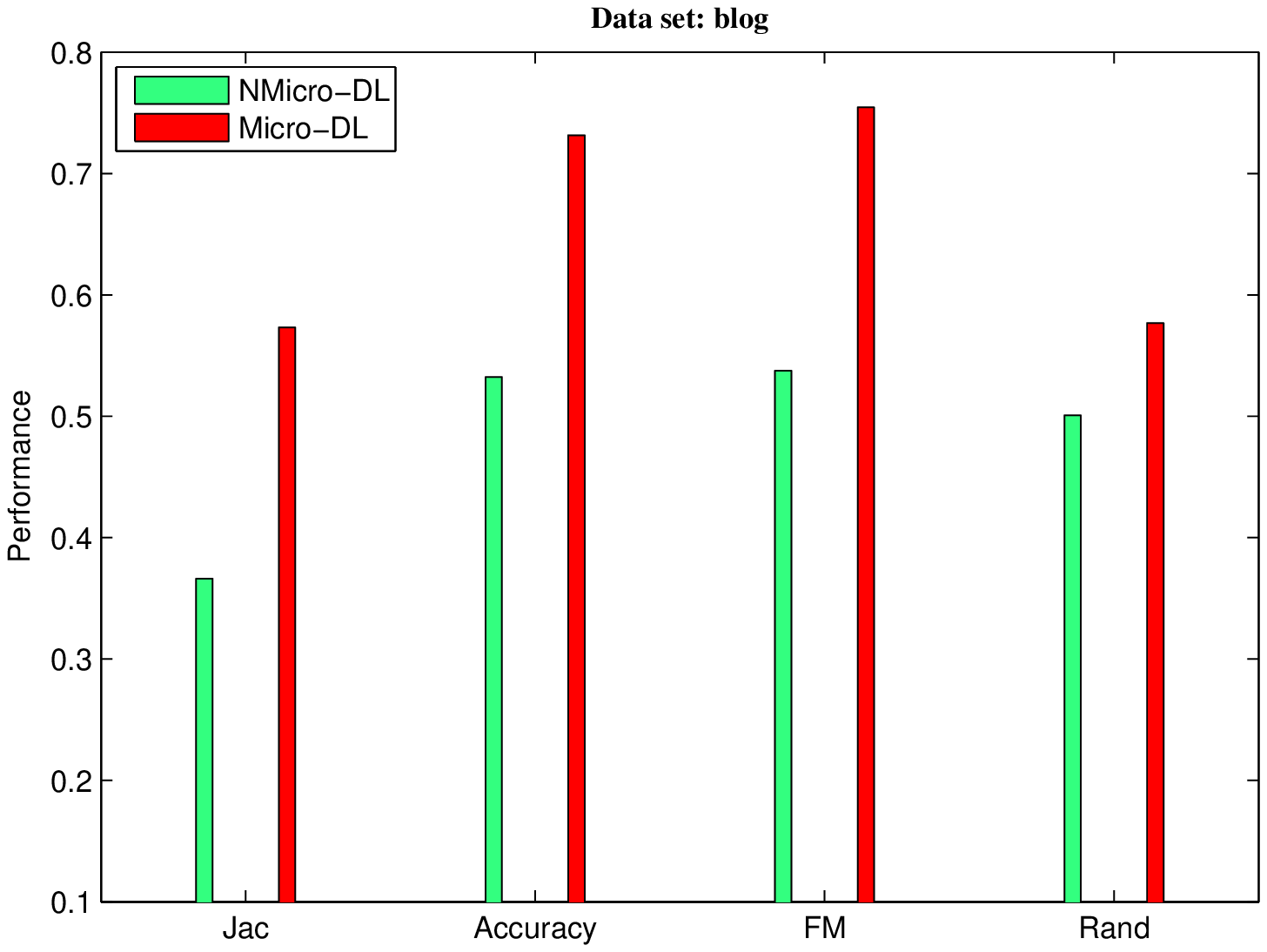}
    \includegraphics[scale=0.4015]{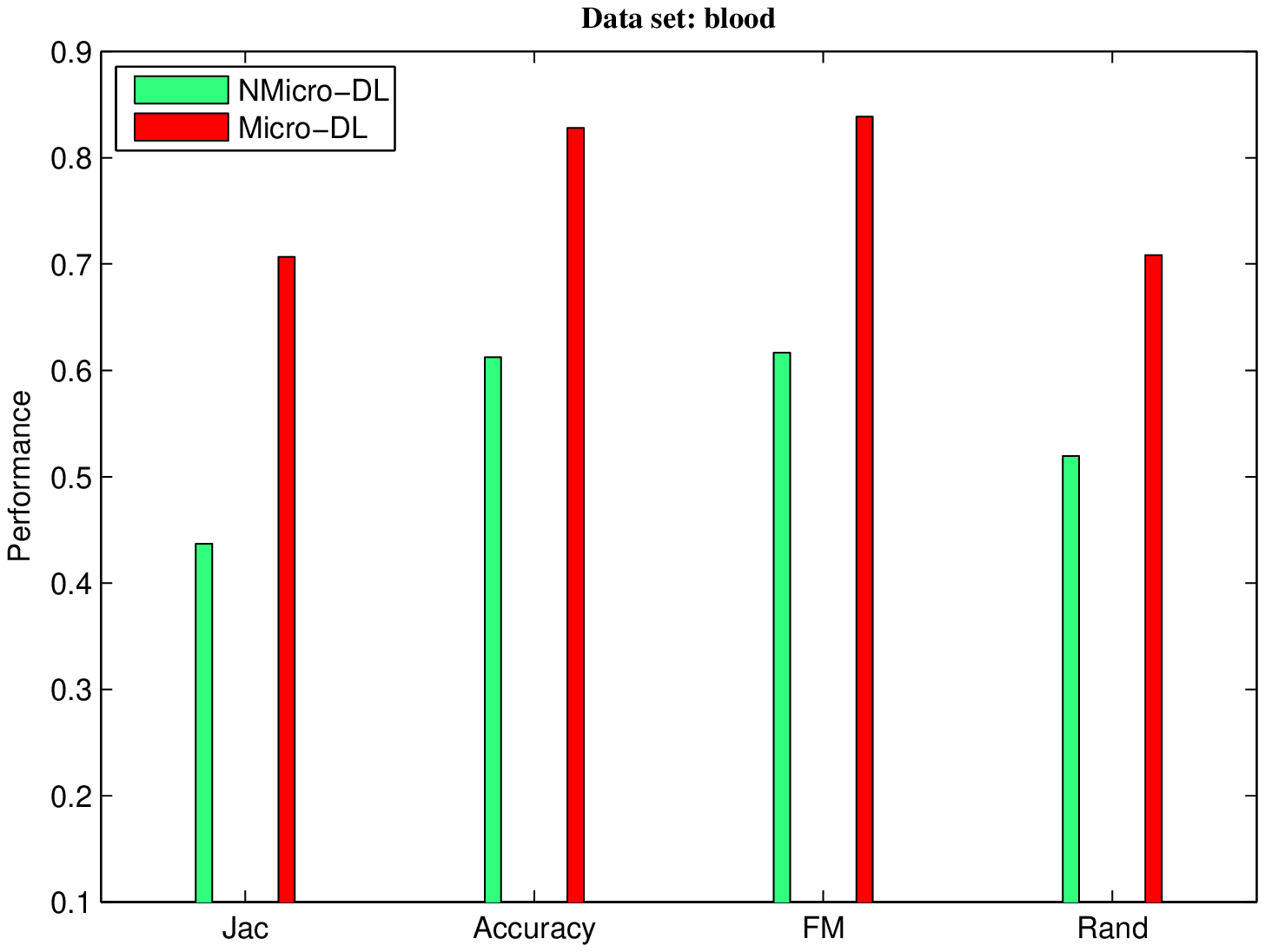}
   \includegraphics[scale=0.4015]{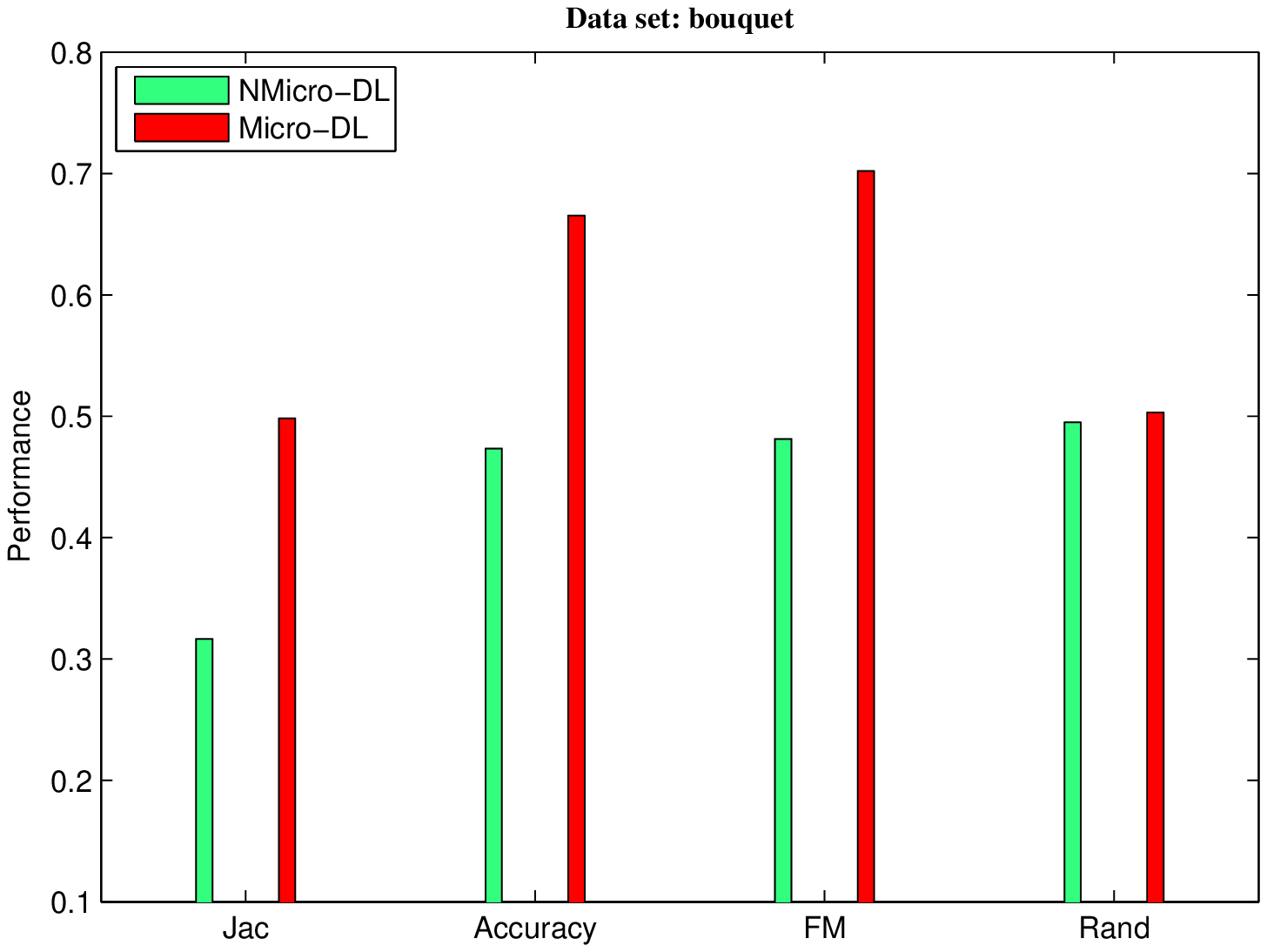}
    \includegraphics[scale=0.4015]{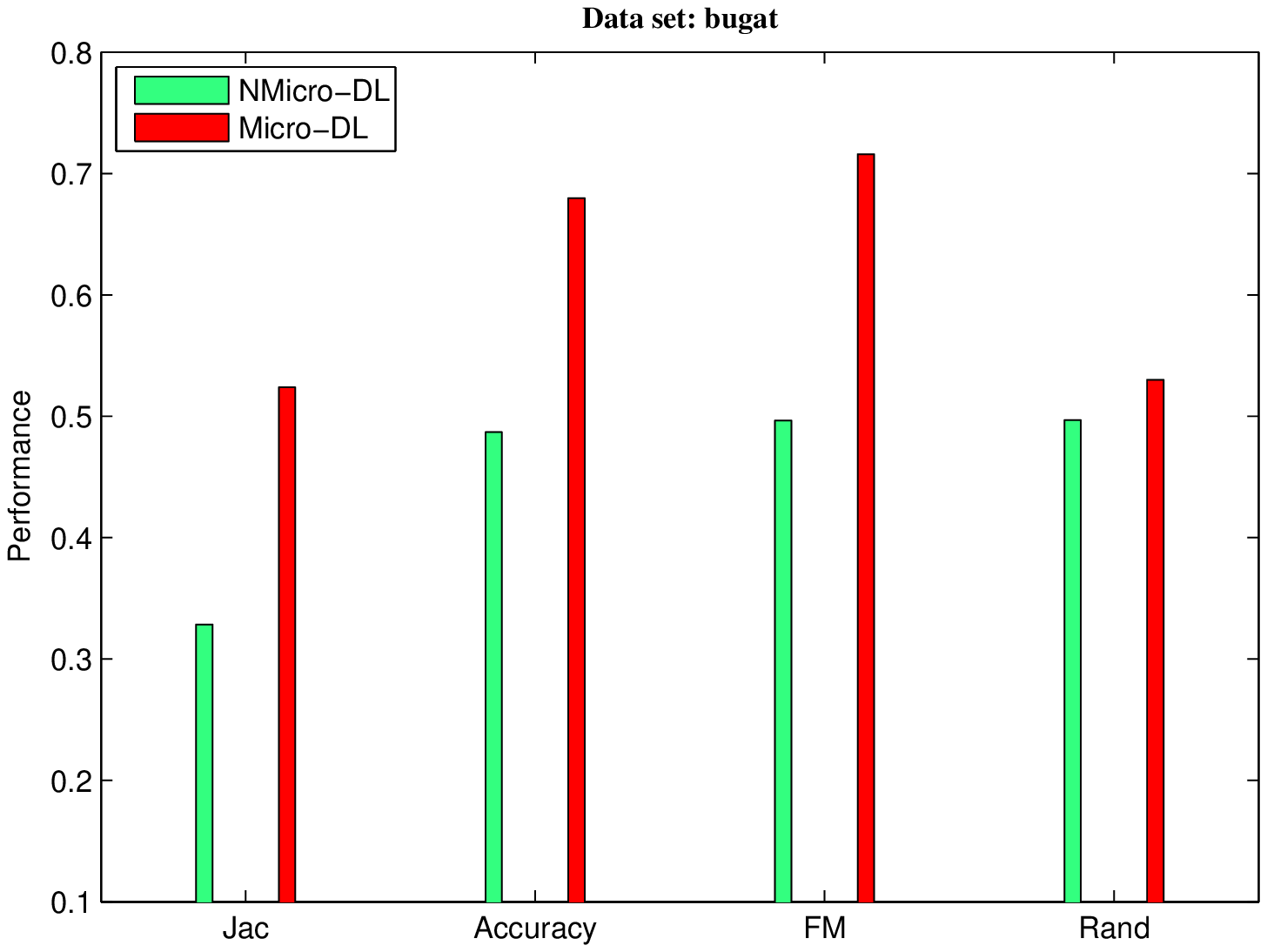}
   \includegraphics[scale=0.4015]{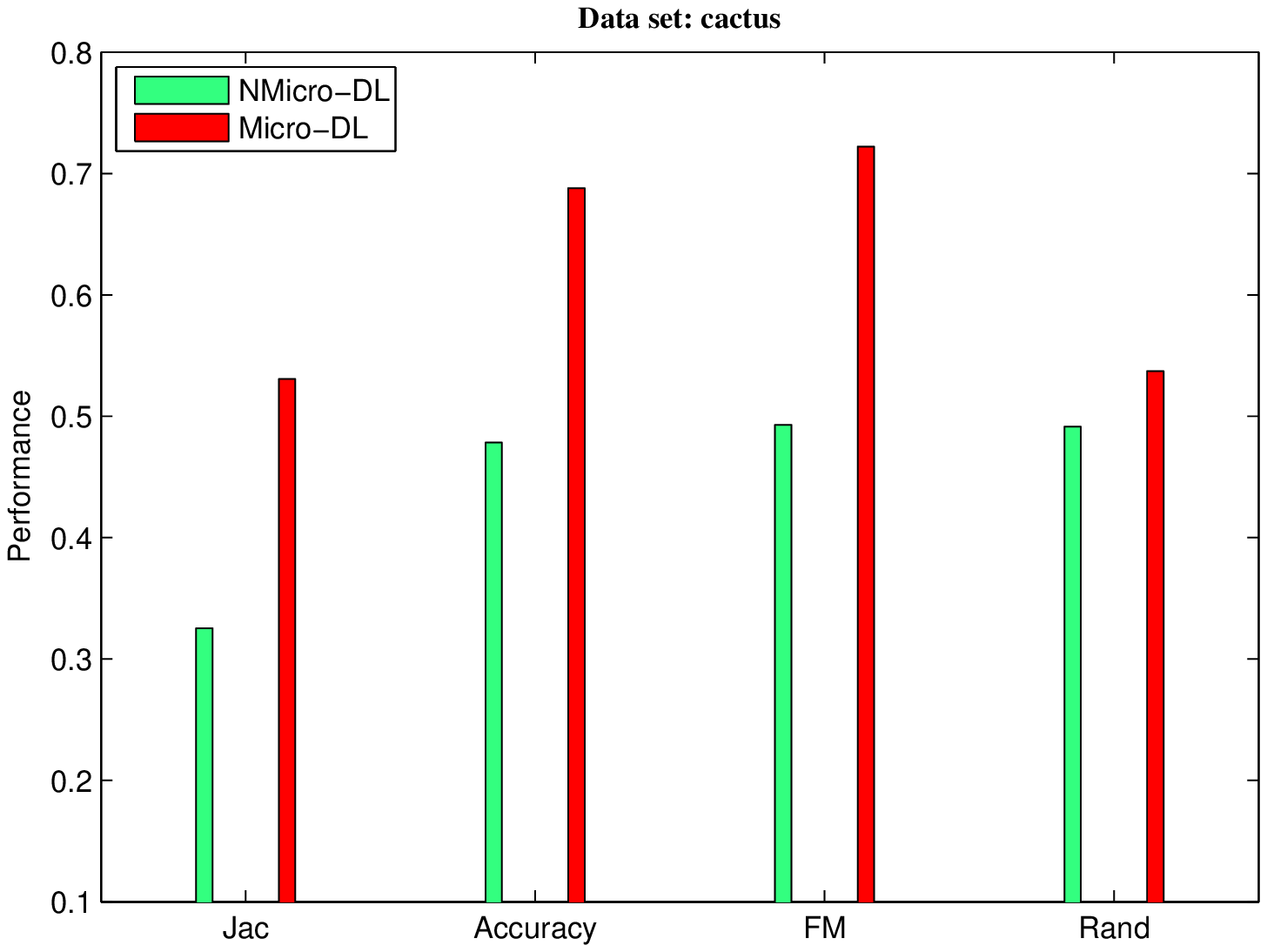}
    \includegraphics[scale=0.4015]{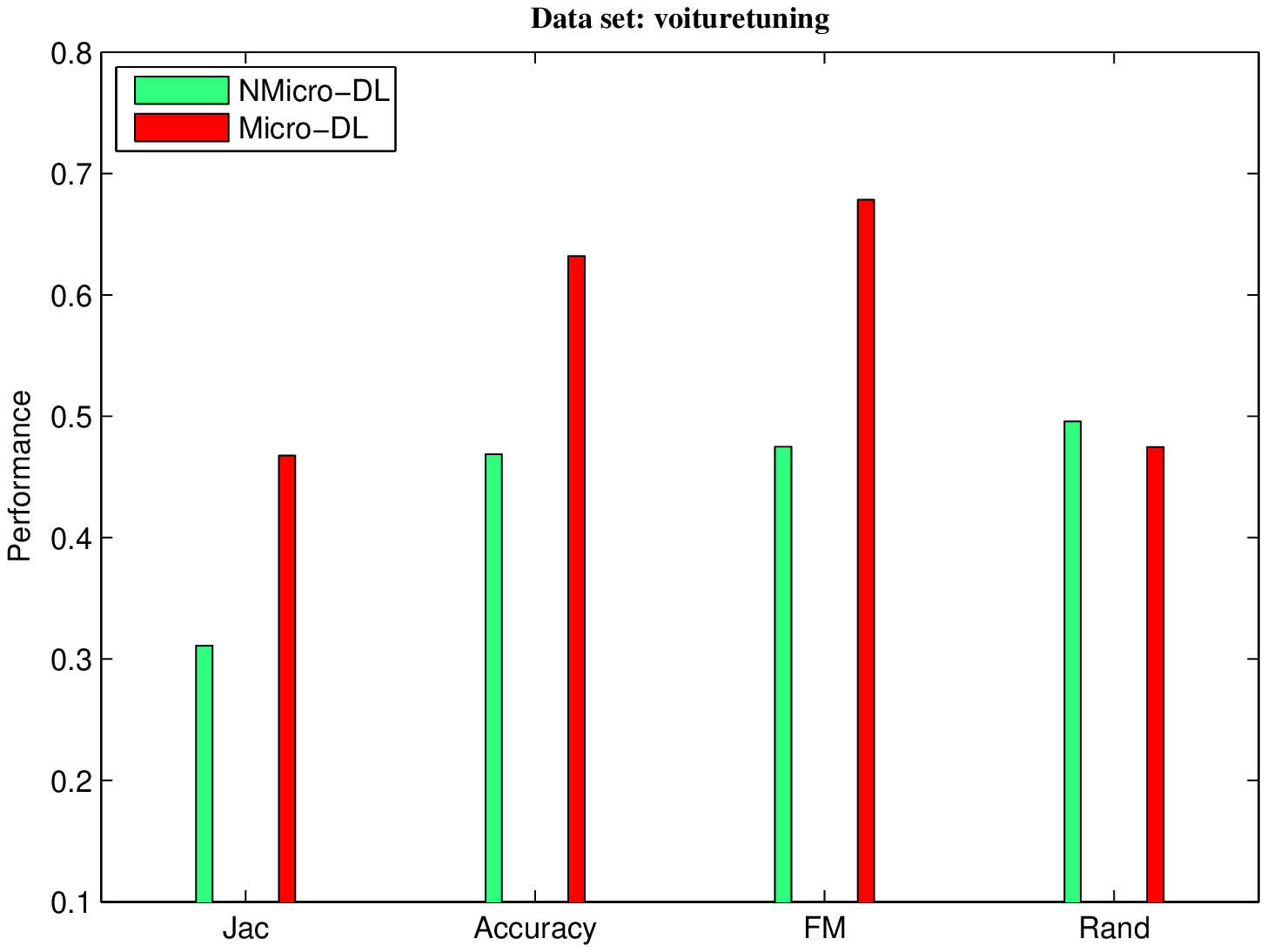}
  \includegraphics[scale=0.4015]{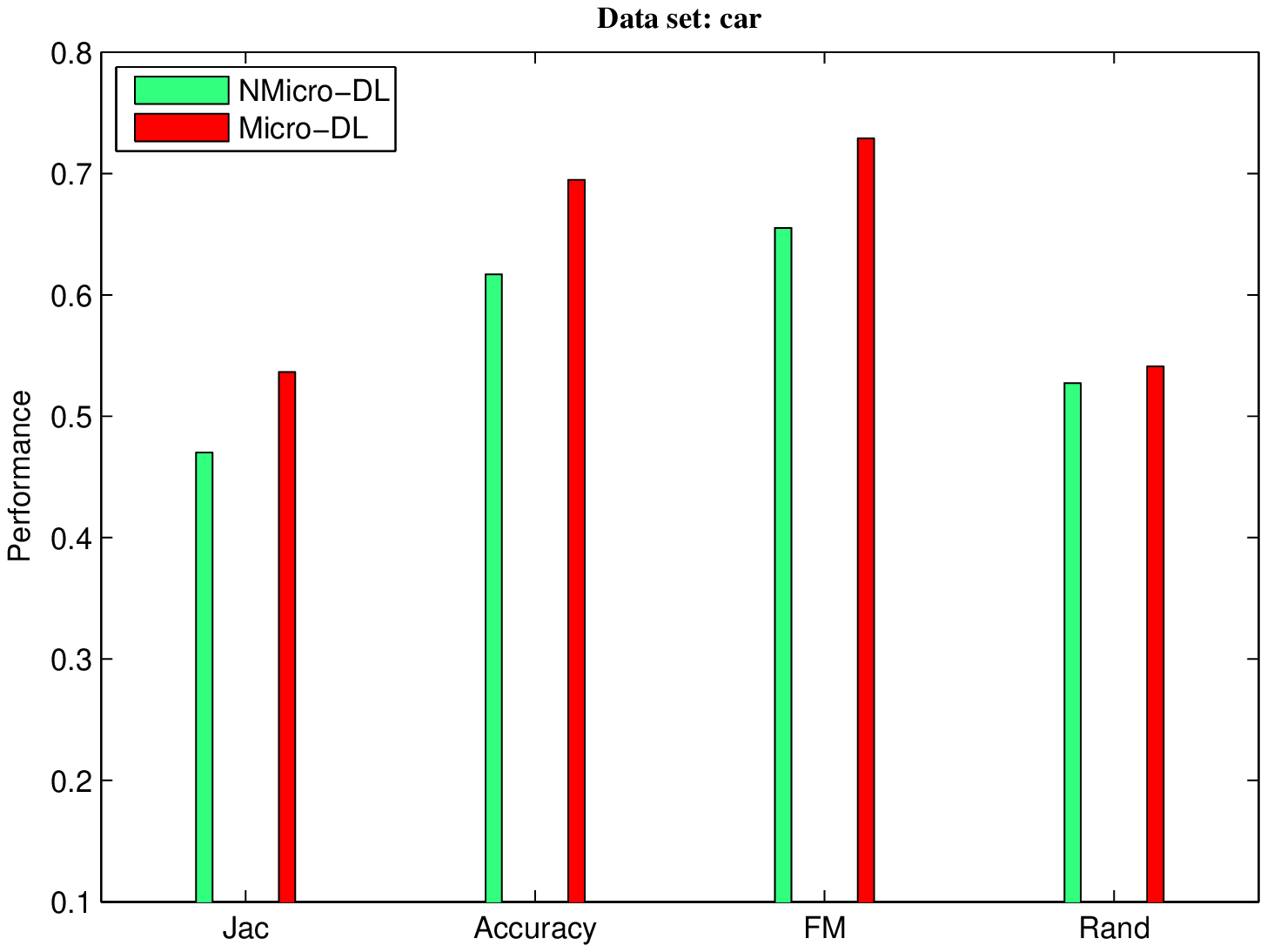}
    \includegraphics[scale=0.4015]{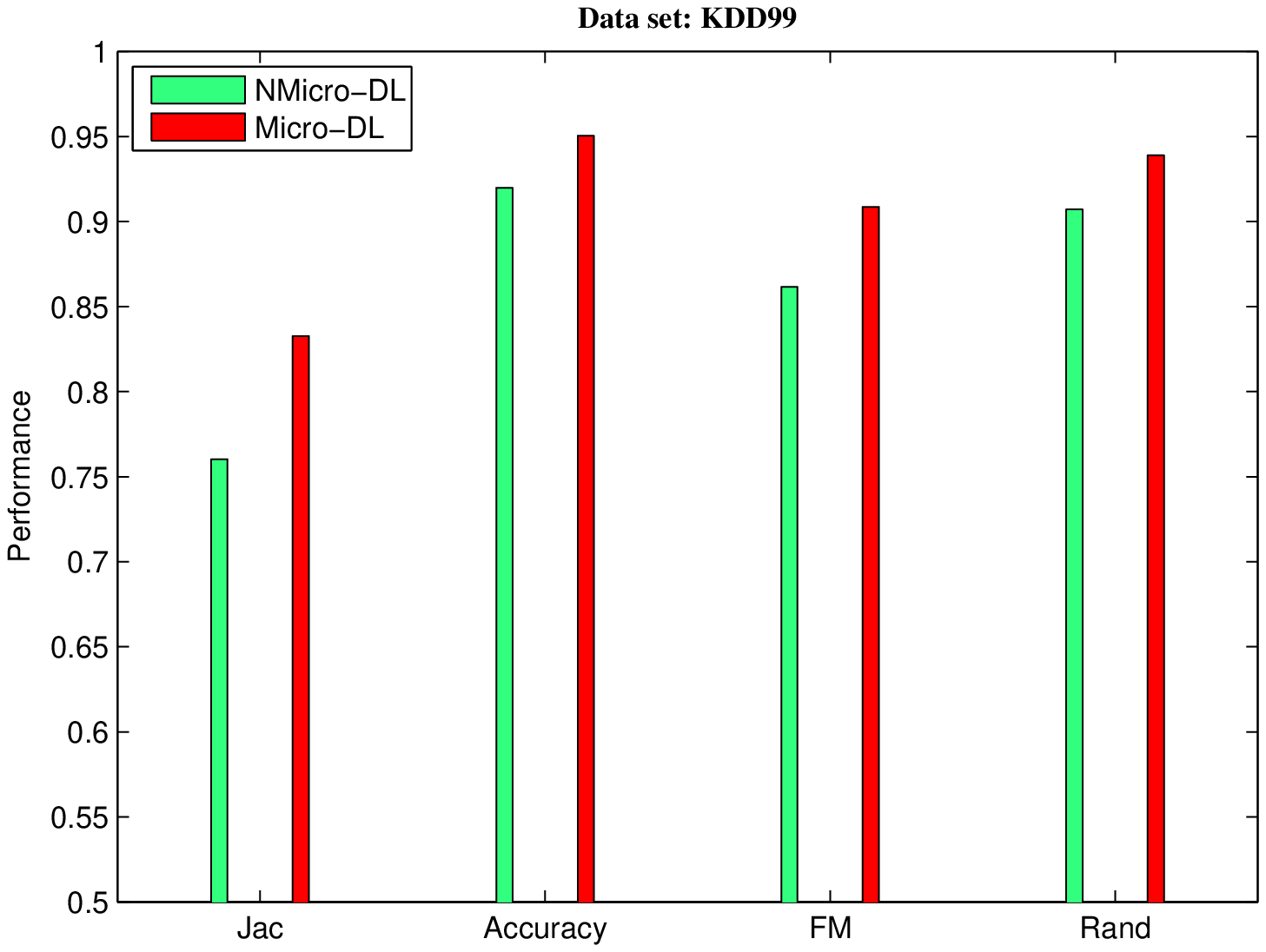}
    \includegraphics[scale=0.4015]{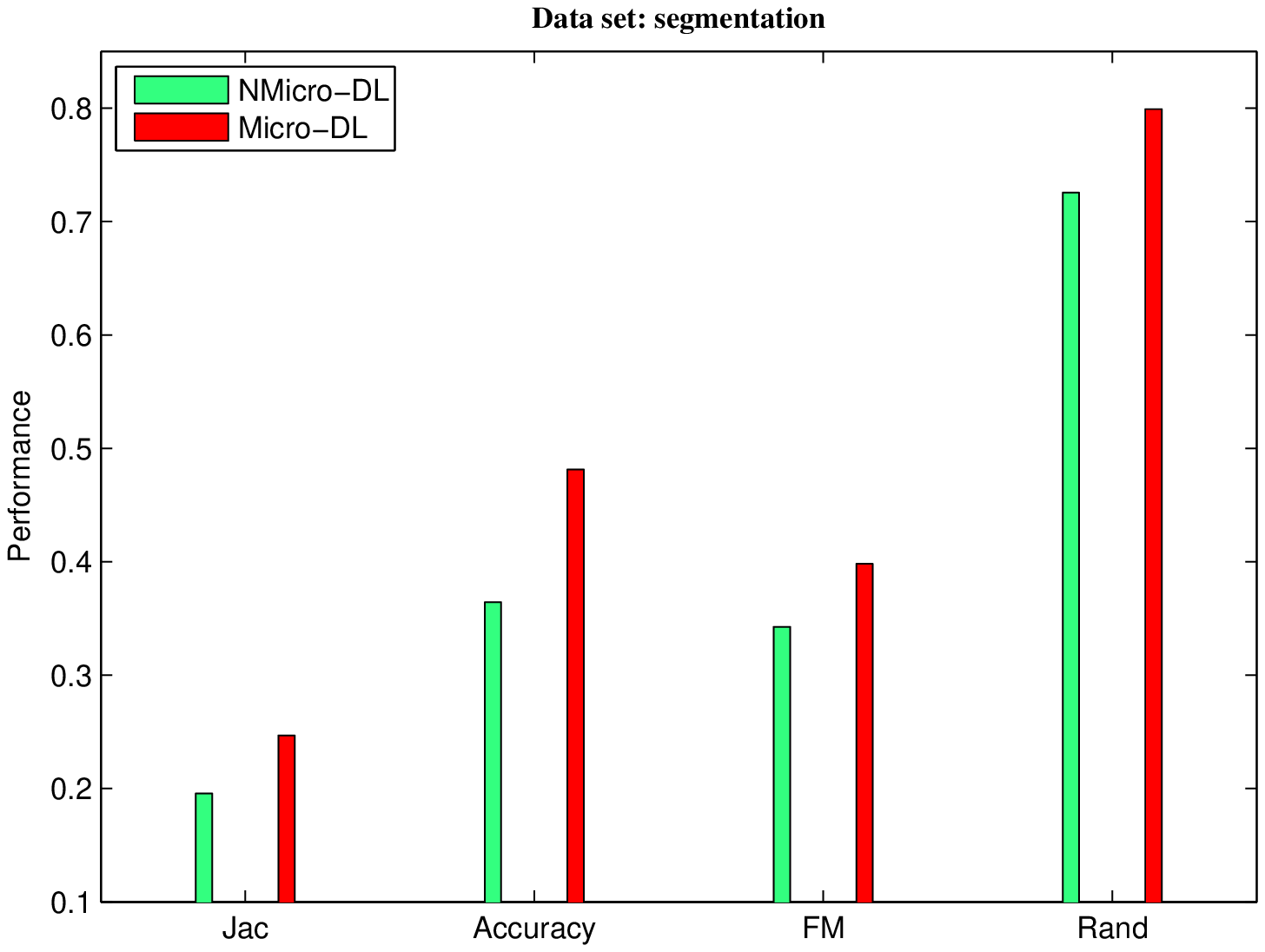}
       \includegraphics[scale=0.4015]{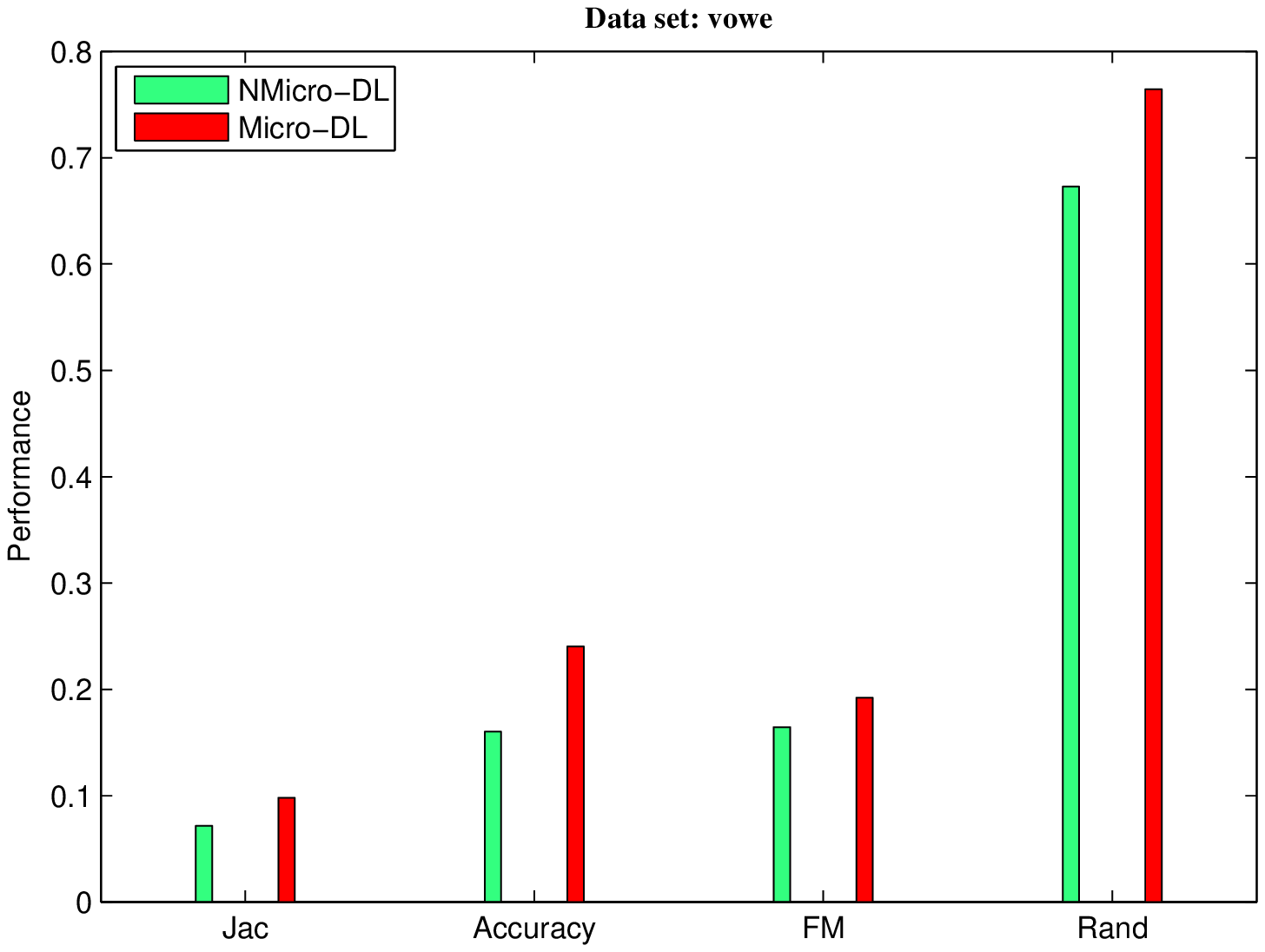}
  \caption{The effectiveness of the SPI.}
\label{rbm}
\end{figure*}
\indent To prove the effectiveness of SPI in the Micro-DL architecture, we compare the performance of our Micro-DL architecture with NMicro-DL which is a most related unsupervised deep representation learning framework. The results are shown in Fig. 4. It is obvious that the proposed Micro-DL architecture shows higher performances than NMicro-DL framework for clustering. Although the depth structures of Micro-DL and NMicro-DL are almost identical, the representation learning capability is improved by constantly stimulating the representation probability distribution with the SPI.
\subsection{Sensibility of the Scale Coefficient}
To show the sensibility of the scale coefficient $\alpha$, we gradually increase it from 0.1 to 0.9 with 0.1 per step in the learning process of our Micro-DL architecture. The effect of the scale coefficient on the clustering performances (e.g. Accuracy, Jac, FM and Rand) are shown in Fig. 5. From the results, we can obtain that the clustering performances increase as the scale coefficient is increased from 0.1 to 0.3. Then the performances begin to decline as $\alpha$ is increased from 0.4 to 0.9.
\begin{figure}[h]
 \centering
 \includegraphics[scale=0.705]{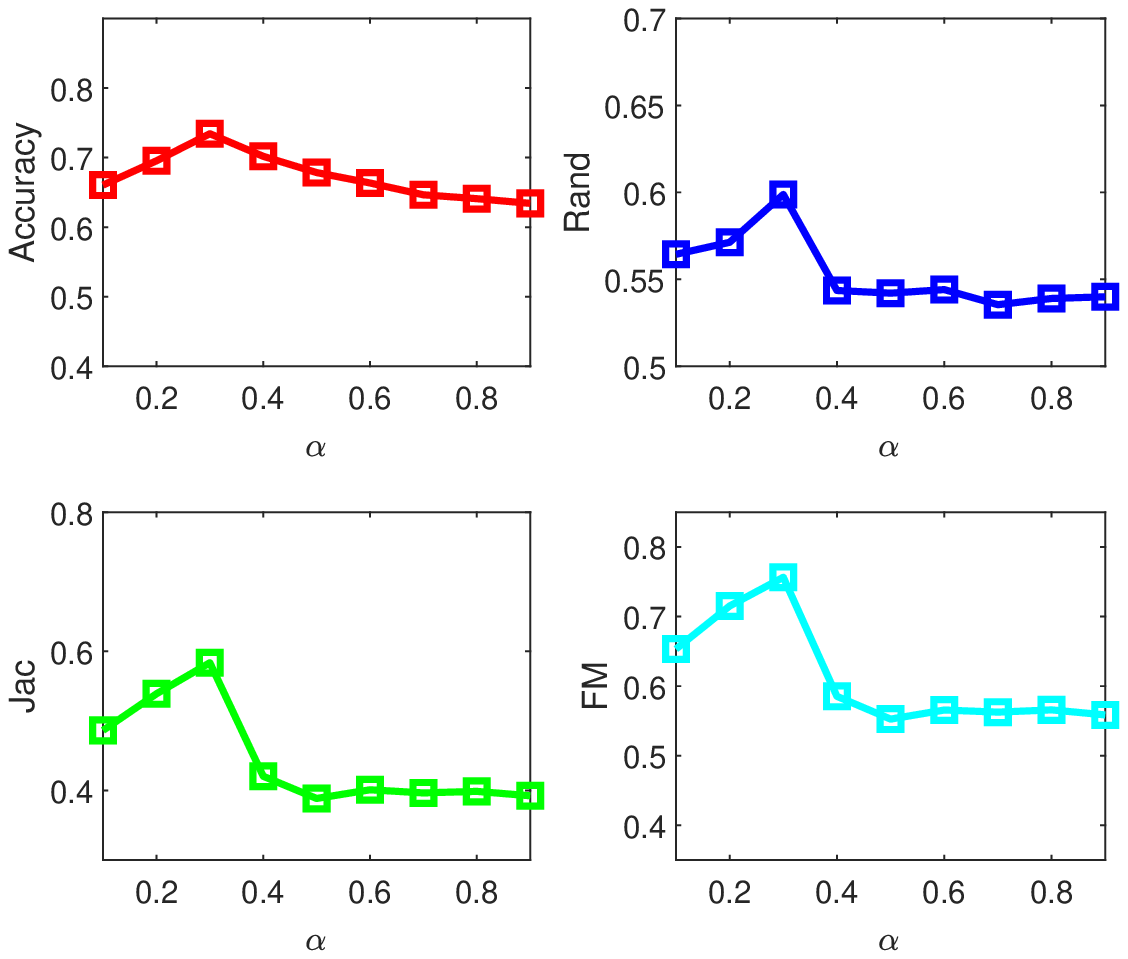}\\
  \caption{Effect of the scale coefficient on the performance.}
\label{rbm}
\end{figure}
\section{Conclusions}
We have presented Micro-DL, a micro-supervised deep representation learning architecture based on two variant models (Micro-DGRBM and Micro-DRBM), which firstly introduced small-perturbation ideology on the representation learning model from the perspective of the representation probability distributions. The positive SPI stimulated the representation probability distributions of Micro-DGRBM and Micro-DRBM models to become more similar and dissimilar in the same and different clusters, respectively. Furthermore, the representation learning capability of Micro-DL has significantly enhanced under the continuous stimulation of SPI. The Micro-DL method improved the stability of representation learning from the perspective of representation
probability distribution and significantly reduced the reliance on the labels. Experimental results demonstrated that the proposed deep Micro-DL architecture shows better performance in comparison to the baseline method, the most related shallow models and deep frameworks for clustering on twelve data sets. \\
\indent In the future, there are several interesting works: 1) to introduce small-perturbation ideology on the other neural networks; 2) to study Micro-supervised representation method with single label in each cluster; 3) to explore the performance of Micro-DL on large-scale datasets.
\section{Acknowledgement}
This work was supported by the National Science Foundation of China (Nos. 62176221, 71901158, 61876158, 61806170) and Sichuan Science and Technology Program (2021YFS0178).
\bibliography{rbm}
\bibliographystyle{IEEEtran}
\begin{IEEEbiography}[{\includegraphics [width=1in,height=1.25in,keepaspectratio]{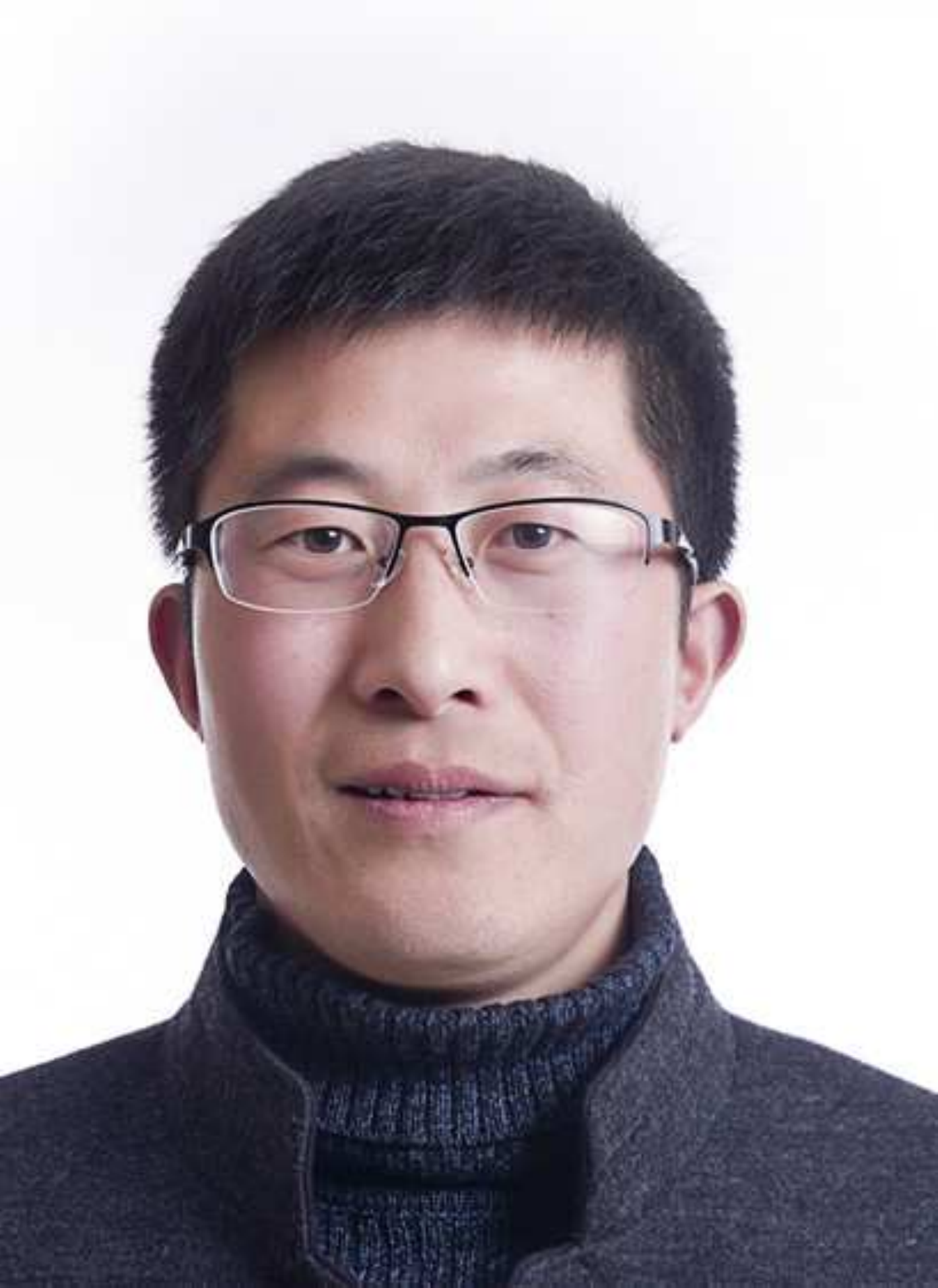}}]{Jielei Chu}
 received the B.S. and Ph.D. degrees in computer application and computer science and technology from Southwest Jiaotong University, Chengdu, China in 2008 and 2020. His research interests include deep learning, big data, semi-supervised learning and ensemble learning. He is a member of the IEEE.
\end{IEEEbiography}
\begin{IEEEbiography}[{\includegraphics [width=1in,height=1.25in,clip,keepaspectratio]{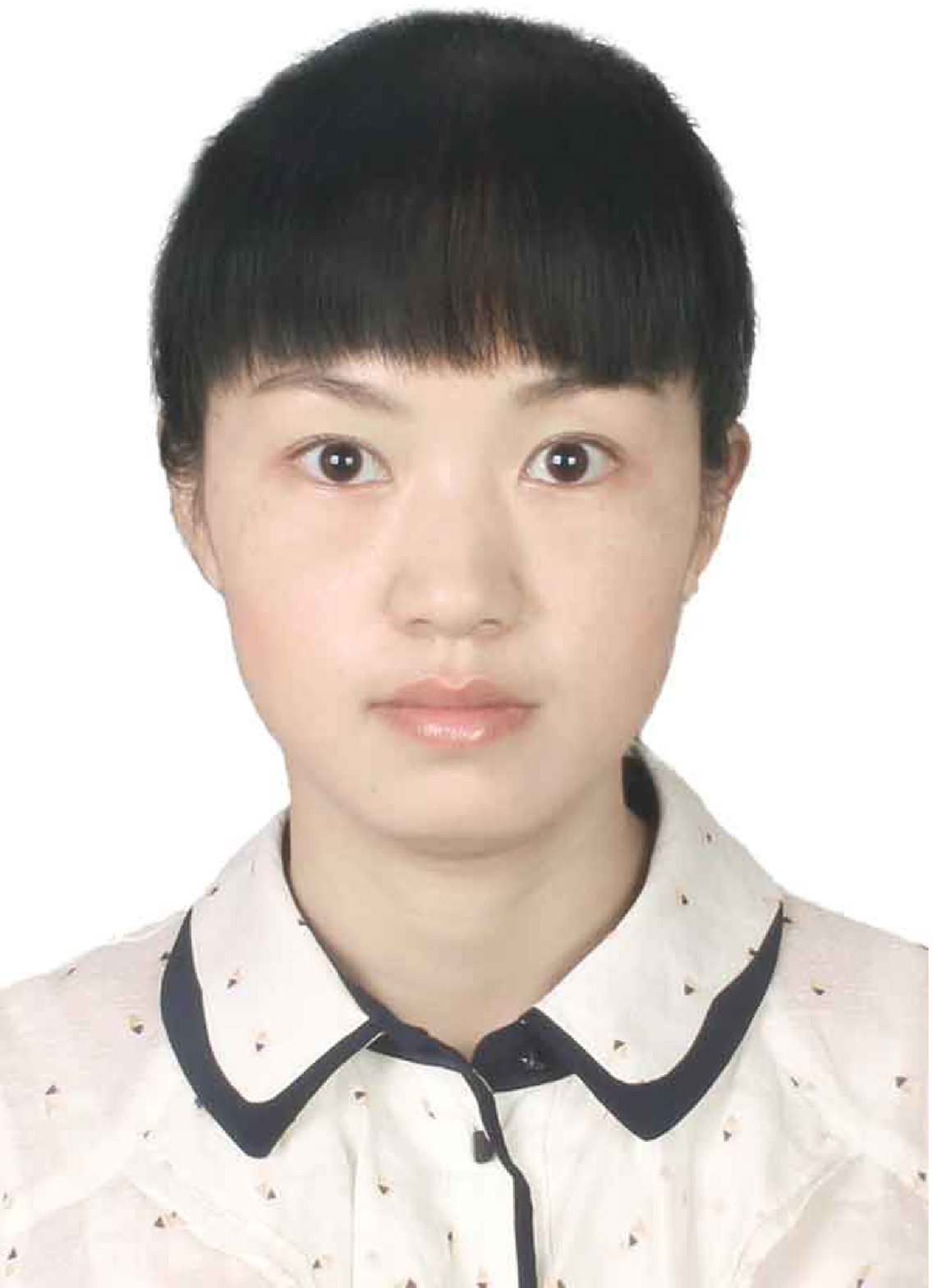}}]{Jing Liu}
 received his Ph.D. degree in management from Southwest Jiaotong University.She is currently an Assistant Professor of Business School in Sichuan University. Her research interests are machine learning, financial technology and modelling and forecasting high-frequency data.
\end{IEEEbiography}
\begin{IEEEbiography}[{\includegraphics [width=1in,height=1.25in,clip,keepaspectratio]{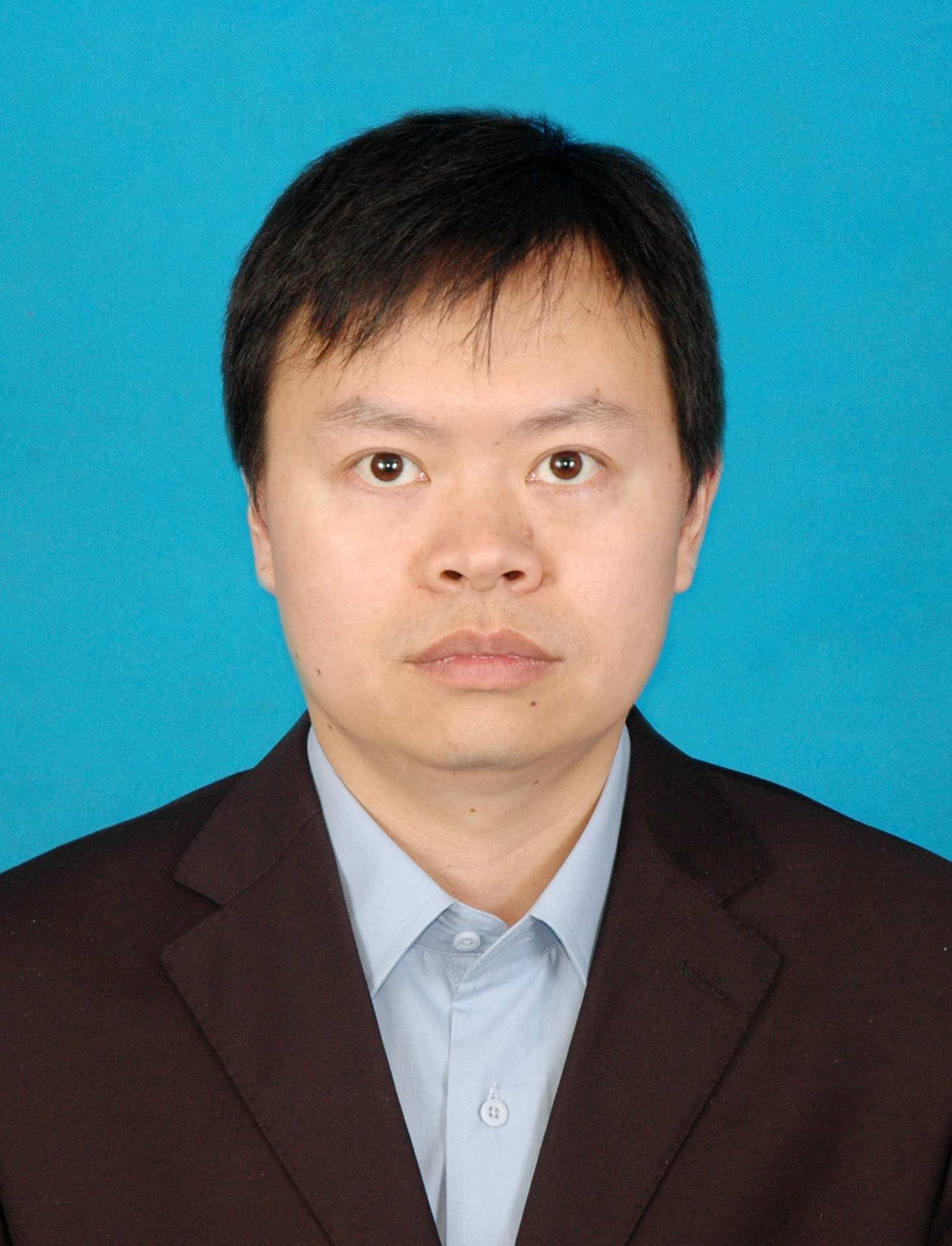}}]{Hongjun Wang}
 received his Ph.D. degree in Computer Science from Sichuan University of China in 2009. He is currently an Associate Professor of the Key Lab of Cloud Computing and Intelligent Techniques in Southwest Jiaotong University. His research interests include machine learning, data mining and ensemble learning. He has published over 50 research papers in journals and conferences and he is a member of ACM and CCF.
\end{IEEEbiography}
\begin{IEEEbiography}[{\includegraphics [width=1in,height=1in,clip,keepaspectratio]{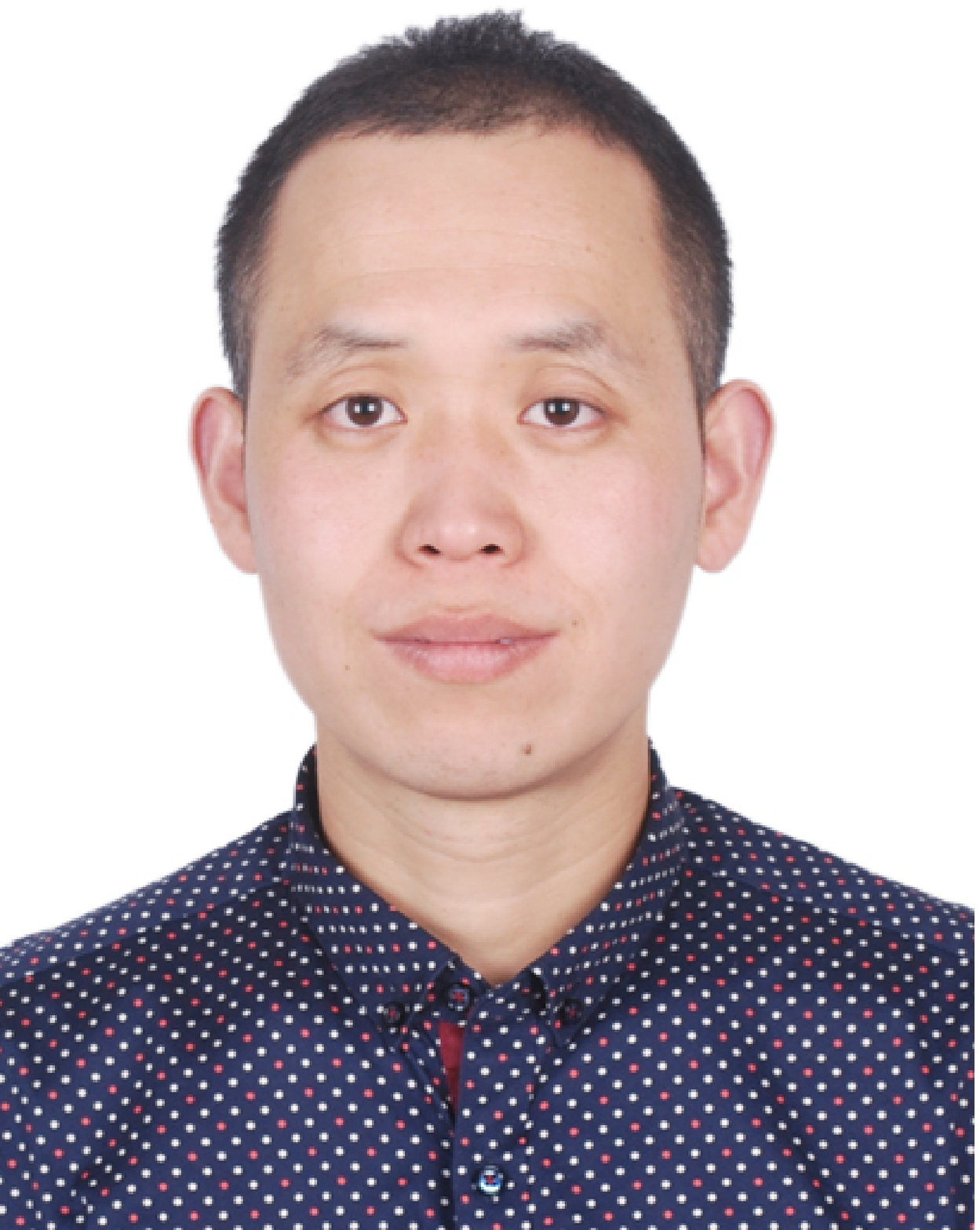}}]{Meng Hua}
 received his Ph.D. degree in Mathematics from Sichuan University of China in 2010. His research interests include belief revision, reasoning with uncertainty, machine learning, general topology.
\end{IEEEbiography}
\begin{IEEEbiography}[{\includegraphics[width=1in,height=1.25in,clip,keepaspectratio]{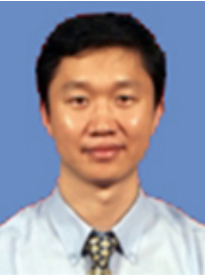}}]{Zhiguo Gong} received the Ph.D. degree in Computer Science from the Institute of Mathematics, Chinese Academy of Science, Beijing, China. He is currently a Professor with the Faculty of Science and Technology, University of Macau, China. His current research interests include machine learning, data mining, database, and information retrieval. \end{IEEEbiography}
\begin{IEEEbiography}[{\includegraphics [width=1in,height=1.25in,clip,keepaspectratio]{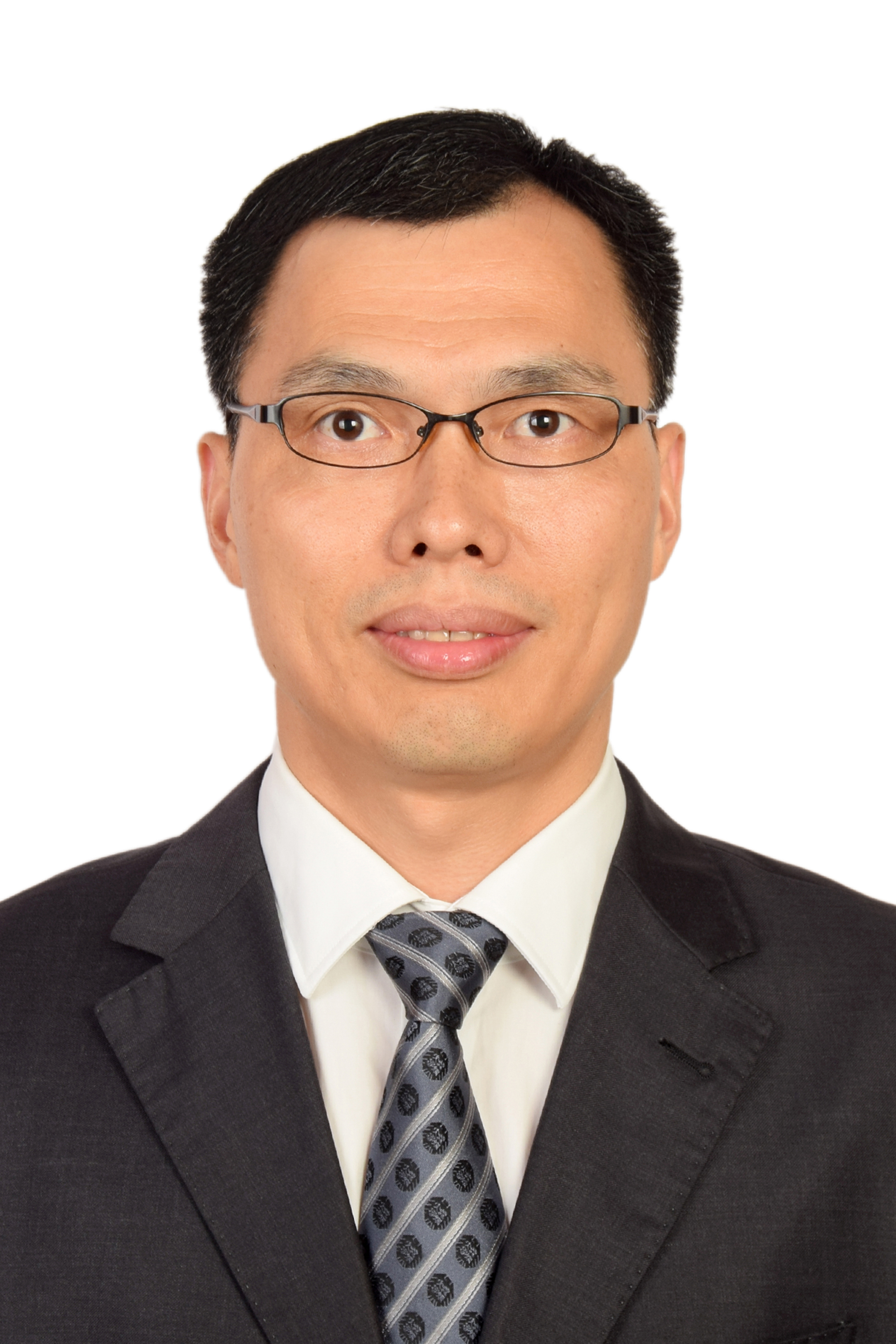}}]{Tianrui Li}
 (SM'11) received the B.S., M.S., and Ph.D. degrees from Southwest Jiaotong University, Chengdu, China, in 1992, 1995, and 2002, respectively. He was a Post-Doctoral Researcher with Belgian Nuclear Research Centre, Mol, Belgium, from 2005 to 2006, and a Visiting Professor with Hasselt University, Hasselt, Belgium, in 2008; University of Technology, Sydney, Australia, in 2009; and University of Regina, Regina, Canada, in 2014. He is currently a Professor and the Director of the Key Laboratory of Cloud Computing and Intelligent Techniques, Southwest Jiaotong University. He has authored or co-authored over 300 research papers in refereed journals and conferences. His research interests include big data, machine learning, data mining, granular computing, and rough sets.
\end{IEEEbiography}
\end{document}